\documentclass[10pt]{article}

\usepackage[utf8]{inputenc}
\usepackage[T1]{fontenc}

\usepackage{amsmath}
\usepackage{amsfonts}
\usepackage{amssymb}

\usepackage{graphicx}
\usepackage{booktabs}
\usepackage{placeins}

\usepackage{natbib}
\usepackage{microtype}
\usepackage{url}
\usepackage[hidelinks]{hyperref}

\usepackage[margin=1in]{geometry}

\title{ArgGYM: A Procedural, Engine-Verified Benchmark for Structured Defeasible Reasoning}

\author{
İbrahim Ethem Deveci, Funda Tan Çalık, Barış Deniz Sağlam, Duygu Ataman \\
Graduate School of Informatics \\
Middle East Technical University \\
Ankara, Türkiye \\
\texttt{\{ethem.deveci, funda.tan, deniz.saglam, dataman\}@metu.edu.tr}
}

\date{}

\begin{document}

\maketitle

\begin{abstract}
Recent progress in large language model reasoning has been driven by benchmarks and reinforcement learning environments with automatically verifiable rewards, particularly in mathematics, code, and formal logic. These settings make model accuracy easier to evaluate and optimize, but it remains unclear how far success under fixed problem specifications and stable evaluation criteria transfers to reasoning outside such domains. Real-world reasoning often proceeds under incomplete and revisable information: conclusions may be supported provisionally, defeated by counter-evidence, reinstated by further arguments, or revised when stronger reasons become available. Reasoning of this kind is generally referred to as \emph{defeasible reasoning}. We introduce ArgGYM, a procedural benchmark and RLVR-compatible training environment for structured defeasible reasoning. ArgGYM decomposes this reasoning into twelve tasks and grounds task-specific scoring in a symbolic argumentation engine that computes the formal states used to evaluate model outputs. It includes a frozen benchmark of 1,440 verified instances across fifteen curriculum configurations, two argument preference orderings (weakest-link and last-link), and two set orderings (elitist and democratic), while the same generators and verifiers can produce fresh instances, reducing dependence on static test sets, for both evaluation and verifiable-reward training. On the frozen benchmark, frontier and open-weight models show sharply different reasoning profiles: they can recover substantial parts of structured answers without solving the complete task, and performance declines in later curriculum configurations with longer dependencies and more interacting structures. Training Qwen3.5-4B with ArgGYM's verifiers as rewards more than doubles complete-task success on held-out instances (20.4\% to 53.3\%), including task configurations and levels never seen in training, with the largest gains on constructive tasks. We release the benchmark, generators, and verifiers for reproducible evaluation and RLVR training.\footnote{\url{https://github.com/ieddeveci/ArgGYM}}
\end{abstract}

\section{Introduction}

Advances in large language models and reasoning-focused models have been driven in part by reinforcement learning with verifiable rewards (RLVR), especially in mathematics, code, and formal logic, where solution accuracy can be checked automatically \citep{deepseekmath,deepseekr1,logicrl}. These domains are important tests of reasoning because they require precise inference and often produce outcomes that can be checked against explicit criteria. Many real-world reasoning problems, however, do not have fixed specifications or stable criteria in the same way. A lawyer may revise an earlier judgment when new evidence is introduced or when competing reasons must be weighed against one another; a physician may have to reach a decision from incomplete information while remaining prepared to revise it as new evidence becomes available. Reasoning of this kind is commonly treated as defeasible: conclusions are warranted given the available information, but may later be withdrawn when stronger or conflicting reasons arise. It is therefore non-monotonic, in the sense that adding new information can invalidate a conclusion that was previously supported. Defeasible and non-monotonic reasoning have long been studied in artificial intelligence, cognitive science, legal reasoning, and argumentation as models of reasoning under incomplete information, conflict, exceptions, and revision \citep{reiter1980default,pollock1987defeasible,prakken1997argument, prakken2010structured}. If language models are to operate reliably in domains where information is incomplete and judgments remain open to revision, the ability to reason defeasibly is likely to be an important part of their broader reasoning competence. Evaluating this ability is therefore necessary for understanding how well current models can handle reasoning under incomplete, conflicting, and revisable information.

Recent efforts have begun to evaluate the defeasible and non-monotonic reasoning capabilities of LLMs
\citep{xiu2022logicnmr,kazemi2023boardgameqa,parmar2024logicbench,
wilie2024belief,allaway2025defreasing,li2025delp}. These studies report weaknesses in generalization, belief revision, and reasoning under conflicting or revisable information. More recent work has also developed procedural or formally verified benchmarks and generation frameworks for particular defeasible reasoning problems
\citep{cooper2026defab,leiva2026delpgen,sadhu2026derelab}. Yet the evaluation and training infrastructure for defeasible reasoning remains much less developed than in mathematics, code, and deductive logic, where procedural generation and exact verification can be used both to benchmark models and to provide training rewards.

Taken together, these developments leave open a complementary gap: a unified executable environment in which multiple structured defeasible-reasoning tasks are instantiated over the same formal substrate, separately verified under task-specific contracts, and varied systematically in their structural and preference-sensitive demands. Such an environment should support both a frozen benchmark for reproducible model comparison and fresh procedural generation using the same verification machinery. This combination is useful for evaluation because it exposes operation-specific capability profiles within a common formal system, while for reinforcement learning it enables fresh verified training instances without drawing from the frozen evaluation suite. Procedural generation can reduce dependence on static evaluation sets and their associated contamination risks \citep{white2025livebench,chen2025contamination,xie2025memorization}, while mechanically verifiable outcomes can serve directly as rewards for reinforcement learning \citep{stojanovski2025reasoninggym}.

Structured argumentation provides a natural formal basis for this setting. \citet{dung1995acceptability} established a fundamental connection between argumentation and non-monotonic reasoning, showing how forms of non-monotonic inference can be understood in terms of arguments, attacks, and acceptability. Structured argumentation frameworks such as ASPIC+ additionally make the
internal construction of arguments explicit: arguments are built from premises and inference rules, conflicts give rise to attacks, preferences determine whether preference-sensitive attacks succeed as defeats, and argumentation semantics determine which arguments, and consequently which conclusions, are accepted \citep{prakken2010structured,modgil2014aspic}. Because the resulting argumentative states can be computed mechanically, structured argumentation supports both controlled decomposition of reasoning capabilities and executable verification of model outputs.

ArgGYM operationalizes this connection as an ASPIC+-based procedural benchmark and RLVR-compatible environment. Its twelve tasks cover formalization and support recovery, defeat diagnosis, semantic evaluation, theory revision, preference reasoning, and constructive intervention. Generated theories vary in structural composition, argument preference ordering, and set ordering, while a symbolic argumentation engine computes or validates the formal states required by the task-specific scorers. For reproducible comparison, ArgGYM includes a frozen benchmark of 1,440 verified instances spanning all twelve tasks, fifteen curriculum configurations, two argument preference orderings (weakest-link and last-link), and two set orderings (elitist and democratic). The same generators and verifiers can also produce fresh verified instances.

Across 18 distinct model variants spanning frontier and open-weight models, performance varies markedly across reasoning operations: models with similar aggregate accuracy can exhibit different task profiles, and success on formalization or support recovery does not imply reliable semantic evaluation, defeat analysis, revision, or intervention. These results motivate evaluating structured defeasible reasoning at the level of its component operations rather than as a single undifferentiated capability. Training a 4B-parameter model with ArgGYM's verifiers as rewards more than doubles its complete-task success on held-out instances, mainly through constructive intervention, while its transfer to external benchmarks is selective.

Our contributions are:
\begin{itemize}

\item We introduce ArgGYM, a twelve-task procedural environment for structured defeasible reasoning based on ASPIC+.

\item We provide a frozen 1,440-instance benchmark together with procedural generators that produce fresh, engine-verified instances across controlled curriculum configurations, argument preference orderings, and set orderings.

\item We show that ArgGYM's verifiers serve as training rewards: GRPO training of Qwen3.5-4B raises complete-task success on held-out instances from 20.4\% to 53.3\%, including configurations and levels never trained on, and we characterize the resulting transfer on thirteen external benchmarks.

\end{itemize}

\section{Related Work}

\paragraph{Defeasible reasoning.}
Several benchmarks test language models in settings where conclusions can change as information is added or conflicts arise. $\delta$-NLI asks whether additional evidence strengthens or weakens a natural-language inference \citep{rudinger2020skeptic}; LogicNMR isolates non-monotonic reasoning through explicit default rules and iterative updates \citep{xiu2022logicnmr}; BoardgameQA introduces contradictory information resolved through preferences over information sources \citep{kazemi2023boardgameqa}; and Belief-R tests whether models revise earlier conclusions when new evidence becomes available \citep{wilie2024belief}. More formal approaches evaluate rule-based defeasible reasoning directly. These include Defeasible Logic reasoning patterns \citep{tachmazidis2024benchmark}, formal defeasible reasoning over synthetic DeLP programs with varying reasoning depth \citep{li2025delp}, and defeasible property inheritance in DEFREASING \citep{allaway2025defreasing}. These benchmarks establish defeasible reasoning as a distinct LLM evaluation problem, but generally focus on particular inference problems or formalisms rather than decomposing structured argumentation into a broader set of separately verifiable operations.

\paragraph{Computational argumentation.}
Recent work has examined the relationship between LLMs and argumentation from two directions. In one, argumentation itself is treated as a model capability: evaluations cover natural-language argument mining and generation \citep{chen2024computationalargumentation}, ArgBench unifies 33 datasets across 46 computational argumentation tasks \citep{ajjour2026argbench}, and work on abstract argumentation evaluates and trains language models to compute argument acceptability over generated argumentation frameworks \citep{li2026conflict}. In the other direction, argumentation is used as formal machinery for improving, verifying, or revising LLM outputs. MQArgEng introduces formal argumentation semantics into an LLM pipeline \citep{castagna2024formalargumentation}; ArgLLMs construct argumentation frameworks that support explainable and contestable claim verification \citep{freedman2025argllms}; LLM-ASPIC$^{+}$ delegates conflict resolution to an ASPIC$^{+}$ component after LLM-based grounding and reasoning \citep{fang2025llmaspic}; and ARGUS represents LLM self-explanations as Dung-style argumentation frameworks and computes minimum-cost repairs when new evidence changes the required acceptability status
\citep{xiao2026argus}.

\paragraph{Procedural and verifiable reasoning.}
Synthetic reasoning benchmarks established that formally generated theories can support controlled evaluation of model reasoning. RuleTaker generates natural-language facts and rules for multi-step inference \citep{clark2020ruletaker}, while ProofWriter extends this setting with explicit proofs and abductive reasoning \citep{tafjord2021proofwriter}. ProverGen, which produces the ProverQA benchmark, combines scalable data generation with symbolic prover verification to construct first-order-logic reasoning problems and their intermediate reasoning steps \citep{qi2025provergen}. More recently, LogicSkills separates formal reasoning into symbolization, countermodel construction, and validity assessment, with instances checked using an SMT solver \citep{rabern2026logicskills}. Reasoning Gym and Enigmata scale this pattern into multi-task procedural environments with controllable difficulty and programmatic verification suitable for RLVR \citep{stojanovski2025reasoninggym,chen2025enigmata}. Related developments are now appearing in defeasible reasoning itself. DeFAb procedurally constructs defeasible-abduction problems with an exact verifier that can serve as a training reward \citep{cooper2026defab}; DeReLab generates formally verified belief-updating problems across default and inheritance reasoning \citep{sadhu2026derelab}; and DPG generates synthetic DeLP knowledge bases with controllable structural properties for benchmarking structured-argumentation systems \citep{leiva2026delpgen}.

\paragraph{Positioning.} ArgGYM is complementary to these lines of work. Existing resources cover computational argumentation broadly through heterogeneous datasets (ArgBench), generate or verify instances of a single defeasible-reasoning problem (DeFAb, DeReLab, DPG), or provide procedural, verifiable environments for RLVR in domains other than defeasible reasoning (Reasoning Gym, Enigmata, ProverGen). ArgGYM brings these properties together in one executable structured-argumentation environment. Its twelve tasks share an ASPIC+ substrate and common engine semantics, and they span both the analysis of an argumentative state and constructive intervention on it, including a matched comparison of two intervention contracts. Each task has its own verification contract, which applies equally to the frozen benchmark and to freshly generated instances; constructive answers are accepted whenever executing them produces the required argumentative state, not by comparison with a stored reference. Because every task rests on the same formalism, failures can be compared across reasoning operations and structural conditions, and the same verifiers serve directly as training rewards (Section~\ref{sec:training}).

\section{ArgGYM}

\subsection{Formal Setting and Verification}

An ASPIC+ argumentation theory consists of a knowledge base, strict and defeasible inference rules, conflict relations, and preferences \citep{modgil2014aspic}. Arguments are constructed recursively from premises by applying strict or defeasible rules. Conflicts give rise to three standard attack types: \emph{undermining} targets an ordinary premise, \emph{rebutting} targets a conclusion reached defeasibly, and \emph{undercutting} targets the applicability of a defeasible inference. Preferences determine whether preference-sensitive attacks succeed as defeats, and argumentation semantics determine which arguments, and consequently which claims, are accepted.

ArgGYM uses a pinned PyArg backend for executable ASPIC+ computation \citep{odekerken2023pyarg}, on which it implements the task-specific procedural layer (Appendix~\ref{app:formal}). The engine validates candidate instances and, for constructive tasks, executes submitted interventions and recomputes the resulting argumentative state, so constructive outputs are evaluated by their formal consequences while the remaining outputs are scored against engine-computed or engine-validated targets. Because every score rests on this backend, we validated it against a separately implemented engine. Arg2P \citep{calegari2022arg2p}, an independent ASPIC+ implementation, reproduces ArgGYM's stored states on every grounded status and every semantics-query answer across all 11,400 theory evaluations of the released taskset, including those obtained after perturbations and reference constructions, and an independent implementation of Dung's semantics recomputes every grounded status from the argument graphs. The two engines differ in one convention, how rules with identical content are identified. In the audited runs, re-scoring under Arg2P changed no reported overall score by more than 0.1 percentage points (Appendix~\ref{app:formal}).

ArgGYM crosses two argument preference orderings, \emph{last-link} and \emph{weakest-link}, with two set orderings, \emph{elitist} and \emph{democratic}, yielding four preference configurations. All tasks except \texttt{semantics\_query} use grounded semantics, which has a unique extension. The \texttt{semantics\_query} task additionally requires reasoning under alternative argumentation semantics and acceptance modes. Full definitions and implementation details are given in Appendix~\ref{app:formal}.

\subsection{Procedural Generation}

Each instance $x = G(\tau,\ell,p,c,s)$ is generated from a task $\tau \in \{1,\ldots,12\}$, curriculum configuration $\ell \in \{1,\ldots,15\}$, argument preference ordering $p$, set ordering $c$, and random seed $s$, following a \emph{generate--verify} procedure. A task-specific generator constructs a candidate theory and task instance, structural gates check that it exhibits the reasoning phenomenon required by the task, and the symbolic engine computes the relevant argumentative state. Candidates that fail either check are rejected.

The fifteen curriculum configurations vary the structural composition of the generated problems, controlling task-specific factors such as derivation depth, branching, interacting attackers and defenders, distractors, preference-sensitive conflicts, and multi-goal interactions (Appendix~\ref{app:generation}). They represent increasing structural richness rather than calibrated difficulty, and together with the orderings and seed they define the space from which both frozen and fresh instances are generated.

\subsection{Task Suite}

ArgGYM operationalizes structured defeasible reasoning through twelve tasks over the same ASPIC+ formal substrate. The suite covers representation and support recovery, diagnosis and semantic evaluation, revision prediction, and constructive intervention. This decomposition allows model performance to be localized to particular reasoning operations. Table~\ref{tab:tasks} summarizes the tasks.

\begin{table*}[t]
\centering
\small
\caption{The twelve ArgGYM tasks. The first six require structured analysis or formalization outputs, while the final six require constructive interventions whose consequences are executed and verified through the task-specific scoring pipeline.}
\label{tab:tasks}
\begin{tabular}{@{}p{0.25\linewidth}p{0.71\linewidth}@{}}
\toprule
Task & Objective \\
\midrule

\multicolumn{2}{@{}l}{\textit{Representation and support}} \\
\addlinespace[2pt]

\texttt{formalization}
& Construct an executable ASPIC+ theory from a controlled natural-language
description whose induced argumentative state is behaviorally equivalent to
the intended theory. \\

\texttt{claim\_chain}
& Recover the complete support derivation for a justified target claim in an ordering consistent with its derivational dependencies. \\

\midrule
\multicolumn{2}{@{}l}{\textit{Diagnosis, semantics, and revision}} \\
\addlinespace[2pt]

\texttt{defeat\_diagnosis}
& Identify where support for a non-justified target fails and the relevant
defeaters responsible for that failure. \\

\texttt{status\_query}
& Determine the statuses of queried claims under grounded semantics. \\

\texttt{semantics\_query}
& Determine claim acceptance under a specified argumentation semantics. \\

\texttt{perturbation}
& Identify which claims of the original theory change status after a specified update and predict the new status of each changed claim. \\

\midrule
\multicolumn{2}{@{}l}{\textit{Constructive intervention}} \\
\addlinespace[2pt]

\texttt{preference\_\allowbreak construction}
& Add preferences that produce a requested configuration of claim statuses. \\

\texttt{counter\_argument}
& Construct an intervention that justifies the opposing claim without adding strict rules. \\

\texttt{counter\_argument\_\allowbreak strict}
& Strict-rule-enabled condition paired with \texttt{counter\_argument} on the same underlying theory and semantic objective. \\

\texttt{attack}
& Modify the theory so that a justified target becomes overruled. \\

\texttt{defence}
& Modify the theory so that an attacked target becomes justified. \\

\texttt{attack\_defense}
& Modify the theory so that one target becomes overruled while another becomes
justified. \\

\bottomrule
\end{tabular}
\end{table*}

\subsection{Frozen Benchmark}

To evaluate models, we freeze a benchmark of $12 \times 15 \times 2 \times 2 \times 2 = 1{,}440$ instances sampled from the ArgGYM generation space: two accepted instances for each combination of task, curriculum configuration, argument preference ordering, and set ordering. These instances remain fixed across all model evaluations. The procedural generators remain separate from the frozen suite and can produce additional verified instances when fresh problems are required.

The same task-specific scoring contracts are used for frozen and newly generated instances. On fresh instances, their verifier outputs can also serve as mechanically computed rewards, making ArgGYM RLVR-compatible without a separate learned reward model. Section~\ref{sec:training} uses these rewards to train a model.

\section{Experimental Setup}

\subsection{Models and Inference Configuration}
\label{sec:model-setup}

We evaluate 18 distinct language-model variants from five model families (Gemma, GPT-OSS, Qwen, DeepSeek, and Gemini), yielding 42 model--reasoning configurations that span a broad range of model scales and inference-time reasoning settings. Where supported, we evaluate several reasoning configurations of the same model: thinking enabled or disabled for Gemma and Qwen3.5, and graded reasoning effort for GPT-OSS, Qwen3.8-27B, DeepSeek-v4.1-Flash, and Gemini. Gemini and DeepSeek-v4.1-Flash are accessed through hosted APIs; the open-weight models are served on NVIDIA H200 GPUs. Every configuration is evaluated on the same frozen 1,440-instance benchmark with the same taskset, prompt version, response template, and scoring version, and inference settings are fixed within each configuration. Model identifiers, reasoning settings, and inference configurations are reported in Appendix~\ref{app:model-configs}.

\subsection{Evaluation Metrics and Success Criteria}
\label{sec:evaluation-metrics}

We report \emph{complete-task success} and a \emph{task-native score}. Complete-task success, our primary metric, indicates whether the full task-specific output contract is satisfied: exact recovery of the required structured output for prediction tasks, behavioral equivalence with the reference theory for \texttt{formalization}, and, for constructive tasks, satisfaction of all requested postconditions after execution with the resulting theory remaining consistent. The task-native score in $[0,1]$ captures task-specific partial correctness, such as partial recovery of structured outputs, diagnostic accuracy, or intervention economy. Because its semantics differ across tasks, it is interpreted only within the corresponding task. Full scoring definitions are given in Appendix~\ref{app:scoring}.

\section{Results}
\label{sec:results}

\subsection{Overall Performance and Capability Profiles}

ArgGYM separates the evaluated systems over a wide range of complete-task success, from near-zero performance in the smallest configurations to 82.4\% for Gemini 3.8 Flash at high reasoning effort (Table~\ref{tab:overall-results}). Additional inference-time reasoning generally helps: Gemma-4-31B rises from 14.4\% to 27.3\% with thinking enabled, GPT-OSS-120B from 12.4\% at low effort to 29.0\% at high effort, and Qwen3.8-27B from 28.4\% to 39.2\% at xhigh effort. The effect is nevertheless model dependent rather than uniformly monotonic; Gemini 2.5 Pro, for example, peaks at medium rather than high effort.

Aggregate performance also does not correspond to a uniform capability profile. Even Gemini 3.8 Flash at high effort, which attains the highest benchmark-wide success, is near ceiling on several tasks yet reaches only 57.5\% complete success on \texttt{defeat\_diagnosis} and 60.0\% on the strict-rule-enabled counter-argument condition. Lower-scoring systems likewise show different mixtures of strengths and weaknesses rather than uniformly scaled versions of the same behavior.

\begin{table}[t]
\centering
\caption{\textbf{Overall ArgGYM performance.} Complete-task success (\%) for every reasoning
configuration; bold marks each model's best configuration. For that configuration, the last four
columns report success in the first and last curriculum bands and under the last-link and
weakest-link principles (full results in Appendices~\ref{app:curriculum-analysis}
and~\ref{app:ordering-analysis}).}
\label{tab:overall-results}
\scriptsize
\setlength{\tabcolsep}{4pt}
\begin{tabular}{@{}llrrrr@{}}
\toprule
& & \multicolumn{2}{c}{Curriculum band} & \multicolumn{2}{c}{Link principle} \\
\cmidrule(lr){3-4}\cmidrule(l){5-6}
Model & Complete-task success by configuration (\%) & L1--5 & L11--15 & Last & Weakest \\
\midrule
Gemma 4 E2B       & off 1.1 $\rightarrow$ on \textbf{1.3}                 & 4.0  & 0.0  & 1.5  & 1.1 \\
Gemma 4 E4B       & off 3.5 $\rightarrow$ on \textbf{3.8}                 & 10.2 & 0.4  & 4.9  & 2.6 \\
Gemma 4 26B-A4B   & off 9.0 $\rightarrow$ on \textbf{14.1}                & 25.8 & 6.9  & 20.8 & 7.4 \\
Gemma 4 31B       & off 14.4 $\rightarrow$ on \textbf{27.3}               & 42.3 & 16.7 & 37.4 & 17.2 \\
\midrule
GPT-OSS 20B       & low 6.2 / med.\ 16.0 / high \textbf{21.2}             & 34.2 & 12.5 & 29.9 & 12.6 \\
GPT-OSS 120B      & low 12.4 / med.\ 21.5 / high \textbf{29.0}            & 45.2 & 16.0 & 35.4 & 22.5 \\
\midrule
Qwen3.5 0.8B      & off 0.0 $\rightarrow$ on \textbf{0.0}                 & 0.0  & 0.0  & 0.0  & 0.0 \\
Qwen3.5 2B        & off 0.3 $\rightarrow$ on \textbf{1.0}                 & 2.9  & 0.0  & 1.2  & 0.8 \\
Qwen3.5 4B        & off 12.4 $\rightarrow$ on \textbf{22.4}               & 40.6 & 10.2 & 31.1 & 13.6 \\
Qwen3.5 9B        & off 17.8 $\rightarrow$ on \textbf{26.1}               & 42.5 & 13.8 & 35.8 & 16.4 \\
Qwen3.5 35B-A3B   & off 21.7 $\rightarrow$ on \textbf{31.7}               & 49.4 & 17.1 & 42.8 & 20.6 \\
Qwen3.5 27B       & off 31.5 $\rightarrow$ on \textbf{39.3}               & 58.8 & 26.2 & 50.1 & 28.5 \\
Qwen3.8 27B       & low 28.4 / med.\ 29.4 / xhigh \textbf{39.2}           & 50.0 & 28.7 & 51.1 & 27.2 \\
\midrule
DeepSeek V4.1 Flash & high \textbf{45.0}                                  & 57.5 & 35.0 & 50.7 & 39.3 \\
\midrule
Gemini 2.5 Pro         & low 16.6 / med.\ \textbf{34.5} / high 33.2       & 48.5 & 23.1 & 44.0 & 25.0 \\
Gemini 3.5 Flash Lite  & low 18.2 / med.\ 29.3 / high \textbf{35.1}       & 54.2 & 20.6 & 44.3 & 26.0 \\
Gemini 3.1 Pro Preview & low 48.3 / med.\ 54.9 / high \textbf{62.9}       & 76.7 & 49.6 & 70.8 & 55.0 \\
Gemini 3.8 Flash       & low 61.5 / med.\ 77.8 / high \textbf{82.4}       & 88.1 & 76.5 & 92.9 & 71.8 \\
\bottomrule
\end{tabular}
\end{table}

Four findings summarize the main empirical patterns, after which we relate ArgGYM to existing defeasible-reasoning benchmarks:

\paragraph{Partial correctness is high where complete success is low.}
A persistent gap separates task-native partial correctness from complete-task
success. Qwen3.5-27B with thinking enabled reaches 88.7 native on
\texttt{status\_query} but only 35.8\% complete success, while
DeepSeek-v4.1-Flash reaches 89.7 native but only 34.2\% complete success on the
same task. Gemini 3.8 Flash at high effort similarly reaches 93.6 native on
\texttt{defeat\_diagnosis} while completing only 57.5\% of instances. Most of these failures are errors of judgment: in 20 of these configurations, 82.6\% of failed analysis answers state at least one wrong judgment, and only 6 of 11,671 (0.1\%) are correct answers in the wrong layout (Appendix~\ref{app:failure-analysis}). Models can therefore recover large portions of the required argumentative state without producing a completely correct answer.

\paragraph{Complete success falls faster than partial correctness in later curriculum configurations.}
Complete-task success generally falls sharply in later curriculum bands. From
L1--L5 to L11--L15, GPT-OSS-120B at high effort falls from 45.2\% to 16.0\%,
Qwen3.5-27B with thinking enabled from 58.8\% to 26.2\%, and
DeepSeek-v4.1-Flash from 57.5\% to 35.0\%; even Gemini 3.8 Flash at high effort
declines from 88.1\% to 76.5\%. The gap between local and complete correctness
can widen sharply: on \texttt{status\_query}, Qwen3.5-27B with thinking enabled
falls from 72.5\% complete success with a 95.1 native score in L1--L5 to only
10.0\% complete success while retaining an 84.5 native score in L11--L15. Once the structural properties of the generated theories are controlled, level no longer has a detectable effect in 11 of the 12 tasks, and on \texttt{status\_query} wrong claims are associated with longer derivations and undecided status rather than with the number of queried claims (Appendix~\ref{app:failure-analysis}). Low complete success is also not an artefact of the all-or-nothing metric compounding independent per-claim errors: observed exact success exceeds the success predicted under independent per-claim errors in 45 of the 47 configuration--band combinations in which a model solves any instance, so errors concentrate in particular instances (Appendix~\ref{app:failure-analysis}). Complete success thus falls as structure grows richer, while high within-task partial correctness remains.

\paragraph{Preference regimes change model behavior, but not
according to a universal difficulty ordering.}
Many configurations perform markedly better under last-link than weakest-link.
Gemini 3.8 Flash at high effort scores 92.9\% versus 71.8\%, and
Qwen3.5-27B with thinking enabled 50.1\% versus 28.5\%. Yet task-level
reversals occur: DeepSeek-v4.1-Flash, for example, performs better under
weakest-link on \texttt{claim\_chain} (66.7\% versus 53.3\%) and
\texttt{defeat\_diagnosis} (73.3\% versus 28.3\%). Weakest-link should
therefore not be interpreted as uniformly harder. Because these ordering cells
are generated independently, the benchmark-wide comparison is descriptive
rather than a causal effect of changing the ordering on a fixed theory. A matched re-evaluation points the same way for one task: re-posed under last-link, the 60 weakest-link \texttt{status\_query} theories are solved more often by the untrained Qwen3.5-4B (16.7\% to 26.7\%, +10.0 points, 95\% CI [1.7, 20.0]), while the 60 last-link theories re-posed under weakest-link fall from 23.3\% to 20.0\% ($-3.3$ [$-15.0$, 10.0]). Weakest-link \texttt{preference\_construction}, whose generated instances all require a preference between premises, is rarely solved by any model: no configuration exceeds 26.7\%, and 31 of the 42 solve at most 2 of 60 instances. Full
ordering analyses are reported in Appendix~\ref{app:ordering-analysis}.

\paragraph{Expanding the intervention language changes success in both directions.}
The matched \texttt{counter\_argument} / \texttt{counter\_argument\_strict} pair holds the underlying theory and target statuses fixed; the strict variant also permits strict-rule additions, which lower the recorded minimum and with it the largest accepted answer, twice that minimum. Model behavior changes in opposite
directions. Gemini 3.8 Flash at high effort falls from 86.7\% success in
\texttt{counter\_argument} to 60.0\% in the strict-rule-enabled condition. Qwen3.8-27B provides an even sharper contrast: from low to xhigh effort, its
overall success rises from 28.4\% to 39.2\% and its
\texttt{counter\_argument} success from 25.8\% to 42.5\%, while success in the
strict-rule-enabled condition falls from 50.8\% to 11.7\%. 
Conversely, across the matched conditions, Qwen3.5-9B with thinking enabled rises from 20.8\% to 44.2\%, and
Qwen3.5-35B-A3B from 15.8\% to 42.5\%. Constructive success therefore depends on the intervention contract, not only on whether an effective intervention exists.

\paragraph{Relation to other defeasible benchmarks.}
On a panel of 12 of these configurations, models that score higher on ArgGYM also tend to score higher on $\delta$-NLI and LogicNMR (rank correlations 0.82 and 0.70), but the correspondence is weaker for BoardgameQA, LogicBench, and Multi-LogiEval, and weakly reversed for DEFREASING ($-0.38$). Several of these benchmarks place the configurations in a narrow range: all twelve configurations fall between 81.7\% and 85.5\% on LogicBench, while their ArgGYM success spans 1.3\% to 39.3\% (Appendix~\ref{app:panel}). ArgGYM therefore complements existing defeasible-reasoning benchmarks.

\section{Training with ArgGYM}
\label{sec:training}

To test whether ArgGYM's verifiers can serve as training rewards, we post-train Qwen3.5-4B (thinking
enabled) with GRPO \citep{deepseekmath} and a LoRA adapter \citep{hu2022lora}, and evaluate it before
and after training on ArgGYM and on thirteen external benchmarks.

\paragraph{Setup.}
Training problems are freshly generated from 82 task--level combinations (\emph{training cells}) in
levels 1--12, with seeds disjoint from every evaluation set. The reward is 1 for complete-task success
and 0 otherwise, with partial credit for failed answers on four analysis tasks (status query, defeat
diagnosis, formalization, perturbation) and zero reward for answers truncated at 40,960 tokens. An
adaptive sampler favours cells the current model solves about half the time. We select step 200, the last checkpoint and the best of steps 20--200 on a 96-problem development split. Hyperparameters, the cell-selection rule and
the schedule are given in Appendix~\ref{app:training}. We evaluate on two sets: the frozen benchmark
of Section~\ref{sec:results}, and a \emph{held-out} set of 1,440 released instances with the same
design, disjoint from the frozen benchmark and never used in training, cell selection or checkpoint
selection. The held-out set is our primary test set, because the base model's frozen-benchmark scores
informed the choice of training cells. ArgGYM evaluations use the protocol of Section~\ref{sec:model-setup}; external benchmarks use the protocol of Appendix~\ref{app:external}. Step 200 is pooled over three inference runs of the selected checkpoint; this is one training run.

\begin{table}[t]
  \centering
  \caption{\textbf{Qwen3.5-4B before and after training with ArgGYM.} Step 200 is pooled over three
  inference runs of the selected checkpoint; $\Delta$ is the change from the base model with a 95\%
  paired bootstrap interval. ArgGYM reports complete-task success; external benchmarks report
  accuracy on their full test sets (Belief-R: BREU; BBEH: micro average). For benchmarks where truncation reaches 1\%, the last column gives the difference among answers neither model truncated (run~1); the rest of the full difference comes from items that at least one model truncated, on which the trained model truncates less often than the base model. Results at every checkpoint and by subset are in
  Appendices~\ref{app:training} and~\ref{app:external}.}
  \label{tab:training}
  \small
  \setlength{\tabcolsep}{4pt}
  \begin{tabular}{@{}llrrcc@{}}
    \toprule
    & & Base & Step 200 & $\Delta$ [95\% CI] & $\Delta$, neither truncated \\
    \midrule
    \multicolumn{6}{@{}l}{\emph{ArgGYM, held-out set}} \\
    & All tasks                          & 20.4 & 53.3 & $+32.9$ [30.5, 35.3] & \\
    & \quad Construction (6 tasks)       & 24.4 & 80.0 & $+55.6$ [52.1, 58.9] & \\
    & \quad Analysis (6 tasks)           & 16.4 & 26.6 & $+10.2$ [7.7, 12.7]  & \\
    & \quad Untrained cells (L1--12)     & 15.5 & 45.2 & $+29.6$ [25.5, 33.7] & \\
    & \quad Levels 13--15                & 6.2  & 38.4 & $+32.2$ [27.1, 37.3] & \\
    \multicolumn{6}{@{}l}{\emph{ArgGYM, frozen benchmark}} \\
    & All tasks                          & 20.8 & 52.7 & $+31.8$ [29.9, 33.9] & \\
    \midrule
    \multicolumn{6}{@{}l}{\emph{External benchmarks}} \\
    Defeasible \&     & BoardgameQA      & 92.8 & 93.9 & $+1.1$ [0.8, 1.5]    & \\
    non-monotonic     & LogicNMR         & 94.1 & 95.5 & $+1.4$ [1.0, 1.8]    & \\
                      & $\delta$-NLI     & 82.4 & 84.4 & $+2.0$ [1.5, 2.5]    & \\
                      & DEFREASING       & 70.0 & 70.2 & $+0.2$ [$-0.4$, 0.8] & \\
                      & Belief-R         & 49.9 & 49.6 & $-0.3$ [$-0.9$, 0.3] & \\
    Logical reasoning & LogicBench       & 84.4 & 84.2 & $-0.2$ [$-1.0$, 0.7] & \\
                      & Multi-LogiEval   & 75.6 & 74.8 & $-0.8$ [$-1.8$, 0.3] & \\
    Mathematics       & GSM8K            & 95.7 & 96.2 & $+0.5$ [$-0.2$, 1.2] & \\
                      & MATH-500         & 98.0 & 98.6 & $+0.6$ [$-0.1$, 1.3] & $+0.5$ [$-0.1$, 1.1] \\
    Knowledge \&      & MMLU-Pro         & 79.7 & 80.2 & $+0.5$ [0.1, 0.9]    & \\
    science           & GPQA Diamond     & 74.4 & 79.8 & $+5.4$ [3.2, 7.8]    & $+1.9$ [$-0.5$, 4.6] \\
    Code              & LiveCodeBench    & 73.2 & 78.5 & $+5.3$ [3.5, 7.0]    & $+2.1$ [0.2, 3.9] \\
    Broad reasoning   & BBEH             & 48.3 & 50.9 & $+2.6$ [1.4, 3.7]    & $+1.1$ [$-0.4$, 2.8] \\
    \bottomrule
  \end{tabular}
\end{table}

\paragraph{Training more than doubles complete-task success.}
On the held-out set, the trained model solves 53.3\% of instances, against 20.4\% for the base model (Table~\ref{tab:training}). The improvement is not limited to what the model practised: instances from task--level combinations never sampled in training improve by 29.6 points, instances from levels 13--15, beyond every training level, by 32.2 points, and the gain is about the same in early and late curriculum levels (+35.6 points in levels 1--5, +31.2 in levels 11--15). Most of it comes from constructive intervention, where success rises from 24.4\% to 80.0\%; on \texttt{attack}, \texttt{defence} and \texttt{attack\_defense}, the trained model now solves about nine instances in ten (Appendix~\ref{app:training}). Analysis improves much less (+10.2 points), mainly on status query, formalization and perturbation, three of the four tasks that received partial credit during training, and claim-chain success does not change; step~200 improves on eleven of the twelve tasks with intervals excluding zero. Training also leaves the effect of the link principle in place: when the same status-query theories are re-posed under the other principle, the trained model is still less accurate under weakest-link, in both directions (Appendix~\ref{app:training}). Finally, verifier rewards cannot teach what the model never gets right: weakest-link preference construction, whose instances require a preference between premises (Section~\ref{sec:results}), was rewarded in 1 of 2,480 sampled answers during training, and no evaluated answer solves it before or after training.

\paragraph{Transfer is selective.}
Training on symbolic theories carries over to the natural-language benchmarks closest to ArgGYM in form, whose rules, exceptions and priorities are stated explicitly. BoardgameQA \citep{kazemi2023boardgameqa} and LogicNMR \citep{xiu2022logicnmr} improve by 1.1 and 1.4 points from a base model that already scores above 92\%, and on BoardgameQA the gain is largest on the deepest problems, $+4.4$ points at depth~3 against $-1.2$ at depth~1. $\delta$-NLI \citep{rudinger2020skeptic} improves by 2.0 points, partly because the trained model more often produces a readable answer. DEFREASING \citep{allaway2025defreasing} and Belief-R \citep{wilie2024belief} do not change; on Belief-R, both models almost never revise an earlier conclusion when new evidence calls for it, and training does not change that. On logical reasoning, LogicBench \citep{parmar2024logicbench} and Multi-LogiEval \citep{patel2024multilogieval} are unchanged overall, but the trained model loses ground on LogicBench's 440 non-monotonic yes/no questions ($-2.0$ points), with the largest declines on one type of reasoning about exceptions ($-6.7$) and on reasoning about priority ($-6.3$; Appendix~\ref{app:external}).

\paragraph{General reasoning is preserved.}
Training on ArgGYM does not cost the model its general abilities: GSM8K \citep{cobbe2021gsm8k}, MATH-500 \citep{lightman2024lets} and MMLU-Pro \citep{wang2024mmlupro} change by at most 0.6 points. On the three benchmarks with the longest answers, GPQA Diamond \citep{rein2024gpqa}, LiveCodeBench \citep{jain2025livecodebench} and BBEH \citep{kazemi2025bbeh}, the trained model gains 5.4, 5.3 and 2.6 points. Much of this comes from finishing within the token limit more often: the share of truncated answers falls from 6.6\% to 1.5\%, from 13.5\% to 6.6\% and from 11.6\% to 7.1\% (run~1). On answers both models completed, LiveCodeBench still improves by 2.1 points, and GPQA Diamond and BBEH by 1.9 and 1.1, with intervals that include zero.

\section{Discussion}

Taken together, the findings suggest separable bottlenecks in current structured defeasible reasoning. Models with similar aggregate scores can fail for different reasons, and even the strongest evaluated systems retain operation-specific weaknesses. Understanding which argumentative state should result is distinct from finding an effective sequence of operations that reaches that state: the matched counter-argument conditions show that changing the intervention contract changes success in both directions rather than uniformly making the task easier. Training makes the separation visible. Verifier rewards improve constructive intervention far more than state analysis (Section~\ref{sec:training}), so improvement on one reasoning operation should not be assumed to imply improvement on another.

ArgGYM makes these distinctions testable within one executable formal environment. Because the tasks share the same ASPIC+ substrate, failures can be localized to support recovery, defeat diagnosis, semantic evaluation, preference reasoning, revision, or constructive intervention without changing the underlying argumentation formalism, while the procedural generators produce fresh verified instances targeting a particular operation or structural regime and the frozen benchmark remains unchanged. The training experiment illustrates the value of this decomposition: a 32.9-point gain on ArgGYM coexists with changes between $-0.8$ and $+5.4$ points on external benchmarks, a difference that a single aggregate score would hide. Further discussion, including the role of preference regimes and constructive search, is given in Appendix~\ref{app:discussion}.

\section{Limitations}

We acknowledge that our evaluation does not cover the full range of currently available models. Major proprietary systems, including Anthropic Claude and OpenAI's proprietary GPT-series models, are not included, and evaluating them with ArgGYM would be an important direction for future work.

Secondly, almost all ArgGYM tasks use symbolic argumentation theories to isolate the reasoning problem and enable exact verification. This leaves open how well the same capabilities transfer to more natural language. The training experiment offers partial evidence: gains from training on symbolic theories transfer to some natural-language defeasible benchmarks, but not to all of them (Section~\ref{sec:training}). ArgGYM also focuses on a single structured-argumentation framework, and most of its tasks use grounded semantics.

Our training experiment covers one model, one training run, and one reward design. We did not ablate the adaptive curriculum and therefore make no claim about its contribution. Training cells were also partly selected where the base model had a learning signal, which limits how cleanly the untrained-cell results separate generalization from difficulty.

Finally, the curriculum and ordering analyses should be interpreted as diagnostic. Curriculum configurations can vary several structural properties at once, while last-link and weakest-link instances are generated independently rather than as matched versions of the same underlying theory. We discuss each of these limitations further in Appendix~\ref{app:limitations}.

\section{Conclusion}

We introduced ArgGYM, a procedural, engine-verified environment for studying
structured defeasible reasoning in language models. Built on a common ASPIC+
substrate, ArgGYM decomposes defeasible reasoning into twelve analysis and
constructive tasks spanning representation, support recovery, defeat
diagnosis, semantic evaluation, revision, preference reasoning, and
intervention. A frozen 1,440-instance benchmark provides a reproducible basis
for model comparison, while the same task generators, formal semantics, and
verification contracts can produce fresh instances without modifying the
evaluation suite. This combination turns structured argumentation into both an
evaluation setting and a source of mechanically verifiable training problems.

Our results show why this decomposition matters. Defeasible reasoning in the tested models is not well described by a single aggregate capability: models can recover large parts of the required argumentative state without solving the complete task, become less reliable as structures interact, respond differently to preference regimes, and vary in their ability to exploit available intervention routes. These findings identify accurate state analysis and constructive search as distinct challenges that final-answer accuracy alone would obscure. Training with ArgGYM's verifiers as rewards more than doubles a 4B model's complete-task success on held-out instances, with gains concentrated in constructive intervention and selective transfer to external benchmarks. By combining a frozen benchmark with fresh, mechanically verified instances, ArgGYM provides a controlled basis for evaluating these weaknesses and for training models toward more reliable reasoning under incomplete, conflicting, and revisable information.

\FloatBarrier
\subsection*{AI Use Statement}

Generative AI tools were used during the preparation of this work to assist with drafting, revising, and polishing manuscript text; to support literature discovery by identifying potentially relevant prior work; and to assist with implementing, debugging, and correcting the procedural generation and evaluation code. The research questions, design of the procedural engine and its tasks, and interpretation of the results were carried out by the authors. All AI-assisted text, literature references, and code changes were reviewed and verified by the authors.

\FloatBarrier
\subsection*{Reproducibility Statement}

We release the frozen 1,440-instance benchmark, procedural generators,
task-specific scorers and verifiers, evaluation harness, and model
configuration files used in this study. The released code supports both
rescoring predictions on the frozen benchmark and generating new instances
under the same task definitions, curriculum configurations, preference
orderings, and verification contracts.

Appendix~\ref{app:formal} documents the formal backend, generation procedure,
argumentation configuration, and executable verification process.
Appendix~\ref{app:model-configs} reports the model identifiers and inference
configurations used in the evaluation. Evaluation metadata records the frozen
taskset identity and relevant prompt and scoring versions, allowing reported
runs to be checked against the intended benchmark and scoring protocol.

For the training experiment, Appendix~\ref{app:training} reports the training settings of the reported checkpoint, the selection of training cells, the learning-rate schedule, compute, and the checkpoint-selection record, and Appendix~\ref{app:external} reports the protocol, test sets, and scoring rules of the thirteen external benchmarks.

The frozen benchmark is kept separate from procedurally generated training
instances. This separation permits reproduction of the reported evaluation
while also allowing independent generation of fresh verified instances without
modifying or training on the frozen evaluation suite. The held-out set used to evaluate the trained model is disjoint from both the frozen benchmark and the training pool.

\section*{Acknowledgements}

This work was supported by the Scientific and Technological Research Council of Türkiye (TÜBİTAK) under Project No. 124C525. The numerical computations reported in this work were partially performed at TÜBİTAK ULAKBİM, High Performance and Grid Computing Center (TRUBA resources). We also acknowledge Google for providing research credits that supported the use of Gemini models in this study.

\bibliography{references}

\appendix

\section{Benchmark Details}
\subsection{Formal Engine and Argumentation Configuration}
\label{app:formal}

\paragraph{PyArg backend.}
ArgGYM uses PyArg as its pinned formal backend for ASPIC+ computation. PyArg is an open-source Python package for computational argumentation with support for abstract argumentation, ASPIC+ argumentation theories, and assumption-based argumentation, together with algorithms for extension-based semantics \citep{odekerken2023pyarg}. The package is distributed as \texttt{python-argumentation}; ArgGYM pins version 2.0.2 for benchmark generation and verification.

ArgGYM does not use PyArg's generic random generators to define the benchmark distribution. Instead, each task has its own procedural generator and task-specific acceptance conditions. Given a task, curriculum configuration, argument preference ordering, set ordering, and random seed, the generator constructs candidate instances from a deterministic pseudorandom process. Structural and semantic gates reject candidates that do not exhibit the phenomenon required by the task and configuration. For an accepted candidate, the ArgGYM adapter instantiates the axioms and ordinary premises, strict and defeasible rules, conflict relations, and preferences, after which the pinned PyArg backend computes the corresponding arguments, attacks, defeats, and semantic outcomes. The benchmark represents ordinary literal conflict through signed negation and undercutting through attacks on named defeasible rules; asymmetric contrary relations are not procedurally generated. The pinned PyArg backend therefore serves as the formal-semantic oracle for ArgGYM: it computes the argumentative states and consequences against which task-specific outputs are computed or validated, while ArgGYM's task-specific generators and scorers define the structured prediction and intervention contracts used by the benchmark.

\paragraph{Argument preferences.}
The benchmark crosses two argument preference orderings with two set orderings. Under the ASPIC+ \emph{last-link} principle, arguments are compared using their last defeasible rules, with ordinary-premise comparison applying when both arguments are strict; under \emph{weakest-link}, relevant ordinary premises and defeasible rules throughout the arguments may contribute to the comparison \citep{modgil2014aspic}. Preference information from an earlier defeasible step can therefore be irrelevant under last-link but decisive under weakest-link, allowing the two principles to induce different defeat relations and acceptance outcomes. Each argument preference ordering is combined with either the \emph{elitist} or \emph{democratic} set ordering, yielding
\[
\{\text{last-link},\text{weakest-link}\}
\times
\{\text{elitist},\text{democratic}\}.
\]
The selected configuration is stated explicitly whenever argument preferences can affect the solution.

\paragraph{Semantics.}
All ArgGYM tasks except \texttt{semantics\_query} use grounded semantics for acceptance. Grounded semantics has a unique extension, so each theory induces a single canonical acceptance state rather than requiring a choice among multiple extensions or acceptance modes \citep{dung1995acceptability,baroni2011semantics}. At the level of abstract argumentation, acceptance under grounded semantics is also polynomial-time decidable \citep{charwat2015survey}. These properties make grounded semantics a natural default for exact and repeated verification: each theory has a deterministic acceptance target that can be computed efficiently.

The \texttt{semantics\_query} task deliberately broadens this setting. It probes grounded, eager, and stable semantics, as well as credulous and skeptical acceptance under preferred semantics. Stable cases include theories with no stable extension, requiring models to distinguish the absence of a stable extension from the non-acceptance of a queried claim.

\paragraph{Executable verification.}
ArgGYM separates the evaluated model from the formal engine. During instance generation and freezing, the engine computes or validates the formal state required by each task, and the state needed by the corresponding scorer is stored in the frozen row. Predictions can therefore be rescored independently without regenerating the instance or relying on a language-model-produced reference answer.

Verification follows the contract of each task. The non-constructive tasks are evaluated against structured targets that are either computed directly by the formal engine or constructed during generation and subsequently validated against engine-computed states and task-specific structural conditions. For \texttt{formalization}, the submitted ASPIC+ directives are additionally
executed. The explicitly queried claim statuses contribute to the task-native behavioral score, while complete-task success requires behavioral equivalence with the intended theory over every literal named by the reference theory. Exact reconstruction of the reference directives is not required. For the six constructive tasks, the submitted directives are parsed, checked for task-specific legality, applied to the base theory, and evaluated by the formal engine. A construction succeeds when all requested target statuses are obtained and the resulting theory remains consistent. Correctness therefore depends on the formal object produced or induced by the answer.

\paragraph{Frozen-gold re-execution audit.}
As an additional safeguard against serialization, scoring, or ordering-propagation errors, we separately re-executed the complete 1,440-instance frozen benchmark from its stored formal state using ArgGYM and PyArg 2.0.2 while explicitly supplying each instance's recorded argument and set ordering. The audit covered all twelve tasks and all four last-link/weakest-link $\times$ democratic/elitist configurations. Every frozen reference answer or verified witness reproduced complete-task success under its corresponding task contract. We observed zero task-decoding failures, zero reference rescoring failures, zero ordering-propagation failures at either the task-payload or formal-framework level, and zero engine-invariant failures. This re-execution audit establishes internal consistency of the frozen benchmark, its serialized formal states, task-specific verification paths, and ordering propagation; it is not an independent second implementation of the underlying ASPIC+ semantics, which the cross-check below provides.

\paragraph{Independent cross-check.}
To validate the backend itself, we re-evaluated every theory in the released taskset with Arg2P \citep{calegari2022arg2p}, a separately implemented ASPIC+ engine, pinned at version 0.25.1, under the ordering configuration recorded for each instance. The comparison covers 11,400 theory evaluations, comprising every generated theory together with the theories obtained after perturbations and after applying reference constructions. Arg2P agrees with ArgGYM's stored states on every grounded claim status and on all 600 \texttt{semantics\_query} answers, and an independent implementation of Dung's semantics, written from the definitions, recomputes every grounded status from the exported argument and defeat graphs with no mismatch. The two engines differ in one convention: PyArg~2.0.2 identifies a rule by its content, so two differently named rules with the same antecedents and consequent are treated as one, whereas Arg2P keeps them apart. This changes the intermediate argument and defeat sets of 380 theories but none of their statuses, so no gold label is affected. The difference can affect scoring when duplicate rules are present in the base theory or introduced by a submitted answer. Across 21 archived runs, we screened 17,640 answers and independently re-scored the 625 flagged for duplicate-rule risk. Three verdicts changed, all from failure to success. In the audited runs, re-scoring under Arg2P changed no reported overall score by more than 0.1 percentage points. All reported scores use the pinned PyArg backend.

\paragraph{Consistency and strict-rule convention.}
For constructive tasks, consistency is evaluated on the recomputed argumentative
state: a result is treated as inconsistent if a literal and its signed contrary
are both justified. The presence of conflicting ordinary premises in the theory
does not by itself constitute failure unless it induces such jointly justified
conclusions. Strict rules, including model-added strict rules, are evaluated as
explicitly declared under the pinned PyArg semantics; ArgGYM does not
automatically close the submitted rule set under transposition or
contraposition.

\paragraph{Minimality and intervention efficiency.}
For the six constructive tasks, ArgGYM records a task-specific intervention minimum to encourage concise solutions. During generation, each task constructs a candidate space of admissible interventions and establishes the smallest number of directives that achieves the required semantic outcome within that space. Depending on the task and instance, this is established through exact subset or combination search or through task-specific decomposition and lower-bound or necessity checks. Instances for which the generator reports that minimality was not established are excluded during freezing.

The resulting \texttt{min\_directives} value is therefore a verified minimum within the generator-defined candidate space for that instance. It is not a claim of global minimality over every syntactically legal intervention. A model may submit a different legal intervention, including one not used to establish the recorded minimum, because its answer is evaluated by execution.

For a successful intervention, let $m$ denote the recorded candidate-space minimum and $n$ the number of submitted directives. Its native score is
\[
0.5 + 0.5\min\left(1,\frac{m}{n}\right).
\]
A successful intervention with $n \leq m$ receives full native credit, while a successful intervention with $n > m$ receives lower credit. If $n > 2m$, the submission is classified as a failed construction before engine execution. Thus semantic success determines correctness within the permitted intervention size, while the native score distinguishes more economical successful constructions.

\FloatBarrier
\subsection{Procedural Generation and Curriculum}
\label{app:generation}

Within the pinned benchmark configuration, generation is deterministic given
\[ 
(\tau,\ell,p,c,s), 
\]
where $\tau$ is the task, $\ell$ the curriculum configuration, $p$ the argument preference ordering, $c$ the set ordering, and $s$ the random seed. In the implementation, the pair $(p,c)$ is represented by a single combined ordering configuration $o=(p,c)$, corresponding to one of the four last-link/weakest-link $\times$ elitist/democratic combinations. These coordinates initialize a task-specific deterministic pseudorandom process. A generator may construct and reject several internal candidates before producing the instance associated with a seed. Candidates are retained only when task-specific structural predicates and engine-computed outcomes confirm that the intended reasoning phenomenon is present.

When the frozen benchmark is constructed, ArgGYM considers each task--level--ordering cell separately and scans seeds until the required number of accepted instances is obtained, subject to a fixed search bound. The accepted seeds and generator rejection statistics are recorded. Thus randomness provides diversity within a controlled task distribution, while deterministic seeding, explicit rejection criteria, and version pinning make the resulting benchmark reproducible.

The fifteen curriculum configurations manipulate task-relevant structural features rather than a single global complexity variable. Depending on the task, these include derivation depth, branching, numbers and interactions of attackers and defenders, mixtures of strict and defeasible inference, distractor material, preference-sensitive conflicts, junctions between support chains, and interactions among multiple target claims. The curriculum is intended to expose progressively richer reasoning structures rather than merely increase input length. Accordingly, level number should be interpreted as curriculum progression and not as a guarantee of monotonic empirical difficulty.

\FloatBarrier
\subsection{Task Definitions}
\label{app:task-definitions}

ArgGYM decomposes structured defeasible reasoning into twelve benchmark tasks. The tasks differ in both the formal operation required from the model and the form in which correctness can be verified. Six tasks produce structured analyses or formalizations, while six require executable interventions whose consequences are recomputed by the formal engine. The descriptions below specify the reasoning problem associated with each task; detailed scoring and verification procedures are given in Appendix~\ref{app:scoring}.

\paragraph{\texttt{formalization}.}
The model receives a controlled natural-language description of an argumentation theory together with the statuses that a correct formalization must induce on a set of queried claims. The description is procedurally realized from pools of sentence templates that express the underlying argumentation-theoretic distinctions, including ordinary premises, axioms, strict and defeasible inference, conflicts, and preferences, using both technical and paraphrastic formulations. The model must translate this description into executable ASPIC+ directives containing the appropriate premises or axioms, rules, and preference information. The submitted theory is then executed. Its agreement on the explicitly queried claims contributes to the task-native behavioral score, while complete-task success requires the submitted theory to reproduce the reference theory's engine-derived statuses over every literal named by the reference theory. Thus a behaviorally equivalent formalization may succeed without exactly reconstructing the generator's reference directives. The task-native score additionally measures structural and type-sensitive agreement with that reference theory.

\paragraph{\texttt{claim\_chain}.}
The model is given a theory and a justified target claim and must recover the complete support derivation responsible for deriving that claim. The required support structure may branch at higher curriculum configurations when a rule depends on multiple supporting lines. The answer therefore consists of the premises and inference rules in the target's generated support derivation, ordered from supporting premises toward the target. Because independent branches need not have a unique textual serialization, the scorer does not require the particular order in which the generator constructed the derivation. Instead, it accepts any ordering consistent with the derivational dependencies: each rule must appear only after the premises or intermediate conclusions on which it depends. Complete-task success requires recovery of all and only the required support directives together with a dependency-consistent ordering.

\paragraph{\texttt{defeat\_diagnosis}.}
The model is given a claim that is not justified and must determine its status and identify every relevant point at which its support fails. A diagnosis records the attacked component, the defeater, and whether the failure is due to undermining, rebutting, or undercutting; selected instances additionally require identifying cases in which an apparent attack on the defeater is itself defeated. The task therefore requires explaining the dialectical structure responsible for non-acceptance.

\paragraph{\texttt{status\_query}.}
Given an ASPIC+ theory and a set of queried literals, the model assigns each
literal one of \textsc{justified}, \textsc{overruled}, or
\textsc{undecided} under grounded semantics. Complete-task success requires the
status of every queried literal to match the engine-derived state. Generator
constraints ensure that all three statuses occur among the queried claims and
limit label imbalance, reducing the usefulness of majority-label shortcuts.

\paragraph{\texttt{semantics\_query}.}
This task extends claim-status reasoning beyond the benchmark's default
grounded setting. The model must determine claim acceptance under the semantics
and reasoning mode stated in the instance. The task covers grounded and eager
semantics, stable semantics, and credulous or skeptical acceptance under
preferred semantics; stable instances may also contain no stable extension.
The model must therefore distinguish changes in semantic interpretation from
changes in the underlying argumentation theory.

\paragraph{\texttt{perturbation}.}
The model is given an argumentation theory together with a specified update and must identify which claims of the original theory change status after the update, as well as the new status of each changed claim. The task therefore requires both detecting the affected claims and computing the downstream consequences of the intervention across the argumentative structure. During generation, the engine computes claim statuses before and after the update. The gold answer consists of the claims whose statuses differ between these two states together with their post-update statuses. Generated instances also retain claims whose statuses survive the perturbation, providing unchanged structure against which spurious change predictions can be distinguished.

\paragraph{\texttt{preference\_construction}.}
The model must add preference statements that transform the current theory into
a requested configuration of claim statuses. No new premises or inference rules
may be introduced: the intervention is restricted to legal preferences over
ordinary premises and defeasible rules. Submitted preferences are executed
together and the resulting argumentative state is recomputed, allowing the task
to test inverse reasoning about how preference relations alter defeat and
acceptance. Successful solutions must satisfy all requested statuses while
preserving consistency.

\paragraph{\texttt{counter\_argument}.}
The model starts from a theory in which a target claim is justified and must construct an intervention that makes the opposing claim justified and the original claim overruled. The permitted intervention language includes new defeasible rules, preferences over ordinary premises or defeasible rules, and the introduction of the contrary of an existing ordinary premise for undermining. Arbitrary new positive premises and new axioms are not permitted, and this task variant does not permit the addition of strict rules. The submitted directives are executed against the original theory, so correctness is determined by the argumentative state they induce.

\paragraph{\texttt{counter\_argument\_strict}.}
This task forms a matched comparison with the \texttt{counter\_argument} task. For each matched configuration, both present the same underlying argumentation theory and require the same semantic outcome: the opposing claim must become justified and the original claim overruled. The conditions differ in the permitted intervention language and, because the strict variant has a lower recorded minimum, in the largest accepted answer (twice that minimum). \texttt{counter\_argument\_strict} permits strict-rule additions, whereas \texttt{counter\_argument} removes that operation while retaining the remaining legal intervention types. Their comparison therefore measures performance under two intervention contracts on the same reasoning problem.

The paired construction is designed so that permitting strict inference provides
a cheaper verified route without changing the theory being solved. Across all
120 frozen matched pairs, the recorded candidate-space minimum is lower in the
strict-rule-enabled condition: it is lower by one directive in 112 pairs and by
two directives in the remaining eight. In the ordinary condition, the recorded
minima are 3 directives for 40 instances, 4 for 48, 5 for 24, and 6 for 8;
in the strict condition, they are 2 directives for 48 instances, 3 for 40,
4 for 24, and 5 for 8. The contrast therefore tests model behavior when a
cheaper verified intervention route becomes available. As for the other
constructive tasks, success is determined by the recomputed argumentative state
and consistency, not by reproducing the stored reference intervention.

\paragraph{\texttt{attack}.}
The model receives a theory in which a designated target is justified and must
modify the theory so that the target becomes overruled. A successful answer must therefore construct an effective attack and ensure that it survives the
resulting preference and defeat relations strongly enough to change the target's
semantic status. The submitted intervention is checked for task-specific
legality and evaluated by executing it against the theory.

\paragraph{\texttt{defence}.}
The model receives an attacked target and must construct an intervention that
makes the target justified. This may require countering or disabling effective
attackers, strengthening relevant support, or otherwise altering the
argumentative configuration within the operations permitted by the task.
Success is determined from the recomputed semantic state rather than from
matching a particular defensive strategy.

\paragraph{\texttt{attack\_defense}.}
The model must satisfy two semantic objectives within a single intervention:
one designated target must become overruled while another becomes justified.
The task therefore requires coordinating offensive and defensive modifications
whose consequences can interact within the same argumentation theory. All requested postconditions must hold simultaneously and the resulting theory
must remain consistent.

\section{Task Verification}
\label{app:scoring}

Each task returns two quantities: a task-native score in $[0,1]$ and a Boolean complete-task success indicator. Complete-task success is the primary quantity used for comparisons across tasks. Native scores retain task-specific partial-credit information and are therefore interpreted only within each task.

\paragraph{Pair-valued prediction tasks.}
The \texttt{status\_query}, \texttt{semantics\_query}, and
\texttt{perturbation} tasks are scored as key--status maps. For a predicted map
$P$ and gold map $G$, a true positive is a key whose submitted status matches
the corresponding gold status exactly. Precision and recall are
\[
    \mathrm{Prec}=\frac{\mathrm{TP}}{|P|},
    \qquad
    \mathrm{Rec}=\frac{\mathrm{TP}}{|G|},
\]
with the usual harmonic-mean $F_1$. A key assigned multiple contradictory
statuses does not count as correct. The native score is this key--status pair
$F_1$, and complete-task success requires exact equality of the predicted and
gold maps. For \texttt{status\_query}, keys are queried claims; for
\texttt{semantics\_query}, keys are claim--semantics pairs; and for
\texttt{perturbation}, keys are claims whose status changes after the specified
update.

\paragraph{\texttt{claim\_chain}.}
Let $G$ be the set of directives in the target support derivation and let the
submitted answer contain $n$ directive lines. Content precision and recall are
computed from the overlap between submitted and required directives, with
repeated submitted directives still counting against precision. Their harmonic
mean gives $F_1^{\mathrm{content}}$. The scorer additionally computes
\[
    q =
    \frac{\text{number of submitted rules appearing after all of their
    antecedents}}
         {\text{number of submitted rules}},
\]
taking $q=1$ when no rule is submitted. The native score is
\[
    F_1^{\mathrm{content}} q.
\]
Complete-task success requires all and only the required support directives and
a dependency-consistent ordering; it does not require the particular ordering
used internally by the generator.

\paragraph{\texttt{defeat\_diagnosis}.}
The scorer compares predicted failure points with the generated diagnosis using
the pair
\[
    (\texttt{defeated\_at},\texttt{defeater})
\]
as the record key and requires the attack type
(\texttt{undermine}, \texttt{undercut}, or \texttt{rebut}) to match for that
record to count as correct. Let $F_1^{\mathrm{diag}}$ denote the resulting
failure-point $F_1$, and let $I_{\mathrm{status}}$ indicate whether the
submitted status of the target claim is correct. When no
\texttt{survives\_because} explanation is required, the native score is
\[
    0.85 F_1^{\mathrm{diag}} + 0.15 I_{\mathrm{status}}.
\]
For instances requiring higher-order survival explanations, let
$F_1^{\mathrm{surv}}$ denote $F_1$ over the required
\texttt{survives\_because} records. The native score becomes
\[
    0.60 F_1^{\mathrm{diag}}
    + 0.15 I_{\mathrm{status}}
    + 0.25 F_1^{\mathrm{surv}}.
\]
Complete-task success requires the target status, all failure points and their attack types, and all required survival explanations to match exactly. If an instance requires no \texttt{survives\_because} explanation but the submitted answer supplies one, the submission receives native score $0$ and cannot achieve complete-task success.

\paragraph{\texttt{formalization}.}
A submitted formalization is parsed as an executable ASPIC+ theory and run under the ordering specified by the instance. Let $B$ be the pair-level $F_1$ between the statuses induced by the submitted theory and the specified statuses of the explicitly queried claims. Structural agreement is measured separately by an alpha-renaming-invariant multiset comparison of the submitted and reference directives; let its $F_1$ be $S$.

When the reference theory contains type-sensitive axiom or strict-rule decisions, let $T$ denote the corresponding type-decision $F_1$. Its recall is computed against the axiom and strict-rule decisions in the reference theory, while its precision also penalizes additional axiom or strict-rule directives introduced by the submitted theory. The native score is
\[
    0.4B + 0.6S
\]
when no such type-sensitive decisions are present, and
\[
    0.25B + 0.35S + 0.40T
\]
otherwise.

Complete-task success is determined by a broader behavioral-equivalence check. The scorer collects every atom named by the reference theory through its premises, axioms, and rule antecedents or consequents, evaluates both polarities of those atoms, and requires the submitted theory to induce the same engine-computed status as the reference theory for every such literal. The queried claims must therefore also be correct, but matching only the queried statuses is insufficient. Exact reconstruction of the reference directive multiset, rule names, or rule grouping is not required. 

\paragraph{Constructive intervention tasks.}
The six intervention tasks---\texttt{preference\_construction},
\texttt{counter\_argument}, \texttt{counter\_argument\_strict},
\texttt{attack}, \texttt{defence}, and \texttt{attack\_defense}---share an
execution-based scorer. Submitted directives are parsed and checked against
task-specific legality constraints. The retained intervention is applied to
the base theory, and the argumentative state is recomputed under the
instance's ordering and semantics.

Complete-task success requires every requested postcondition to hold
simultaneously and the resulting theory to remain consistent. Let $m$ be the
recorded candidate-space minimum and $n$ the number of submitted readable
directives. Successful interventions with a known minimum receive
\[
    0.5 + 0.5\min\left(1,\frac{m}{n}\right).
\]
Thus a successful intervention using at most $m$ directives receives native score $1$, while longer successful constructions receive decreasing efficiency credit. A submission using more than $2m$ directives is classified as failed before execution. When a construction does not achieve complete success, partial credit is
computed from semantic progress relative to the argumentative state before the
intervention. Consistent unsuccessful constructions receive at most $0.25$.
For each requested goal that was not already satisfied in the base theory, the
progress value is $1$ if the intervention achieves the requested status and
$0.4$ if it moves the goal to \textsc{undecided} from a different initial
status when either \textsc{justified} or \textsc{overruled} was requested.
Thus, leaving a goal unchanged at \textsc{undecided} does not receive deadlock
credit, and a goal that was already satisfied before the intervention does not
receive free positive credit. A goal that is already satisfied in the base theory is excluded from the progress average if it remains satisfied, but contributes zero if the intervention breaks it.

For an \textsc{overruled} goal with derived support subgoals, an otherwise
unsatisfied goal may instead receive $0.3u$, where $u$ is the fraction of
relevant subgoals whose support has been defeated by the intervention. A
subgoal is included in this calculation only if it was initially
\textsc{justified} or its status was changed by the submitted intervention,
preventing unchanged irrelevant subgoals from contributing credit. Progress is
averaged across the applicable requested goals, and the native score of a
consistent unsuccessful construction is $0.25$ times this average.
Inconsistent resulting theories receive no partial credit.

Every readable submitted directive counts toward $n$, including duplicates and
directives subsequently rejected by legality checks. Unparseable responses are
assigned zero. The recorded minimum is a verified minimum within the
generator-defined candidate space for the instance, not a claim of global
minimality over every syntactically legal intervention.

\paragraph{Scoring interpretation.}
Because the native scores encode different notions of partial correctness
across tasks, they are not averaged into a benchmark-wide native score.
Benchmark-wide performance is summarized using complete-task success; native
scores are used only for within-task diagnostics and comparisons.

\section{Representative Benchmark Instances}
\label{app:examples}

The examples below are drawn from frozen benchmark instances and are presented in abridged form for readability. Omitted material consists of generated theory lines or detailed parser-level legality instructions; the evaluated models received the complete original prompts. Constructive-task reference answers are verified witnesses rather than strings that the model must reproduce.

\subsection{Grounded Claim-Status}
\label{ex:status-query}

The first example illustrates the basic acceptance-query interface. Although
relatively small, the instance contains strict and defeasible inference,
conflicting arguments, and premise preferences.

\begin{center}
\begin{tabular}{ll}
\toprule
Task & \texttt{status\_query} \\
Level & 2 \\
Ordering & last-link democratic \\
Semantics & grounded \\
Output type & claim-status map \\
\bottomrule
\end{tabular}
\end{center}

\begin{small}
\begin{verbatim}
The following is a defeasible argumentation theory, evaluated under
grounded semantics with the last-link democratic strength ordering.

[premise: ko9]
[axiom: mp0]
[premise: -lp3]
[premise: kk6]
[premise: lp3]

... additional generated premises and rules omitted here ...

[defeasible vo8: ko9 => ls2]
[defeasible mi1: ls2 => ik2]
[defeasible hy5: lq8 => dv8]
[defeasible hi2: gn5 => -ls2]
[defeasible ve8: mk7 => lq8]

[prefer_premise: -mk7 > mk7]
[prefer_premise: -lp3 > lp3]

State the status of each of the following claims:
ls2, mo0, ik2, dv8, mk7, ho7.

Possible statuses: justified, overruled, undecided.
A claim is justified when some argument for it is accepted,
overruled when every argument for it is defeated, and undecided
otherwise.

Answer format: one line per claim, written as `claim: status`.
\end{verbatim}
\end{small}

The engine-derived answer is

\begin{small}
\begin{verbatim}
ls2: undecided
mo0: justified
ik2: undecided
dv8: overruled
mk7: overruled
ho7: justified
\end{verbatim}
\end{small}

\subsection{Defeat Diagnosis}
\label{ex:diagnosis}

In this question, the model must identify \emph{where} each possible support route for a claim fails, \emph{how} it is attacked, and, where required, why the defeater remains effective.

\begin{center}
\begin{tabular}{ll}
\toprule
Task & \texttt{defeat\_diagnosis} \\
Level & 5 \\
Ordering & last-link democratic \\
Semantics & grounded \\
Output type & structured diagnostic record \\
Target claim & \texttt{im1} \\
\bottomrule
\end{tabular}
\end{center}

A presentation excerpt of the generated theory is:

\begin{small}
\begin{verbatim}
[premise: al9]
[premise: fl4]
[premise: -fl4]
[premise: br3]
[premise: fv8]

... additional generated premises and rules omitted here ...

[defeasible ta7: iq0 => im1]
[defeasible so1: bw7 => lp5]
[defeasible fu0: -ik2 => bw7]
[defeasible ve3: mo9 => bv5]
[defeasible ga3: bv5 => cs7]

[prefer_premise: -fl4 > fl4]
[prefer_premise: -ik2 > ik2]
[prefer_rule: vo7 > ga3]

The claim im1 is not justified.
State its status, and identify every point at which its support fails.

For defeated_at:
  - use the attacked ordinary premise for an undermine,
  - the attacked defeasible rule name for an undercut,
  - the attacked conclusion literal for a rebut.

For defeater:
  - use the attacking premise literal for an undermine,
  - otherwise use the defeasible rule whose conclusion performs the attack.

Append
  survives_because: <rule>
only when the listed defeater is itself attacked but that attack is defeated.

Answer format:
   first line: status: overruled or status: undecided
   then one line per failure point:
   defeated_at: <target>; defeater: <defeater>;
   kind: undermine|undercut|rebut
\end{verbatim}
\end{small}

The generator-constructed, engine-validated reference diagnosis is

\begin{small}
\begin{verbatim}
status: overruled
defeated_at: fl4; defeater: -fl4; kind: undermine
defeated_at: so1; defeater: ze6; kind: undercut;
             survives_because: si5
defeated_at: cs7; defeater: vo7; kind: rebut;
             survives_because: pu0
\end{verbatim}
\end{small}

This example exposes information that a single claim-status label would hide.
The model must distinguish undermining, undercutting, and rebutting, locate the
attacked component on each failed support route, and represent a higher-order
case in which an apparent counterattack does not neutralize the operative
defeater.

\FloatBarrier
\subsection{Strict Counterargument Example}
\label{ex:counter-strict}

This instance illustrates the strict-rule-enabled condition of the matched counter-argument comparison. Its paired \texttt{counter\_argument} instance uses the same base theory and semantic objective; the strict variant additionally permits strict-rule additions and has a lower recorded minimum. The model must therefore construct a small intervention that changes the argumentative state while having access to an intervention form unavailable in the paired non-strict arm. 

\begin{center}
\begin{tabular}{ll}
\toprule
Task & \texttt{counter\_argument\_strict} \\
Level & 6 \\
Ordering & weakest-link elitist \\
Semantics & grounded \\
Output type & executable directive list \\
Recorded minimum & 2 directives \\
\bottomrule
\end{tabular}
\end{center}

\begin{small}
\begin{verbatim}
The following is a defeasible argumentation theory, evaluated under
grounded semantics with the weakest-link elitist strength ordering.

[premise: gv9]
[premise: hq8]
[premise: jr5]
[premise: -kp2]
[premise: -ay2]
[axiom: bv4]

... additional generated premises and rules omitted here ...

[defeasible fy9: by3 AND hu0 => lq4]
[defeasible za8: -kp2 => gu6]
[strict hu2: lq4 -> au0]
[defeasible ji8: bt8 => by3]
[defeasible su9: km2 AND gq4 => by7]
[strict xi0: bv4 -> kt8]
[defeasible ha8: cm8 => -ju8]

[prefer_premise: -kp2 > kp2]
[prefer_rule: mi6 > ha8]

The claim au0 is currently justified.
What is the minimal set of directives that makes -au0 justified
and au0 overruled?

Permitted additions, written exactly in these forms:
   [defeasible <name>: <antecedent> => <consequent>]
   [strict <name>: <antecedent> -> <consequent>]
   [prefer_rule: <defeasible rule name> > <defeasible rule name>]
   [prefer_premise: <ordinary premise literal> >
                    <ordinary premise literal>]
   [premise: -<literal>]

New axioms may not be added.

... additional parser-level syntax and legality constraints omitted here ...

The answer must be minimal: one using more than twice the fewest
directives that work scores zero. A directive that cannot be read
at all scores the whole answer zero. An answer that reaches every
goal while leaving the theory inconsistent also scores zero.

Answer format: one directive per line.
\end{verbatim}
\end{small}

One verified reference intervention is

\begin{small}
\begin{verbatim}
[strict cs: gv9 -> -au0]
[strict u1: gv9 -> -ji8]
\end{verbatim}
\end{small}

The scorer does not require these particular rule names or this particular
construction. It parses the submitted directives, applies them to the base
theory, recomputes the argumentative state, and checks whether
\texttt{-au0} is justified and \texttt{au0} is overruled while the resulting
theory remains consistent. The instance records a two-directive minimum within
the generator's verified candidate space.

\section{Detailed Evaluation Results}

\subsection{Model and Inference Configurations}
\label{app:model-configs}

Table~\ref{tab:model-configs} reports the inference configurations used in the evaluation. We evaluate 18 distinct model variants under 42 model--reasoning configurations. Reasoning configurations for the same model share decoding parameters unless otherwise indicated. Inference settings are fixed within each model--reasoning configuration, while output limits, sampling parameters, context windows, numerical precision, and serving configurations may differ across systems. For locally served models, context denotes the \texttt{max\_model\_len} configured in the corresponding vLLM serving profile. A dash indicates a parameter that is not explicitly configured by the evaluation harness.

\begin{table*}[t]
\centering
\scriptsize
\setlength{\tabcolsep}{4pt}
\caption{\textbf{Model and inference configurations.} $T$ denotes temperature, $p$ top-$p$, $k$ top-$k$, and PP presence penalty. Max.\ out.\ is the maximum generated-token budget.}
\label{tab:model-configs}
\begin{tabular}{@{}llllrr@{}}
\toprule
Model & Reasoning & Sampling & Seed & Max.\ out. & Context \\
\midrule
\multicolumn{6}{l}{\textit{Gemma}} \\

\texttt{google/gemma-4-E2B-it}
& off / on
& $T=1.0,\ p=.95,\ k=64$
& 0 & 65536 & 131072 \\

\texttt{google/gemma-4-E4B-it}
& off / on
& $T=1.0,\ p=.95,\ k=64$
& 0 & 65536 & 131072 \\

\texttt{google/gemma-4-26B-A4B-it}
& off / on
& $T=1.0,\ p=.95,\ k=64$
& 0 & 65536 & 262144 \\

\texttt{google/gemma-4-31B-it}
& off / on
& $T=1.0,\ p=.95,\ k=64$
& 0 & 65536 & 262144 \\

\midrule
\multicolumn{6}{l}{\textit{GPT-OSS}} \\

\texttt{openai/gpt-oss-20b}
& low / medium / high
& $T=1.0,\ p=1.0$
& 0 & 65536 & 131072 \\

\texttt{openai/gpt-oss-120b}
& low / medium / high
& $T=1.0,\ p=1.0$
& 0 & 65536 & 131072 \\

\midrule
\multicolumn{6}{l}{\textit{Qwen}} \\

\texttt{Qwen/Qwen3.5-0.8B}
& off
& $T=1.0,\ p=1.0,\ k=20,\ \mathrm{PP}=2.0$
& 0 & 65536 & 262144 \\

&
& \multicolumn{1}{l}{} & & & \\[-8pt]

& on
& $T=1.0,\ p=.95,\ k=20,\ \mathrm{PP}=1.5$
& 0 & 65536 & 262144 \\

\texttt{Qwen/Qwen3.5-2B}
& off
& $T=1.0,\ p=1.0,\ k=20,\ \mathrm{PP}=2.0$
& 0 & 65536 & 262144 \\

& on
& $T=1.0,\ p=.95,\ k=20,\ \mathrm{PP}=1.5$
& 0 & 65536 & 262144 \\

\texttt{Qwen/Qwen3.5-4B}
& off / on
& $T=1.0,\ p=.95,\ k=20,\ \mathrm{PP}=1.5$
& 0 & 65536 & 262144 \\

\texttt{Qwen/Qwen3.5-9B}
& off / on
& $T=1.0,\ p=.95,\ k=20,\ \mathrm{PP}=1.5$
& 0 & 65536 & 262144 \\

\texttt{Qwen/Qwen3.5-35B-A3B}
& off / on
& $T=1.0,\ p=.95,\ k=20,\ \mathrm{PP}=1.5$
& 0 & 65536 & 262144 \\

\texttt{Qwen/Qwen3.5-27B}
& off / on
& $T=1.0,\ p=.95,\ k=20,\ \mathrm{PP}=1.5$
& 0 & 65536 & 262144 \\

\texttt{Qwen/Qwen3.8-27B}
& low / medium / xhigh
& $T=1.0,\ p=.95,\ k=20,\ \mathrm{PP}=0$
& 0 & 65536 & 131072 \\

\midrule
\multicolumn{6}{l}{\textit{DeepSeek}} \\

\texttt{deepseek-v4.1-flash}
& high
& $T=1.0,\ p=.95$
& -- & 65536 & -- \\

\midrule
\multicolumn{6}{l}{\textit{Gemini}} \\

\texttt{gemini-2.5-pro}
& low / medium / high
& provider defaults
& -- & 65536 & -- \\

\texttt{gemini-3.5-flash-lite}
& low / medium / high
& provider defaults
& -- & 65536 & -- \\

\texttt{gemini-3.1-pro-preview}
& low / medium / high
& provider defaults
& -- & 65536 & -- \\

\texttt{gemini-3.8-flash}
& low / medium / high
& provider defaults
& -- & 65536 & -- \\

\bottomrule
\end{tabular}
\end{table*}

For Gemma and Qwen3.5, thinking is controlled through the model chat template. GPT-OSS and Gemini 2.5 Pro/3.5 Flash Lite receive explicit reasoning-effort settings. Gemini 3.1 Pro Preview and Gemini 3.8 Flash use the provider's \texttt{thinking\_level} control. Qwen3.8-27B uses explicit \texttt{reasoning\_effort} values of \texttt{low}, \texttt{medium}, and \texttt{xhigh}. DeepSeek-v4.1-Flash is evaluated with \texttt{reasoning\_effort=high} through its chat-template configuration.

The Qwen configurations additionally set \texttt{min\_p=0.0} and \texttt{repetition\_penalty=1.0}; these neutral settings are omitted from the table for readability.

\FloatBarrier
\subsection{Task-Level Performance}
\label{app:task-results}

Here we provide the corresponding task-level results for all twelve ArgGYM tasks. Each cell reports
\emph{complete-task success (\%)} followed by the \emph{task-native score} multiplied by 100. Native scores retain task-specific semantics and should therefore be compared only within the same task.

\begin{table*}[t]
\centering
\tiny
\setlength{\tabcolsep}{2.5pt}
\caption{\textbf{Analysis and evaluation tasks.} Each cell reports complete-task success (\%) $\,|\, $ task-native score $\times 100$.}
\label{tab:task-results-analysis}
\resizebox{\textwidth}{!}{
\begin{tabular}{@{}llrrrrrr@{}}
\toprule
Model & Reasoning & Formal. & Chain & Diagnosis & Status & Semantics & Perturb. \\
\midrule

\multicolumn{8}{l}{\textit{Gemma}} \\
\texttt{google/gemma-4-E2B-it}
& off & 0.8$|$1.0 & 2.5$|$3.5 & 5.0$|$13.1 & 1.7$|$47.6 & 0.0$|$38.0 & 0.0$|$3.7 \\
& on & 0.8$|$1.5 & 5.0$|$5.3 & 0.0$|$10.0 & 4.2$|$52.1 & 0.0$|$42.2 & 0.0$|$4.8 \\

\texttt{google/gemma-4-E4B-it}
& off & 4.2$|$16.8 & 2.5$|$3.3 & 5.8$|$23.5 & 11.7$|$69.4 & 2.5$|$58.6 & 0.8$|$12.7 \\
& on & 4.2$|$25.1 & 6.7$|$7.9 & 9.2$|$26.1 & 10.8$|$71.4 & 0.0$|$56.8 & 0.8$|$16.2 \\

\texttt{google/gemma-4-26B-A4B-it}
& off & 7.5$|$43.2 & 2.5$|$3.3 & 9.2$|$35.5 & 15.0$|$75.1 & 4.2$|$64.5 & 1.7$|$22.8 \\
& on & 10.8$|$44.8 & 15.8$|$18.8 & 10.8$|$53.3 & 10.8$|$72.9 & 10.0$|$66.3 & 2.5$|$36.8 \\

\texttt{google/gemma-4-31B-it}
& off & 14.2$|$64.9 & 24.2$|$24.6 & 7.5$|$42.2 & 6.7$|$74.9 & 10.0$|$71.6 & 3.3$|$46.1 \\
& on & 50.8$|$78.9 & 30.8$|$31.6 & 25.8$|$69.8 & 20.0$|$77.4 & 22.5$|$81.8 & 6.7$|$58.3 \\

\midrule
\multicolumn{8}{l}{\textit{GPT-OSS}} \\
\texttt{openai/gpt-oss-20b}
& low & 5.8$|$14.4 & 9.2$|$12.7 & 8.3$|$19.0 & 4.2$|$41.4 & 0.8$|$39.2 & 0.0$|$2.0 \\
& medium & 5.8$|$50.1 & 31.7$|$35.0 & 14.2$|$38.9 & 3.3$|$65.7 & 5.0$|$57.1 & 0.0$|$14.9 \\
& high & 10.8$|$66.1 & 40.0$|$42.0 & 16.7$|$48.1 & 5.0$|$63.6 & 1.7$|$57.6 & 0.0$|$14.5 \\

\texttt{openai/gpt-oss-120b}
& low & 7.5$|$33.5 & 27.5$|$29.2 & 7.5$|$27.0 & 5.8$|$67.5 & 3.3$|$46.7 & 0.0$|$10.3 \\
& medium & 9.2$|$60.0 & 49.2$|$50.2 & 10.8$|$47.0 & 15.0$|$77.3 & 10.8$|$67.1 & 0.0$|$24.5 \\
& high & 14.2$|$75.2 & 45.0$|$45.9 & 20.0$|$64.0 & 15.8$|$78.1 & 13.3$|$70.6 & 5.0$|$34.5 \\

\midrule
\multicolumn{8}{l}{\textit{Qwen}} \\
\texttt{Qwen/Qwen3.5-0.8B}
& off & 0.0$|$0.2 & 0.0$|$0.0 & 0.0$|$3.9 & 0.0$|$12.0 & 0.0$|$4.8 & 0.0$|$0.9 \\
& on & 0.0$|$0.3 & 0.0$|$0.0 & 0.0$|$0.0 & 0.0$|$0.0 & 0.0$|$0.0 & 0.0$|$0.0 \\

\texttt{Qwen/Qwen3.5-2B}
& off & 1.7$|$1.9 & 0.8$|$1.2 & 0.8$|$5.8 & 0.0$|$24.1 & 0.0$|$23.0 & 0.0$|$1.0 \\
& on & 0.8$|$1.8 & 2.5$|$3.5 & 5.8$|$8.7 & 0.8$|$6.6 & 0.0$|$15.2 & 0.0$|$0.2 \\

\texttt{Qwen/Qwen3.5-4B}
& off & 5.8$|$26.8 & 31.7$|$37.8 & 12.5$|$41.9 & 12.5$|$79.6 & 11.7$|$63.6 & 1.7$|$23.8 \\
& on & 7.5$|$26.2 & 50.0$|$52.5 & 20.0$|$53.1 & 20.8$|$75.8 & 8.3$|$65.5 & 2.5$|$24.8 \\

\texttt{Qwen/Qwen3.5-9B}
& off & 8.3$|$36.0 & 38.3$|$42.1 & 16.7$|$55.2 & 23.3$|$86.5 & 7.5$|$66.6 & 5.8$|$41.9 \\
& on & 10.0$|$37.6 & 42.5$|$47.4 & 22.5$|$64.8 & 20.0$|$81.3 & 13.3$|$73.3 & 3.3$|$47.2 \\

\texttt{Qwen/Qwen3.5-35B-A3B}
& off & 15.8$|$43.3 & 36.7$|$41.9 & 20.0$|$65.2 & 29.2$|$89.6 & 30.0$|$83.5 & 7.5$|$54.5 \\
& on & 12.5$|$43.9 & 44.2$|$48.2 & 31.7$|$76.6 & 39.2$|$91.9 & 24.2$|$83.0 & 15.8$|$59.8 \\

\texttt{Qwen/Qwen3.5-27B}
& off & 18.3$|$58.5 & 45.8$|$48.0 & 20.0$|$72.5 & 25.8$|$86.2 & 21.7$|$80.0 & 7.5$|$63.8 \\
& on & 22.5$|$64.1 & 50.0$|$60.1 & 39.2$|$86.6 & 35.8$|$88.7 & 22.5$|$82.2 & 15.0$|$71.8 \\

\texttt{Qwen/Qwen3.8-27B}
& low & 35.0$|$75.9 & 38.3$|$38.9 & 27.5$|$74.6 & 5.8$|$69.2 & 5.8$|$59.4 & 7.5$|$43.1 \\
& medium & 30.8$|$77.1 & 46.7$|$49.7 & 35.0$|$80.1 & 6.7$|$71.9 & 9.2$|$68.6 & 1.7$|$52.8 \\
& xhigh & 34.2$|$86.0 & 61.7$|$63.2 & 50.0$|$92.3 & 44.2$|$91.4 & 18.3$|$77.3 & 24.2$|$72.0 \\

\midrule
\multicolumn{8}{l}{\textit{DeepSeek}} \\
\texttt{deepseek-v4.1-flash}
& high & 67.5$|$91.5 & 60.0$|$60.1 & 50.8$|$88.2 & 34.2$|$89.7 & 15.8$|$72.2 & 46.7$|$83.0 \\

\midrule
\multicolumn{8}{l}{\textit{Gemini}} \\
\texttt{gemini-2.5-pro}
& low & 41.7$|$58.2 & 24.2$|$31.6 & 14.2$|$35.8 & 6.7$|$71.1 & 1.7$|$57.4 & 0.8$|$21.6 \\
& medium & 44.2$|$68.6 & 49.2$|$55.8 & 35.8$|$88.7 & 35.0$|$92.4 & 26.7$|$78.8 & 25.0$|$87.7 \\
& high & 40.0$|$65.4 & 49.2$|$56.9 & 38.3$|$88.4 & 37.5$|$93.2 & 21.7$|$78.3 & 22.5$|$88.8 \\

\texttt{gemini-3.5-flash-lite}
& low & 21.7$|$46.8 & 32.5$|$33.2 & 16.7$|$34.6 & 16.7$|$66.2 & 10.8$|$65.4 & 0.8$|$25.4 \\
& medium & 58.3$|$85.1 & 38.3$|$39.3 & 22.5$|$73.6 & 23.3$|$83.7 & 14.2$|$73.2 & 3.3$|$41.1 \\
& high & 84.2$|$87.0 & 50.0$|$50.5 & 35.0$|$84.8 & 25.8$|$89.3 & 19.2$|$79.0 & 10.0$|$68.9 \\

\texttt{gemini-3.1-pro-preview}
& low & 95.8$|$89.6 & 52.5$|$69.8 & 51.7$|$92.1 & 23.3$|$45.2 & 29.2$|$84.9 & 32.5$|$80.2 \\
& medium & 99.2$|$89.9 & 57.5$|$85.3 & 56.7$|$93.3 & 20.8$|$24.6 & 56.7$|$92.6 & 50.0$|$92.6 \\
& high & 97.5$|$90.6 & 58.3$|$92.3 & 60.8$|$94.1 & 32.5$|$61.4 & 76.7$|$96.3 & 64.2$|$95.4 \\

\texttt{gemini-3.8-flash}
& low & 97.5$|$90.8 & 58.3$|$62.8 & 57.5$|$93.2 & 55.0$|$95.4 & 67.5$|$96.0 & 53.3$|$80.3 \\
& medium & 98.3$|$94.3 & 88.3$|$94.1 & 57.5$|$93.6 & 63.3$|$97.2 & 92.5$|$98.6 & 84.2$|$97.3 \\
& high & 99.2$|$97.5 & 85.8$|$94.3 & 57.5$|$93.6 & 61.7$|$97.1 & 98.3$|$99.8 & 86.7$|$98.3 \\

\bottomrule
\end{tabular}}
\end{table*}

\begin{table*}[t]
\centering
\tiny
\setlength{\tabcolsep}{2.5pt}
\caption{\textbf{Constructive intervention tasks.} Each cell reports complete-task success (\%) $\,|\, $ task-native score $\times 100$.}
\label{tab:task-results-construction}
\resizebox{\textwidth}{!}{
\begin{tabular}{@{}llrrrrrr@{}}
\toprule
Model & Reasoning & Pref. & Counter & Counter+S & Attack & Defence & Attack--Def. \\
\midrule

\multicolumn{8}{l}{\textit{Gemma}} \\
\texttt{google/gemma-4-E2B-it}
& off & 2.5$|$2.7 & 0.0$|$0.0 & 0.0$|$0.1 & 0.0$|$0.9 & 0.8$|$0.8 & 0.0$|$0.6 \\
& on & 3.3$|$3.5 & 0.0$|$0.2 & 0.0$|$0.2 & 0.0$|$1.2 & 2.5$|$2.5 & 0.0$|$1.0 \\

\texttt{google/gemma-4-E4B-it}
& off & 12.5$|$14.7 & 0.0$|$2.1 & 0.0$|$1.8 & 0.0$|$3.9 & 2.5$|$4.5 & 0.0$|$2.4 \\
& on & 8.3$|$11.6 & 0.0$|$2.3 & 0.0$|$3.6 & 0.0$|$4.2 & 4.2$|$6.8 & 0.8$|$5.0 \\

\texttt{google/gemma-4-26B-A4B-it}
& off & 30.8$|$33.4 & 2.5$|$4.9 & 4.2$|$6.8 & 0.0$|$4.3 & 20.8$|$22.1 & 10.0$|$15.2 \\
& on & 35.0$|$36.7 & 8.3$|$12.9 & 15.8$|$18.1 & 9.2$|$14.0 & 22.5$|$25.8 & 17.5$|$22.7 \\

\texttt{google/gemma-4-31B-it}
& off & 40.8$|$42.5 & 5.0$|$8.6 & 15.0$|$16.3 & 8.3$|$10.5 & 28.3$|$31.8 & 9.2$|$15.9 \\
& on & 46.7$|$47.0 & 14.2$|$17.8 & 16.7$|$18.7 & 20.0$|$21.1 & 39.2$|$41.9 & 34.2$|$37.8 \\

\midrule
\multicolumn{8}{l}{\textit{GPT-OSS}} \\
\texttt{openai/gpt-oss-20b}
& low & 8.3$|$8.7 & 3.3$|$4.3 & 2.5$|$2.7 & 4.2$|$5.0 & 15.8$|$15.7 & 12.5$|$17.6 \\
& medium & 18.3$|$21.1 & 19.2$|$19.6 & 11.7$|$11.4 & 15.8$|$15.9 & 38.3$|$38.7 & 29.2$|$32.8 \\
& high & 23.3$|$25.7 & 28.3$|$26.9 & 24.2$|$22.6 & 18.3$|$18.9 & 40.0$|$41.0 & 46.7$|$49.2 \\

\texttt{openai/gpt-oss-120b}
& low & 13.3$|$16.0 & 10.8$|$11.2 & 6.7$|$6.2 & 12.5$|$12.3 & 30.0$|$30.0 & 23.3$|$23.5 \\
& medium & 34.2$|$36.5 & 15.8$|$15.6 & 13.3$|$13.0 & 13.3$|$12.9 & 41.7$|$42.3 & 45.0$|$43.7 \\
& high & 56.7$|$57.5 & 30.8$|$29.8 & 19.2$|$18.1 & 13.3$|$14.0 & 60.0$|$59.2 & 54.2$|$54.0 \\

\midrule
\multicolumn{8}{l}{\textit{Qwen}} \\
\texttt{Qwen/Qwen3.5-0.8B}
& off & 0.0$|$0.0 & 0.0$|$0.0 & 0.0$|$0.0 & 0.0$|$0.0 & 0.0$|$0.0 & 0.0$|$0.1 \\
& on & 0.0$|$0.0 & 0.0$|$0.0 & 0.0$|$0.0 & 0.0$|$0.0 & 0.0$|$0.0 & 0.0$|$0.0 \\

\texttt{Qwen/Qwen3.5-2B}
& off & 0.8$|$0.8 & 0.0$|$0.0 & 0.0$|$0.0 & 0.0$|$0.2 & 0.0$|$0.0 & 0.0$|$0.3 \\
& on & 1.7$|$1.9 & 0.0$|$0.1 & 0.0$|$0.1 & 0.0$|$0.1 & 0.8$|$0.8 & 0.0$|$0.0 \\

\texttt{Qwen/Qwen3.5-4B}
& off & 31.7$|$34.1 & 3.3$|$6.2 & 5.8$|$7.5 & 9.2$|$12.2 & 17.5$|$20.0 & 5.8$|$12.0 \\
& on & 43.3$|$44.3 & 21.7$|$21.8 & 21.7$|$21.4 & 17.5$|$19.0 & 30.0$|$31.8 & 25.0$|$27.4 \\

\texttt{Qwen/Qwen3.5-9B}
& off & 40.8$|$41.8 & 10.8$|$14.9 & 9.2$|$13.3 & 5.8$|$10.2 & 25.0$|$27.3 & 22.5$|$25.9 \\
& on & 43.3$|$43.7 & 20.8$|$23.6 & 44.2$|$43.6 & 24.2$|$25.5 & 32.5$|$34.9 & 36.7$|$38.3 \\

\texttt{Qwen/Qwen3.5-35B-A3B}
& off & 40.0$|$41.6 & 10.8$|$12.7 & 15.8$|$16.6 & 5.8$|$8.8 & 26.7$|$29.1 & 21.7$|$26.2 \\
& on & 46.7$|$47.1 & 15.8$|$19.9 & 42.5$|$42.8 & 30.8$|$32.4 & 37.5$|$39.6 & 39.2$|$40.9 \\

\texttt{Qwen/Qwen3.5-27B}
& off & 47.5$|$47.7 & 23.3$|$25.4 & 31.7$|$29.2 & 37.5$|$38.2 & 46.7$|$48.3 & 51.7$|$52.8 \\
& on & 49.2$|$49.3 & 41.7$|$44.2 & 42.5$|$41.9 & 44.2$|$46.7 & 51.7$|$54.0 & 57.5$|$58.4 \\

\texttt{Qwen/Qwen3.8-27B}
& low & 33.3$|$33.5 & 25.8$|$29.3 & 50.8$|$49.4 & 21.7$|$25.2 & 45.8$|$48.3 & 43.3$|$44.4 \\
& medium & 39.2$|$39.5 & 27.5$|$31.4 & 40.8$|$41.5 & 16.7$|$20.5 & 57.5$|$59.5 & 41.7$|$43.6 \\
& xhigh & 50.0$|$49.0 & 42.5$|$43.7 & 11.7$|$11.7 & 27.5$|$24.9 & 45.8$|$48.3 & 60.0$|$57.9 \\

\midrule
\multicolumn{8}{l}{\textit{DeepSeek}} \\
\texttt{deepseek-v4.1-flash}
& high & 60.0$|$59.9 & 34.2$|$35.5 & 13.3$|$13.5 & 32.5$|$32.1 & 54.2$|$54.4 & 70.8$|$67.6 \\

\midrule
\multicolumn{8}{l}{\textit{Gemini}} \\
\texttt{gemini-2.5-pro}
& low & 45.0$|$45.6 & 3.3$|$4.6 & 5.0$|$4.6 & 15.0$|$17.9 & 29.2$|$30.1 & 12.5$|$17.6 \\
& medium & 53.3$|$53.5 & 23.3$|$26.2 & 15.8$|$15.9 & 15.0$|$16.8 & 40.0$|$42.2 & 50.8$|$51.3 \\
& high & 52.5$|$52.6 & 25.8$|$28.0 & 4.2$|$5.0 & 20.0$|$21.8 & 41.7$|$43.8 & 45.0$|$46.4 \\

\texttt{gemini-3.5-flash-lite}
& low & 20.0$|$22.9 & 9.2$|$11.2 & 4.2$|$4.9 & 11.7$|$12.9 & 45.0$|$45.4 & 29.2$|$33.9 \\
& medium & 41.7$|$42.7 & 22.5$|$25.0 & 10.8$|$10.7 & 18.3$|$20.7 & 60.8$|$61.6 & 37.5$|$43.4 \\
& high & 44.2$|$45.5 & 30.0$|$31.4 & 15.0$|$13.8 & 13.3$|$14.8 & 53.3$|$54.2 & 41.7$|$48.3 \\

\texttt{gemini-3.1-pro-preview}
& low & 49.2$|$49.2 & 46.7$|$47.6 & 18.3$|$16.4 & 41.7$|$41.6 & 70.0$|$70.7 & 69.2$|$70.8 \\
& medium & 52.5$|$52.4 & 56.7$|$58.1 & 20.8$|$19.7 & 35.8$|$35.6 & 65.8$|$66.8 & 86.7$|$86.4 \\
& high & 61.7$|$61.7 & 57.5$|$58.8 & 23.3$|$22.5 & 54.2$|$54.3 & 76.7$|$77.7 & 91.7$|$92.0 \\

\texttt{gemini-3.8-flash}
& low & 51.7$|$51.9 & 51.7$|$51.9 & 30.0$|$27.2 & 71.7$|$71.5 & 83.3$|$83.4 & 60.0$|$61.7 \\
& medium & 57.5$|$57.5 & 76.7$|$77.3 & 31.7$|$30.5 & 98.3$|$98.3 & 90.8$|$90.8 & 94.2$|$94.6 \\
& high & 63.3$|$63.3 & 86.7$|$87.3 & 60.0$|$58.2 & 99.2$|$99.2 & 92.5$|$92.5 & 97.5$|$97.7 \\

\bottomrule
\end{tabular}}
\end{table*}

\FloatBarrier
\subsection{Failure and Structure Analysis of Untrained Models}
\label{app:failure-analysis}

This section analyzes why untrained models fail, using the generations of 20 configurations: Gemma~4 E4B (thinking on and off), GPT-OSS-20B and GPT-OSS-120B (each at three effort levels), Qwen3.5-4B, 9B, 27B, and 35B-A3B (thinking on and off), Qwen3.8-27B (three effort levels), and DeepSeek-v4.1-Flash. Re-scoring these generations reproduces the task-level results of Table~\ref{tab:task-results-analysis} exactly for 18 configurations; the Gemma~4 E4B generations are a different sample of the same two configurations. All analyses are descriptive and use the frozen benchmark. They address three questions: whether failures reflect wrong reasoning or answer format, whether low complete success merely compounds independent per-claim errors, and whether curriculum level matters once the structure of a theory is controlled.

\paragraph{Failure classes.} Each failed answer to an analysis task is classified in order: truncated generation, no answer region, correct content in a wrong layout (\emph{format only}), or unreadable lines. Among the remaining answers, a wrong stated judgment takes priority; an answer whose stated judgments are all correct counts as \emph{incomplete} if required items are missing, or \emph{order only} if only their order is wrong. Table~\ref{tab:failure-classes} pools the 11,671 failed analysis answers. Wrong judgments account for 82.6\% of failures, between 74.2\% and 95.1\% in every configuration; only 6 answers (0.1\%) fail on layout alone, and at most 0.3\% in any configuration. Most failures are therefore errors of judgment; the 10.4\% that are unreadable, lack an answer region, or are truncated are not adjudicated.

\begin{table}[h]
\centering
\caption{\textbf{Why analysis-task answers fail}, by task, pooled over the 20 untrained configurations (counts of failed answers; last row: share of all failures).}
\label{tab:failure-classes}
\scriptsize
\setlength{\tabcolsep}{3pt}
\begin{tabular}{@{}lrrrrrrrr@{}}
\toprule
Task & Failed & Truncated & No region & Format only & Unreadable & Incomplete & Order only & Wrong judgment \\
\midrule
\texttt{claim\_chain}      & 1,491 & 8  & 136 & 0 & 103 & 22  & 3 & 1,219 \\
\texttt{defeat\_diagnosis} & 1,878 & 15 & 28  & 0 & 1   & 416 & 0 & 1,418 \\
\texttt{formalization}     & 2,005 & 8  & 74  & 6 & 631 & 0   & 0 & 1,286 \\
\texttt{perturbation}      & 2,223 & 35 & 39  & 0 & 35  & 363 & 0 & 1,751 \\
\texttt{semantics\_query}  & 2,127 & 3  & 39  & 0 & 5   & 5   & 0 & 2,075 \\
\texttt{status\_query}     & 1,947 & 17 & 32  & 0 & 7   & 1   & 0 & 1,890 \\
\midrule
Total                      & 11,671 & 86 & 348 & 6 & 782 & 807 & 3 & 9,639 \\
Share (\%)                 & 100   & 0.7 & 3.0 & 0.1 & 6.7 & 6.9 & 0.0 & 82.6 \\
\bottomrule
\end{tabular}
\end{table}

\paragraph{Are claim errors independent?} If a model judged each queried claim of a \texttt{status\_query} instance independently with its observed per-claim accuracy, its exact success would be the product of those accuracies. The prediction for an instance is the product, over its queried claims, of the configuration's accuracy in that band on claims with the same correct status. Observed exact success exceeds this independent-error prediction in 45 of the 47 configuration--band combinations in which a model solves any instance, by up to a factor of 12.5; in the remaining 13 combinations the model solves no instance, while the prediction is at most 1.6\%, so they do not discriminate. Errors therefore concentrate in particular instances; the test does not distinguish harder instances from dependence among the claims of one instance.

\paragraph{Which claims are judged wrongly?} A logistic regression of claim-level correctness on structural properties, with intervals clustered by instance, identifies where errors concentrate. Longer derivations lower per-claim accuracy in 19 of the 20 configurations, and claims whose correct status is \textsc{undecided} are judged worse in 16. Neither the number of queried claims nor curriculum level is associated with lower accuracy in any configuration once these properties are controlled: the interval for the number of claims includes zero in 19 configurations and is positive in one, and that for level includes zero in 17 and is positive in three.

\paragraph{Level versus structure.} For each task, we regress complete-task success on curriculum level alone and then on level together with the structural properties recorded by the generator and computed from the argument graph (for example derivation depth, numbers of attackers and junctions, numbers of arguments and defeats, and the argument and set orderings), pooling the 20 configurations. Table~\ref{tab:level-structure} reports the level coefficient. Level alone predicts lower success in 11 of the 12 tasks; once structure is controlled, its interval includes zero in 11 of the 12, the exception being \texttt{counter\_argument\_strict}. Conditional on these correlated structural features, the level index has no detectable association with success in 11 of the 12 tasks.

\begin{table}[h]
\centering
\caption{\textbf{Curriculum level with and without structural controls.} Log-odds per standard deviation of level, pooled over the 20 configurations (120 instances per task each); 95\% intervals resample instances.}
\label{tab:level-structure}
\small
\begin{tabular}{@{}lcc@{}}
\toprule
Task & Level only & Level with structure \\
\midrule
\texttt{attack} & $-0.56$ [$-0.73$, $-0.45$] & $-0.36$ [$-1.04$, 0.37] \\
\texttt{attack\_defense} & $-0.22$ [$-0.41$, $-0.03$] & $+0.34$ [$-0.31$, 0.93] \\
\texttt{claim\_chain} & $-1.22$ [$-1.35$, $-1.09$] & $-0.19$ [$-0.74$, 0.34] \\
\texttt{counter\_argument} & $-0.39$ [$-0.53$, $-0.27$] & $+0.06$ [$-0.33$, 0.50] \\
\texttt{counter\_argument\_strict} & $-0.54$ [$-0.67$, $-0.44$] & $-0.49$ [$-0.95$, $-0.08$] \\
\texttt{defeat\_diagnosis} & $-1.52$ [$-1.83$, $-1.22$] & $-0.55$ [$-1.35$, 0.07] \\
\texttt{defence} & $-0.75$ [$-0.98$, $-0.55$] & $+0.30$ [$-0.04$, 0.72] \\
\texttt{formalization} & $-0.75$ [$-1.14$, $-0.48$] & $+0.46$ [$-0.74$, 1.35] \\
\texttt{perturbation} & $-1.22$ [$-1.55$, $-0.93$] & $-0.03$ [$-0.68$, 0.53] \\
\texttt{preference\_construction} & $-0.26$ [$-0.60$, 0.04] & $+0.29$ [$-0.45$, 0.86] \\
\texttt{semantics\_query} & $-1.94$ [$-2.42$, $-1.53$] & $-0.18$ [$-0.69$, 0.35] \\
\texttt{status\_query} & $-1.37$ [$-1.68$, $-1.13$] & $-0.59$ [$-1.47$, 0.43] \\
\bottomrule
\end{tabular}
\end{table}

\FloatBarrier
\subsection{Curriculum Analysis}
\label{app:curriculum-analysis}

The fifteen curriculum configurations vary task-specific structural properties such as derivation depth, branching, interacting attackers and defenders, strict and defeasible structure, distractors, and multi-goal interactions, and, for \texttt{status\_query}, the number of queried literals. They are intended to provide progressively richer configurations, not a calibrated one-dimensional difficulty scale. We therefore summarize robustness in three five-configuration bands: L1--L5, L6--L10, and L11--L15. Each band contains 480 instances for a complete evaluation configuration.

Table~\ref{tab:curriculum-bands} reports complete-task success separately for every model--reasoning configuration. Across most configurations, success decreases in the later bands, while stronger models and higher reasoning settings generally retain more performance as structural richness increases.

\begin{table*}[t]
\centering
\scriptsize
\setlength{\tabcolsep}{5pt}
\caption{\textbf{Complete-task success across structural-richness bands.} Values are percentages.}
\label{tab:curriculum-bands}
\begin{tabular}{@{}llrrr@{}}
\toprule
Model & Reasoning & L1--L5 & L6--L10 & L11--L15 \\
\midrule

\multicolumn{5}{l}{\textit{Gemma}} \\
\texttt{google/gemma-4-E2B-it}
& off & 3.3 & 0.0 & 0.0 \\
& on & 4.0 & 0.0 & 0.0 \\

\texttt{google/gemma-4-E4B-it}
& off & 9.6 & 0.8 & 0.2 \\
& on & 10.2 & 0.6 & 0.4 \\

\texttt{google/gemma-4-26B-A4B-it}
& off & 17.5 & 5.6 & 4.0 \\
& on & 25.8 & 9.6 & 6.9 \\

\texttt{google/gemma-4-31B-it}
& off & 25.8 & 11.0 & 6.2 \\
& on & 42.3 & 22.9 & 16.7 \\

\midrule
\multicolumn{5}{l}{\textit{GPT-OSS}} \\
\texttt{openai/gpt-oss-20b}
& low & 16.5 & 1.2 & 1.0 \\
& medium & 27.7 & 14.2 & 6.2 \\
& high & 34.2 & 17.1 & 12.5 \\

\texttt{openai/gpt-oss-120b}
& low & 24.6 & 9.8 & 2.7 \\
& medium & 34.8 & 18.5 & 11.2 \\
& high & 45.2 & 25.6 & 16.0 \\

\midrule
\multicolumn{5}{l}{\textit{Qwen}} \\
\texttt{Qwen/Qwen3.5-0.8B}
& off & 0.0 & 0.0 & 0.0 \\
& on & 0.0 & 0.0 & 0.0 \\

\texttt{Qwen/Qwen3.5-2B}
& off & 1.0 & 0.0 & 0.0 \\
& on & 2.9 & 0.2 & 0.0 \\

\texttt{Qwen/Qwen3.5-4B}
& off & 26.5 & 7.5 & 3.3 \\
& on & 40.6 & 16.2 & 10.2 \\

\texttt{Qwen/Qwen3.5-9B}
& off & 34.0 & 11.9 & 7.7 \\
& on & 42.5 & 22.1 & 13.8 \\

\texttt{Qwen/Qwen3.5-35B-A3B}
& off & 43.3 & 14.2 & 7.5 \\
& on & 49.4 & 28.5 & 17.1 \\

\texttt{Qwen/Qwen3.5-27B}
& off & 52.7 & 25.0 & 16.7 \\
& on & 58.8 & 32.9 & 26.2 \\

\texttt{Qwen/Qwen3.8-27B}
& low & 40.4 & 26.0 & 18.8 \\
& medium & 37.5 & 27.1 & 23.8 \\
& xhigh & 50.0 & 38.8 & 28.7 \\

\midrule
\multicolumn{5}{l}{\textit{DeepSeek}} \\
\texttt{deepseek-v4.1-flash}
& high & 57.5 & 42.5 & 35.0 \\

\midrule
\multicolumn{5}{l}{\textit{Gemini}} \\
\texttt{gemini-2.5-pro}
& low & 27.9 & 14.4 & 7.5 \\
& medium & 48.5 & 31.9 & 23.1 \\
& high & 48.3 & 29.2 & 22.1 \\

\texttt{gemini-3.5-flash-lite}
& low & 38.1 & 11.9 & 4.6 \\
& medium & 47.5 & 21.9 & 18.5 \\
& high & 54.2 & 30.6 & 20.6 \\

\texttt{gemini-3.1-pro-preview}
& low & 62.7 & 42.9 & 39.4 \\
& medium & 69.8 & 51.9 & 43.1 \\
& high & 76.7 & 62.5 & 49.6 \\

\texttt{gemini-3.8-flash}
& low & 76.7 & 60.4 & 47.3 \\
& medium & 83.1 & 76.7 & 73.5 \\
& high & 88.1 & 82.5 & 76.5 \\

\bottomrule
\end{tabular}
\end{table*}

\subsubsection{Task-Specific Curriculum Profiles}
\label{app:curriculum-task-results}

To identify which reasoning operations are most sensitive to structural richness, Tables~\ref{tab:curriculum-early-analysis}--\ref{tab:curriculum-late-construction} report task-level performance separately for the three curriculum bands. Each task-specific band contains 40 frozen benchmark instances for a complete model--reasoning configuration. Each cell reports complete-task success (\%) followed by the task-native score multiplied by 100. Native scores retain task-specific semantics and should therefore be compared only within the same task.

\begin{table*}[t]
\centering
\tiny
\setlength{\tabcolsep}{2.4pt}
\caption{\textbf{Task-specific analysis/evaluation performance for curriculum configurations L1--L5.} Each cell reports complete-task success (\%) $\,|\, $ task-native score $\times 100$.}
\label{tab:curriculum-early-analysis}
\resizebox{\textwidth}{!}{
\begin{tabular}{@{}llrrrrrr@{}}
\toprule
Model & Reasoning & Formal. & Chain & Diagnosis & Status & Semantics & Perturb. \\
\midrule

\multicolumn{8}{l}{\textit{Gemma}} \\
\texttt{google/gemma-4-E2B-it}
& off & 2.5$|$3.1 & 7.5$|$8.4 & 15.0$|$26.2 & 5.0$|$56.0 & 0.0$|$40.2 & 0.0$|$8.2 \\
& on & 2.5$|$4.6 & 15.0$|$15.7 & 0.0$|$16.1 & 12.5$|$65.2 & 0.0$|$47.3 & 0.0$|$9.1 \\

\texttt{google/gemma-4-E4B-it}
& off & 12.5$|$32.0 & 5.0$|$5.0 & 17.5$|$50.7 & 35.0$|$81.5 & 7.5$|$71.0 & 2.5$|$28.0 \\
& on & 12.5$|$35.3 & 15.0$|$16.7 & 27.5$|$54.4 & 32.5$|$84.8 & 0.0$|$62.1 & 2.5$|$30.7 \\

\texttt{google/gemma-4-26B-A4B-it}
& off & 20.0$|$49.0 & 7.5$|$7.5 & 27.5$|$73.8 & 42.5$|$86.9 & 12.5$|$75.2 & 5.0$|$37.5 \\
& on & 17.5$|$50.4 & 42.5$|$48.8 & 30.0$|$85.7 & 27.5$|$80.5 & 30.0$|$82.1 & 7.5$|$45.8 \\

\texttt{google/gemma-4-31B-it}
& off & 27.5$|$66.0 & 45.0$|$45.0 & 22.5$|$68.0 & 15.0$|$78.8 & 30.0$|$82.8 & 10.0$|$54.5 \\
& on & 55.0$|$76.1 & 65.0$|$67.3 & 57.5$|$93.3 & 45.0$|$86.3 & 65.0$|$94.4 & 20.0$|$72.2 \\

\midrule
\multicolumn{8}{l}{\textit{GPT-OSS}} \\
\texttt{openai/gpt-oss-20b}
& low & 17.5$|$34.3 & 27.5$|$31.8 & 25.0$|$40.4 & 12.5$|$60.2 & 2.5$|$47.9 & 0.0$|$5.3 \\
& medium & 12.5$|$49.6 & 52.5$|$52.5 & 42.5$|$66.5 & 10.0$|$70.2 & 12.5$|$69.8 & 0.0$|$24.0 \\
& high & 25.0$|$75.1 & 60.0$|$60.0 & 50.0$|$81.6 & 15.0$|$70.4 & 5.0$|$64.1 & 0.0$|$21.2 \\

\texttt{openai/gpt-oss-120b}
& low & 17.5$|$54.4 & 52.5$|$53.8 & 22.5$|$53.6 & 17.5$|$80.0 & 10.0$|$60.2 & 0.0$|$18.1 \\
& medium & 20.0$|$60.7 & 77.5$|$77.5 & 32.5$|$76.3 & 45.0$|$83.8 & 32.5$|$80.6 & 0.0$|$38.4 \\
& high & 32.5$|$75.1 & 70.0$|$72.4 & 57.5$|$89.5 & 30.0$|$79.9 & 37.5$|$80.1 & 15.0$|$51.5 \\

\midrule
\multicolumn{8}{l}{\textit{Qwen}} \\
\texttt{Qwen/Qwen3.5-0.8B}
& off & 0.0$|$0.5 & 0.0$|$0.0 & 0.0$|$2.6 & 0.0$|$14.9 & 0.0$|$8.7 & 0.0$|$1.2 \\
& on & 0.0$|$1.0 & 0.0$|$0.0 & 0.0$|$0.0 & 0.0$|$0.0 & 0.0$|$0.0 & 0.0$|$0.0 \\

\texttt{Qwen/Qwen3.5-2B}
& off & 5.0$|$5.6 & 2.5$|$2.9 & 2.5$|$7.5 & 0.0$|$26.8 & 0.0$|$25.9 & 0.0$|$1.7 \\
& on & 2.5$|$2.5 & 7.5$|$10.6 & 17.5$|$23.7 & 2.5$|$18.8 & 0.0$|$24.5 & 0.0$|$0.6 \\

\texttt{Qwen/Qwen3.5-4B}
& off & 17.5$|$45.2 & 60.0$|$65.7 & 37.5$|$69.5 & 35.0$|$83.9 & 35.0$|$81.6 & 5.0$|$33.5 \\
& on & 17.5$|$49.0 & 87.5$|$87.5 & 57.5$|$86.8 & 50.0$|$87.9 & 22.5$|$75.9 & 7.5$|$38.5 \\

\texttt{Qwen/Qwen3.5-9B}
& off & 25.0$|$59.4 & 80.0$|$84.2 & 50.0$|$87.0 & 50.0$|$89.2 & 22.5$|$77.9 & 17.5$|$55.3 \\
& on & 17.5$|$50.9 & 77.5$|$79.8 & 55.0$|$88.4 & 42.5$|$87.8 & 40.0$|$87.1 & 10.0$|$61.3 \\

\texttt{Qwen/Qwen3.5-35B-A3B}
& off & 35.0$|$52.5 & 67.5$|$72.2 & 52.5$|$86.6 & 67.5$|$95.3 & 72.5$|$94.8 & 22.5$|$66.2 \\
& on & 27.5$|$64.1 & 87.5$|$91.6 & 65.0$|$92.3 & 65.0$|$95.0 & 65.0$|$89.0 & 45.0$|$81.6 \\

\texttt{Qwen/Qwen3.5-27B}
& off & 30.0$|$70.2 & 80.0$|$86.6 & 55.0$|$91.4 & 47.5$|$89.8 & 62.5$|$94.4 & 22.5$|$67.2 \\
& on & 27.5$|$70.1 & 85.0$|$87.3 & 67.5$|$94.8 & 72.5$|$95.1 & 62.5$|$91.9 & 42.5$|$79.7 \\

\texttt{Qwen/Qwen3.8-27B}
& low & 47.5$|$79.6 & 67.5$|$69.2 & 55.0$|$92.9 & 12.5$|$72.3 & 17.5$|$74.0 & 17.5$|$52.8 \\
& medium & 37.5$|$82.5 & 77.5$|$81.4 & 55.0$|$92.4 & 12.5$|$74.1 & 25.0$|$77.0 & 5.0$|$56.6 \\
& xhigh & 37.5$|$85.8 & 95.0$|$95.0 & 60.0$|$94.0 & 67.5$|$95.6 & 40.0$|$84.6 & 47.5$|$82.7 \\

\midrule
\multicolumn{8}{l}{\textit{DeepSeek}} \\
\texttt{deepseek-v4.1-flash}
& high & 85.0$|$91.3 & 90.0$|$90.0 & 60.0$|$92.7 & 52.5$|$90.4 & 37.5$|$81.8 & 50.0$|$82.8 \\

\midrule
\multicolumn{8}{l}{\textit{Gemini}} \\
\texttt{gemini-2.5-pro}
& low & 52.5$|$65.3 & 42.5$|$46.7 & 42.5$|$66.4 & 20.0$|$75.9 & 2.5$|$64.3 & 2.5$|$22.4 \\
& medium & 52.5$|$71.1 & 92.5$|$94.8 & 52.5$|$93.0 & 60.0$|$93.2 & 47.5$|$86.3 & 52.5$|$89.7 \\
& high & 52.5$|$72.0 & 77.5$|$79.8 & 67.5$|$95.4 & 75.0$|$97.1 & 50.0$|$85.2 & 47.5$|$87.9 \\

\texttt{gemini-3.5-flash-lite}
& low & 45.0$|$70.1 & 67.5$|$67.5 & 50.0$|$74.1 & 47.5$|$89.7 & 30.0$|$79.2 & 2.5$|$46.2 \\
& medium & 72.5$|$81.0 & 85.0$|$85.0 & 60.0$|$90.1 & 57.5$|$92.2 & 37.5$|$89.1 & 10.0$|$66.6 \\
& high & 90.0$|$84.8 & 90.0$|$90.0 & 62.5$|$92.7 & 50.0$|$90.5 & 52.5$|$92.7 & 27.5$|$79.0 \\

\texttt{gemini-3.1-pro-preview}
& low & 95.0$|$88.1 & 90.0$|$96.8 & 70.0$|$95.3 & 62.5$|$79.7 & 75.0$|$96.7 & 67.5$|$93.0 \\
& medium & 100.0$|$89.0 & 92.5$|$97.0 & 70.0$|$95.5 & 62.5$|$69.1 & 87.5$|$94.9 & 70.0$|$94.9 \\
& high & 100.0$|$89.0 & 97.5$|$99.4 & 75.0$|$96.3 & 60.0$|$79.8 & 95.0$|$99.2 & 75.0$|$95.8 \\

\texttt{gemini-3.8-flash}
& low & 100.0$|$90.0 & 97.5$|$97.5 & 72.5$|$95.9 & 75.0$|$97.0 & 95.0$|$99.4 & 72.5$|$93.3 \\
& medium & 100.0$|$92.1 & 100.0$|$100.0 & 72.5$|$95.9 & 77.5$|$97.4 & 97.5$|$99.5 & 77.5$|$96.1 \\
& high & 100.0$|$98.1 & 100.0$|$100.0 & 72.5$|$95.9 & 77.5$|$97.4 & 100.0$|$100.0 & 82.5$|$96.8 \\

\bottomrule
\end{tabular}}
\end{table*}

\begin{table*}[t]
\centering
\tiny
\setlength{\tabcolsep}{2.4pt}
\caption{\textbf{Task-specific constructive performance for curriculum configurations L1--L5.} Each cell reports complete-task success (\%) $\,|\, $ task-native score $\times 100$.}
\label{tab:curriculum-early-construction}
\resizebox{\textwidth}{!}{
\begin{tabular}{@{}llrrrrrr@{}}
\toprule
Model & Reasoning & Pref. & Counter & Counter+S & Attack & Defence & Attack--Def. \\
\midrule

\multicolumn{8}{l}{\textit{Gemma}} \\
\texttt{google/gemma-4-E2B-it}
& off & 7.5$|$8.2 & 0.0$|$0.1 & 0.0$|$0.2 & 0.0$|$2.2 & 2.5$|$2.5 & 0.0$|$1.4 \\
& on & 10.0$|$10.0 & 0.0$|$0.4 & 0.0$|$0.6 & 0.0$|$3.1 & 7.5$|$7.5 & 0.0$|$1.9 \\

\texttt{google/gemma-4-E4B-it}
& off & 30.0$|$31.8 & 0.0$|$4.8 & 0.0$|$4.4 & 0.0$|$7.0 & 5.0$|$5.0 & 0.0$|$3.9 \\
& on & 20.0$|$21.4 & 0.0$|$3.6 & 0.0$|$5.8 & 0.0$|$6.9 & 12.5$|$12.5 & 0.0$|$5.2 \\

\texttt{google/gemma-4-26B-A4B-it}
& off & 42.5$|$43.2 & 7.5$|$10.2 & 5.0$|$8.0 & 0.0$|$8.5 & 32.5$|$32.5 & 7.5$|$13.9 \\
& on & 42.5$|$42.5 & 17.5$|$19.5 & 22.5$|$23.1 & 10.0$|$16.1 & 40.0$|$40.0 & 22.5$|$27.5 \\

\texttt{google/gemma-4-31B-it}
& off & 50.0$|$50.0 & 10.0$|$12.2 & 22.5$|$20.3 & 15.0$|$17.2 & 50.0$|$50.0 & 12.5$|$20.2 \\
& on & 50.0$|$50.0 & 12.5$|$15.2 & 22.5$|$23.4 & 20.0$|$20.6 & 57.5$|$57.5 & 37.5$|$39.6 \\

\midrule
\multicolumn{8}{l}{\textit{GPT-OSS}} \\
\texttt{openai/gpt-oss-20b}
& low & 25.0$|$25.9 & 10.0$|$12.2 & 5.0$|$5.6 & 12.5$|$13.9 & 35.0$|$34.6 & 25.0$|$27.2 \\
& medium & 32.5$|$34.3 & 32.5$|$32.6 & 10.0$|$10.8 & 35.0$|$30.8 & 55.0$|$54.4 & 37.5$|$41.4 \\
& high & 37.5$|$37.5 & 37.5$|$35.8 & 30.0$|$25.7 & 27.5$|$26.4 & 62.5$|$62.5 & 60.0$|$60.9 \\

\texttt{openai/gpt-oss-120b}
& low & 37.5$|$38.6 & 22.5$|$21.7 & 17.5$|$15.3 & 27.5$|$23.8 & 40.0$|$39.6 & 30.0$|$28.2 \\
& medium & 45.0$|$45.7 & 25.0$|$24.6 & 15.0$|$13.5 & 30.0$|$26.1 & 55.0$|$55.0 & 40.0$|$39.2 \\
& high & 62.5$|$63.1 & 35.0$|$35.5 & 37.5$|$35.0 & 17.5$|$17.8 & 95.0$|$93.1 & 52.5$|$52.7 \\

\midrule
\multicolumn{8}{l}{\textit{Qwen}} \\
\texttt{Qwen/Qwen3.5-0.8B}
& off & 0.0$|$0.0 & 0.0$|$0.0 & 0.0$|$0.0 & 0.0$|$0.0 & 0.0$|$0.0 & 0.0$|$0.1 \\
& on & 0.0$|$0.0 & 0.0$|$0.0 & 0.0$|$0.0 & 0.0$|$0.0 & 0.0$|$0.0 & 0.0$|$0.0 \\

\texttt{Qwen/Qwen3.5-2B}
& off & 2.5$|$2.5 & 0.0$|$0.0 & 0.0$|$0.0 & 0.0$|$0.4 & 0.0$|$0.0 & 0.0$|$0.5 \\
& on & 5.0$|$5.7 & 0.0$|$0.2 & 0.0$|$0.0 & 0.0$|$0.3 & 0.0$|$0.0 & 0.0$|$0.0 \\

\texttt{Qwen/Qwen3.5-4B}
& off & 40.0$|$40.8 & 10.0$|$13.3 & 10.0$|$11.9 & 17.5$|$21.4 & 37.5$|$36.9 & 12.5$|$18.4 \\
& on & 47.5$|$47.5 & 37.5$|$36.1 & 42.5$|$38.6 & 30.0$|$29.8 & 45.0$|$45.0 & 42.5$|$43.4 \\

\texttt{Qwen/Qwen3.5-9B}
& off & 45.0$|$45.0 & 17.5$|$22.3 & 20.0$|$25.2 & 12.5$|$18.7 & 37.5$|$37.5 & 30.0$|$31.8 \\
& on & 47.5$|$47.5 & 27.5$|$30.1 & 72.5$|$67.5 & 35.0$|$36.1 & 47.5$|$47.5 & 37.5$|$37.9 \\

\texttt{Qwen/Qwen3.5-35B-A3B}
& off & 50.0$|$50.0 & 22.5$|$23.9 & 35.0$|$32.3 & 12.5$|$16.9 & 47.5$|$47.5 & 35.0$|$38.3 \\
& on & 50.0$|$50.0 & 17.5$|$21.7 & 45.0$|$45.8 & 42.5$|$44.7 & 45.0$|$45.0 & 37.5$|$39.3 \\

\texttt{Qwen/Qwen3.5-27B}
& off & 47.5$|$47.5 & 42.5$|$43.1 & 47.5$|$44.0 & 70.0$|$70.2 & 65.0$|$65.0 & 62.5$|$62.0 \\
& on & 50.0$|$50.0 & 47.5$|$49.9 & 55.0$|$54.7 & 55.0$|$56.6 & 72.5$|$72.5 & 67.5$|$67.4 \\

\texttt{Qwen/Qwen3.8-27B}
& low & 30.0$|$28.3 & 30.0$|$31.8 & 50.0$|$48.7 & 32.5$|$34.5 & 70.0$|$70.0 & 55.0$|$53.1 \\
& medium & 35.0$|$34.2 & 12.5$|$15.4 & 42.5$|$42.8 & 22.5$|$25.1 & 80.0$|$80.0 & 45.0$|$46.0 \\
& xhigh & 45.0$|$42.0 & 37.5$|$38.4 & 12.5$|$12.5 & 17.5$|$14.9 & 72.5$|$72.5 & 67.5$|$63.1 \\

\midrule
\multicolumn{8}{l}{\textit{DeepSeek}} \\
\texttt{deepseek-v4.1-flash}
& high & 62.5$|$62.1 & 37.5$|$38.8 & 25.0$|$24.6 & 32.5$|$31.0 & 80.0$|$80.0 & 77.5$|$71.7 \\

\midrule
\multicolumn{8}{l}{\textit{Gemini}} \\
\texttt{gemini-2.5-pro}
& low & 50.0$|$49.9 & 10.0$|$10.7 & 10.0$|$8.2 & 32.5$|$36.2 & 55.0$|$54.4 & 15.0$|$20.7 \\
& medium & 52.5$|$52.5 & 17.5$|$19.4 & 15.0$|$14.0 & 17.5$|$20.3 & 65.0$|$65.0 & 57.5$|$57.1 \\
& high & 55.0$|$55.0 & 17.5$|$18.2 & 5.0$|$4.4 & 15.0$|$16.8 & 60.0$|$60.0 & 57.5$|$56.6 \\

\texttt{gemini-3.5-flash-lite}
& low & 45.0$|$45.6 & 17.5$|$20.3 & 10.0$|$10.6 & 20.0$|$22.7 & 67.5$|$67.5 & 55.0$|$59.5 \\
& medium & 42.5$|$42.9 & 30.0$|$32.7 & 25.0$|$22.7 & 32.5$|$35.1 & 72.5$|$72.5 & 45.0$|$50.2 \\
& high & 52.5$|$52.5 & 50.0$|$51.4 & 30.0$|$26.6 & 15.0$|$17.8 & 80.0$|$80.0 & 50.0$|$54.2 \\

\texttt{gemini-3.1-pro-preview}
& low & 50.0$|$50.0 & 15.0$|$16.2 & 2.5$|$2.1 & 60.0$|$59.4 & 90.0$|$90.0 & 75.0$|$76.1 \\
& medium & 55.0$|$54.6 & 35.0$|$35.8 & 20.0$|$19.6 & 60.0$|$59.6 & 95.0$|$95.0 & 90.0$|$89.6 \\
& high & 70.0$|$70.0 & 32.5$|$32.8 & 27.5$|$27.1 & 95.0$|$94.6 & 100.0$|$100.0 & 92.5$|$92.7 \\

\texttt{gemini-3.8-flash}
& low & 50.0$|$50.0 & 57.5$|$58.7 & 52.5$|$50.0 & 92.5$|$91.9 & 87.5$|$87.5 & 67.5$|$67.9 \\
& medium & 55.0$|$55.0 & 65.0$|$66.2 & 62.5$|$62.1 & 100.0$|$100.0 & 95.0$|$95.0 & 95.0$|$95.7 \\
& high & 57.5$|$57.5 & 85.0$|$86.2 & 90.0$|$90.0 & 100.0$|$100.0 & 97.5$|$97.5 & 95.0$|$95.4 \\

\bottomrule
\end{tabular}}
\end{table*}

\begin{table*}[t]
\centering
\tiny
\setlength{\tabcolsep}{2.4pt}
\caption{\textbf{Task-specific analysis/evaluation performance for curriculum configurations L6--L10.} Each cell reports complete-task success (\%) $\,|\, $ task-native score $\times 100$.}
\label{tab:curriculum-middle-analysis}
\resizebox{\textwidth}{!}{
\begin{tabular}{@{}llrrrrrr@{}}
\toprule
Model & Reasoning & Formal. & Chain & Diagnosis & Status & Semantics & Perturb. \\
\midrule

\multicolumn{8}{l}{\textit{Gemma}} \\
\texttt{google/gemma-4-E2B-it}
& off & 0.0$|$0.0 & 0.0$|$1.8 & 0.0$|$6.4 & 0.0$|$44.6 & 0.0$|$36.2 & 0.0$|$2.6 \\
& on & 0.0$|$0.0 & 0.0$|$0.2 & 0.0$|$7.1 & 0.0$|$47.8 & 0.0$|$39.9 & 0.0$|$2.7 \\

\texttt{google/gemma-4-E4B-it}
& off & 0.0$|$2.4 & 2.5$|$4.0 & 0.0$|$11.7 & 0.0$|$67.8 & 0.0$|$54.0 & 0.0$|$6.9 \\
& on & 0.0$|$15.2 & 2.5$|$3.8 & 0.0$|$13.2 & 0.0$|$70.7 & 0.0$|$55.6 & 0.0$|$10.5 \\

\texttt{google/gemma-4-26B-A4B-it}
& off & 2.5$|$40.8 & 0.0$|$0.0 & 0.0$|$22.2 & 2.5$|$72.2 & 0.0$|$57.1 & 0.0$|$20.6 \\
& on & 10.0$|$34.9 & 0.0$|$1.2 & 2.5$|$45.8 & 2.5$|$74.6 & 0.0$|$58.5 & 0.0$|$34.9 \\

\texttt{google/gemma-4-31B-it}
& off & 7.5$|$60.0 & 20.0$|$20.0 & 0.0$|$32.3 & 0.0$|$74.0 & 0.0$|$66.9 & 0.0$|$46.9 \\
& on & 50.0$|$79.9 & 15.0$|$15.0 & 17.5$|$71.5 & 10.0$|$71.8 & 2.5$|$76.0 & 0.0$|$54.0 \\

\midrule
\multicolumn{8}{l}{\textit{GPT-OSS}} \\
\texttt{openai/gpt-oss-20b}
& low & 0.0$|$7.9 & 0.0$|$5.2 & 0.0$|$9.1 & 0.0$|$35.8 & 0.0$|$36.7 & 0.0$|$0.5 \\
& medium & 5.0$|$50.3 & 40.0$|$45.3 & 0.0$|$20.5 & 0.0$|$69.2 & 0.0$|$50.2 & 0.0$|$13.6 \\
& high & 7.5$|$62.2 & 42.5$|$44.7 & 0.0$|$34.0 & 0.0$|$64.5 & 0.0$|$52.9 & 0.0$|$13.5 \\

\texttt{openai/gpt-oss-120b}
& low & 5.0$|$26.5 & 30.0$|$30.7 & 0.0$|$14.1 & 0.0$|$66.2 & 0.0$|$38.8 & 0.0$|$8.5 \\
& medium & 5.0$|$57.7 & 50.0$|$52.9 & 0.0$|$34.7 & 0.0$|$74.8 & 0.0$|$58.3 & 0.0$|$22.0 \\
& high & 7.5$|$74.8 & 37.5$|$37.5 & 2.5$|$57.6 & 15.0$|$78.7 & 2.5$|$65.6 & 0.0$|$28.0 \\

\midrule
\multicolumn{8}{l}{\textit{Qwen}} \\
\texttt{Qwen/Qwen3.5-0.8B}
& off & 0.0$|$0.0 & 0.0$|$0.0 & 0.0$|$5.2 & 0.0$|$14.2 & 0.0$|$1.7 & 0.0$|$1.4 \\
& on & 0.0$|$0.0 & 0.0$|$0.0 & 0.0$|$0.0 & 0.0$|$0.0 & 0.0$|$0.0 & 0.0$|$0.0 \\

\texttt{Qwen/Qwen3.5-2B}
& off & 0.0$|$0.0 & 0.0$|$0.0 & 0.0$|$6.7 & 0.0$|$30.4 & 0.0$|$22.7 & 0.0$|$1.1 \\
& on & 0.0$|$0.9 & 0.0$|$0.0 & 0.0$|$1.1 & 0.0$|$1.0 & 0.0$|$7.7 & 0.0$|$0.0 \\

\texttt{Qwen/Qwen3.5-4B}
& off & 0.0$|$22.5 & 22.5$|$28.1 & 0.0$|$33.4 & 0.0$|$78.0 & 0.0$|$55.1 & 0.0$|$24.5 \\
& on & 2.5$|$22.6 & 35.0$|$37.8 & 0.0$|$40.2 & 7.5$|$78.8 & 2.5$|$60.3 & 0.0$|$22.8 \\

\texttt{Qwen/Qwen3.5-9B}
& off & 0.0$|$25.3 & 20.0$|$22.7 & 0.0$|$45.3 & 12.5$|$85.3 & 0.0$|$62.5 & 0.0$|$41.7 \\
& on & 5.0$|$36.2 & 30.0$|$36.2 & 10.0$|$58.7 & 15.0$|$79.2 & 0.0$|$63.2 & 0.0$|$52.4 \\

\texttt{Qwen/Qwen3.5-35B-A3B}
& off & 10.0$|$45.9 & 25.0$|$30.3 & 7.5$|$70.2 & 17.5$|$89.3 & 10.0$|$76.9 & 0.0$|$57.8 \\
& on & 5.0$|$41.5 & 35.0$|$39.5 & 25.0$|$79.1 & 40.0$|$93.1 & 5.0$|$80.4 & 2.5$|$58.6 \\

\texttt{Qwen/Qwen3.5-27B}
& off & 15.0$|$54.1 & 40.0$|$40.0 & 5.0$|$67.7 & 10.0$|$79.9 & 0.0$|$72.3 & 0.0$|$68.2 \\
& on & 17.5$|$54.4 & 37.5$|$65.6 & 35.0$|$86.2 & 25.0$|$86.4 & 0.0$|$75.0 & 2.5$|$75.5 \\

\texttt{Qwen/Qwen3.8-27B}
& low & 27.5$|$71.3 & 32.5$|$32.5 & 22.5$|$73.2 & 2.5$|$69.1 & 0.0$|$42.9 & 5.0$|$43.0 \\
& medium & 30.0$|$81.8 & 35.0$|$39.7 & 32.5$|$80.2 & 5.0$|$69.3 & 0.0$|$63.5 & 0.0$|$50.5 \\
& xhigh & 32.5$|$84.6 & 62.5$|$64.9 & 47.5$|$91.8 & 40.0$|$93.2 & 10.0$|$75.4 & 17.5$|$67.0 \\

\midrule
\multicolumn{8}{l}{\textit{DeepSeek}} \\
\texttt{deepseek-v4.1-flash}
& high & 65.0$|$90.0 & 52.5$|$52.5 & 55.0$|$88.9 & 27.5$|$90.3 & 7.5$|$67.9 & 50.0$|$84.6 \\

\midrule
\multicolumn{8}{l}{\textit{Gemini}} \\
\texttt{gemini-2.5-pro}
& low & 45.0$|$57.7 & 25.0$|$34.1 & 0.0$|$25.6 & 0.0$|$70.4 & 0.0$|$52.9 & 0.0$|$25.4 \\
& medium & 32.5$|$59.2 & 35.0$|$44.4 & 37.5$|$89.4 & 35.0$|$93.0 & 20.0$|$74.8 & 20.0$|$92.0 \\
& high & 35.0$|$56.5 & 50.0$|$61.7 & 27.5$|$86.0 & 25.0$|$92.2 & 2.5$|$73.8 & 17.5$|$91.2 \\

\texttt{gemini-3.5-flash-lite}
& low & 15.0$|$36.0 & 20.0$|$20.0 & 0.0$|$19.0 & 2.5$|$60.1 & 0.0$|$57.3 & 0.0$|$18.3 \\
& medium & 55.0$|$85.7 & 20.0$|$22.2 & 0.0$|$63.8 & 7.5$|$83.3 & 2.5$|$63.8 & 0.0$|$36.8 \\
& high & 82.5$|$86.9 & 37.5$|$38.8 & 30.0$|$82.6 & 20.0$|$89.4 & 2.5$|$71.9 & 2.5$|$67.5 \\

\texttt{gemini-3.1-pro-preview}
& low & 97.5$|$89.1 & 42.5$|$67.3 & 40.0$|$89.7 & 5.0$|$28.3 & 10.0$|$76.5 & 20.0$|$78.9 \\
& medium & 97.5$|$87.2 & 65.0$|$90.1 & 45.0$|$91.5 & 0.0$|$0.0 & 42.5$|$90.4 & 50.0$|$90.6 \\
& high & 97.5$|$89.3 & 50.0$|$93.1 & 52.5$|$92.9 & 25.0$|$56.8 & 77.5$|$95.2 & 72.5$|$94.9 \\

\texttt{gemini-3.8-flash}
& low & 95.0$|$89.0 & 62.5$|$68.3 & 50.0$|$92.5 & 60.0$|$96.4 & 60.0$|$95.0 & 57.5$|$86.2 \\
& medium & 97.5$|$94.2 & 85.0$|$93.5 & 50.0$|$92.5 & 62.5$|$97.0 & 87.5$|$97.9 & 95.0$|$99.5 \\
& high & 100.0$|$96.3 & 92.5$|$96.5 & 50.0$|$92.5 & 57.5$|$96.8 & 97.5$|$99.8 & 95.0$|$99.5 \\

\bottomrule
\end{tabular}}
\end{table*}

\begin{table*}[t]
\centering
\tiny
\setlength{\tabcolsep}{2.4pt}
\caption{\textbf{Task-specific constructive performance for curriculum configurations L6--L10.} Each cell reports complete-task success (\%) $\,|\, $ task-native score $\times 100$.}
\label{tab:curriculum-middle-construction}
\resizebox{\textwidth}{!}{
\begin{tabular}{@{}llrrrrrr@{}}
\toprule
Model & Reasoning & Pref. & Counter & Counter+S & Attack & Defence & Attack--Def. \\
\midrule

\multicolumn{8}{l}{\textit{Gemma}} \\
\texttt{google/gemma-4-E2B-it}
& off & 0.0$|$0.0 & 0.0$|$0.0 & 0.0$|$0.0 & 0.0$|$0.4 & 0.0$|$0.0 & 0.0$|$0.4 \\
& on & 0.0$|$0.4 & 0.0$|$0.2 & 0.0$|$0.0 & 0.0$|$0.4 & 0.0$|$0.0 & 0.0$|$0.6 \\

\texttt{google/gemma-4-E4B-it}
& off & 5.0$|$6.8 & 0.0$|$1.4 & 0.0$|$0.9 & 0.0$|$2.6 & 2.5$|$4.5 & 0.0$|$2.1 \\
& on & 2.5$|$7.4 & 0.0$|$2.6 & 0.0$|$2.5 & 0.0$|$3.3 & 0.0$|$1.7 & 2.5$|$6.2 \\

\texttt{google/gemma-4-26B-A4B-it}
& off & 27.5$|$30.9 & 0.0$|$2.1 & 2.5$|$5.8 & 0.0$|$3.6 & 20.0$|$20.7 & 12.5$|$17.9 \\
& on & 30.0$|$33.2 & 2.5$|$8.3 & 22.5$|$24.0 & 12.5$|$18.4 & 17.5$|$20.5 & 15.0$|$20.1 \\

\texttt{google/gemma-4-31B-it}
& off & 40.0$|$41.8 & 5.0$|$8.9 & 10.0$|$13.6 & 10.0$|$11.8 & 30.0$|$33.2 & 10.0$|$15.5 \\
& on & 45.0$|$45.3 & 15.0$|$18.2 & 15.0$|$17.7 & 20.0$|$20.9 & 47.5$|$48.8 & 37.5$|$41.0 \\

\midrule
\multicolumn{8}{l}{\textit{GPT-OSS}} \\
\texttt{openai/gpt-oss-20b}
& low & 0.0$|$0.1 & 0.0$|$0.5 & 0.0$|$0.1 & 0.0$|$1.0 & 12.5$|$11.6 & 2.5$|$10.7 \\
& medium & 12.5$|$17.3 & 12.5$|$13.5 & 15.0$|$13.5 & 10.0$|$11.9 & 42.5$|$43.1 & 32.5$|$35.8 \\
& high & 25.0$|$27.8 & 30.0$|$27.7 & 22.5$|$21.8 & 10.0$|$12.6 & 35.0$|$35.9 & 32.5$|$36.3 \\

\texttt{openai/gpt-oss-120b}
& low & 2.5$|$6.4 & 7.5$|$8.1 & 2.5$|$2.5 & 10.0$|$11.7 & 37.5$|$37.2 & 22.5$|$23.1 \\
& medium & 25.0$|$28.7 & 17.5$|$15.5 & 17.5$|$16.7 & 7.5$|$7.4 & 42.5$|$42.9 & 57.5$|$55.1 \\
& high & 62.5$|$63.3 & 42.5$|$38.9 & 12.5$|$12.3 & 17.5$|$18.0 & 55.0$|$54.3 & 52.5$|$52.4 \\

\midrule
\multicolumn{8}{l}{\textit{Qwen}} \\
\texttt{Qwen/Qwen3.5-0.8B}
& off & 0.0$|$0.0 & 0.0$|$0.0 & 0.0$|$0.0 & 0.0$|$0.0 & 0.0$|$0.0 & 0.0$|$0.0 \\
& on & 0.0$|$0.0 & 0.0$|$0.0 & 0.0$|$0.0 & 0.0$|$0.0 & 0.0$|$0.0 & 0.0$|$0.0 \\

\texttt{Qwen/Qwen3.5-2B}
& off & 0.0$|$0.0 & 0.0$|$0.0 & 0.0$|$0.0 & 0.0$|$0.0 & 0.0$|$0.0 & 0.0$|$0.2 \\
& on & 0.0$|$0.0 & 0.0$|$0.0 & 0.0$|$0.2 & 0.0$|$0.0 & 2.5$|$2.5 & 0.0$|$0.0 \\

\texttt{Qwen/Qwen3.5-4B}
& off & 35.0$|$36.4 & 0.0$|$2.7 & 5.0$|$6.8 & 10.0$|$12.6 & 15.0$|$16.0 & 2.5$|$9.1 \\
& on & 45.0$|$45.5 & 22.5$|$21.9 & 17.5$|$18.4 & 10.0$|$13.2 & 32.5$|$33.9 & 20.0$|$22.6 \\

\texttt{Qwen/Qwen3.5-9B}
& off & 42.5$|$43.0 & 15.0$|$17.9 & 5.0$|$8.6 & 5.0$|$7.8 & 20.0$|$22.2 & 22.5$|$25.6 \\
& on & 45.0$|$45.0 & 30.0$|$30.6 & 32.5$|$33.6 & 17.5$|$19.3 & 35.0$|$37.4 & 45.0$|$45.0 \\

\texttt{Qwen/Qwen3.5-35B-A3B}
& off & 40.0$|$41.2 & 7.5$|$10.6 & 10.0$|$12.6 & 5.0$|$7.1 & 22.5$|$24.1 & 15.0$|$20.8 \\
& on & 45.0$|$45.0 & 20.0$|$24.2 & 57.5$|$53.8 & 27.5$|$29.9 & 37.5$|$38.5 & 42.5$|$42.6 \\

\texttt{Qwen/Qwen3.5-27B}
& off & 47.5$|$48.0 & 17.5$|$20.7 & 40.0$|$35.0 & 27.5$|$29.2 & 50.0$|$50.5 & 47.5$|$49.0 \\
& on & 50.0$|$50.0 & 42.5$|$44.5 & 40.0$|$39.8 & 45.0$|$47.2 & 50.0$|$52.3 & 50.0$|$50.8 \\

\texttt{Qwen/Qwen3.8-27B}
& low & 32.5$|$32.8 & 22.5$|$26.6 & 60.0$|$58.6 & 17.5$|$21.7 & 45.0$|$46.8 & 45.0$|$46.1 \\
& medium & 35.0$|$36.1 & 32.5$|$36.3 & 42.5$|$44.4 & 20.0$|$24.7 & 55.0$|$56.6 & 37.5$|$40.5 \\
& xhigh & 50.0$|$50.0 & 47.5$|$48.5 & 15.0$|$15.0 & 42.5$|$38.4 & 42.5$|$44.5 & 57.5$|$56.5 \\

\midrule
\multicolumn{8}{l}{\textit{DeepSeek}} \\
\texttt{deepseek-v4.1-flash}
& high & 57.5$|$57.5 & 32.5$|$32.8 & 12.5$|$12.6 & 37.5$|$37.4 & 52.5$|$53.0 & 60.0$|$58.8 \\

\midrule
\multicolumn{8}{l}{\textit{Gemini}} \\
\texttt{gemini-2.5-pro}
& low & 47.5$|$47.7 & 0.0$|$1.6 & 5.0$|$5.2 & 12.5$|$16.2 & 27.5$|$28.7 & 10.0$|$14.6 \\
& medium & 57.5$|$57.5 & 22.5$|$25.7 & 15.0$|$14.6 & 17.5$|$18.9 & 35.0$|$36.8 & 55.0$|$55.1 \\
& high & 47.5$|$47.9 & 30.0$|$32.5 & 5.0$|$5.8 & 25.0$|$27.4 & 45.0$|$47.0 & 40.0$|$42.1 \\

\texttt{gemini-3.5-flash-lite}
& low & 15.0$|$18.7 & 7.5$|$9.3 & 2.5$|$3.8 & 10.0$|$10.6 & 47.5$|$47.7 & 22.5$|$25.6 \\
& medium & 40.0$|$41.5 & 20.0$|$21.9 & 7.5$|$8.1 & 15.0$|$18.1 & 62.5$|$63.8 & 32.5$|$39.9 \\
& high & 45.0$|$46.0 & 32.5$|$34.0 & 7.5$|$6.9 & 15.0$|$15.4 & 50.0$|$50.3 & 42.5$|$50.2 \\

\texttt{gemini-3.1-pro-preview}
& low & 50.0$|$50.0 & 55.0$|$55.8 & 20.0$|$17.4 & 32.5$|$33.5 & 72.5$|$73.0 & 70.0$|$71.4 \\
& medium & 50.0$|$50.0 & 67.5$|$68.8 & 25.0$|$23.7 & 30.0$|$29.8 & 60.0$|$61.5 & 90.0$|$88.9 \\
& high & 60.0$|$60.0 & 70.0$|$71.0 & 32.5$|$30.9 & 50.0$|$50.3 & 72.5$|$73.5 & 90.0$|$90.3 \\

\texttt{gemini-3.8-flash}
& low & 50.0$|$50.0 & 45.0$|$44.7 & 27.5$|$22.9 & 72.5$|$72.6 & 85.0$|$84.9 & 60.0$|$61.6 \\
& medium & 52.5$|$52.5 & 75.0$|$75.2 & 27.5$|$25.1 & 100.0$|$100.0 & 90.0$|$90.0 & 97.5$|$97.2 \\
& high & 57.5$|$57.5 & 85.0$|$85.5 & 65.0$|$62.2 & 100.0$|$100.0 & 90.0$|$90.0 & 100.0$|$100.0 \\

\bottomrule
\end{tabular}}
\end{table*}

\begin{table*}[t]
\centering
\tiny
\setlength{\tabcolsep}{2.4pt}
\caption{\textbf{Task-specific analysis/evaluation performance for curriculum configurations L11--L15.} Each cell reports complete-task success (\%) $\,|\, $ task-native score $\times 100$.}
\label{tab:curriculum-late-analysis}
\resizebox{\textwidth}{!}{
\begin{tabular}{@{}llrrrrrr@{}}
\toprule
Model & Reasoning & Formal. & Chain & Diagnosis & Status & Semantics & Perturb. \\
\midrule

\multicolumn{8}{l}{\textit{Gemma}} \\
\texttt{google/gemma-4-E2B-it}
& off & 0.0$|$0.0 & 0.0$|$0.3 & 0.0$|$6.7 & 0.0$|$42.1 & 0.0$|$37.7 & 0.0$|$0.5 \\
& on & 0.0$|$0.0 & 0.0$|$0.0 & 0.0$|$6.7 & 0.0$|$43.4 & 0.0$|$39.5 & 0.0$|$2.5 \\

\texttt{google/gemma-4-E4B-it}
& off & 0.0$|$15.9 & 0.0$|$0.8 & 0.0$|$8.2 & 0.0$|$59.0 & 0.0$|$50.7 & 0.0$|$3.3 \\
& on & 0.0$|$24.9 & 2.5$|$3.3 & 0.0$|$10.7 & 0.0$|$58.6 & 0.0$|$52.8 & 0.0$|$7.5 \\

\texttt{google/gemma-4-26B-A4B-it}
& off & 0.0$|$39.8 & 0.0$|$2.3 & 0.0$|$10.6 & 0.0$|$66.2 & 0.0$|$61.2 & 0.0$|$10.2 \\
& on & 5.0$|$49.1 & 5.0$|$6.4 & 0.0$|$28.4 & 2.5$|$63.7 & 0.0$|$58.3 & 0.0$|$29.8 \\

\texttt{google/gemma-4-31B-it}
& off & 7.5$|$68.7 & 7.5$|$8.9 & 0.0$|$26.4 & 5.0$|$72.0 & 0.0$|$65.0 & 0.0$|$36.8 \\
& on & 47.5$|$80.6 & 12.5$|$12.5 & 2.5$|$44.6 & 5.0$|$74.2 & 0.0$|$75.0 & 0.0$|$48.7 \\

\midrule
\multicolumn{8}{l}{\textit{GPT-OSS}} \\
\texttt{openai/gpt-oss-20b}
& low & 0.0$|$1.2 & 0.0$|$1.1 & 0.0$|$7.5 & 0.0$|$28.1 & 0.0$|$33.1 & 0.0$|$0.3 \\
& medium & 0.0$|$50.5 & 2.5$|$7.1 & 0.0$|$29.8 & 0.0$|$57.7 & 2.5$|$51.2 & 0.0$|$7.1 \\
& high & 0.0$|$61.1 & 17.5$|$21.3 & 0.0$|$28.8 & 0.0$|$55.9 & 0.0$|$55.8 & 0.0$|$8.8 \\

\texttt{openai/gpt-oss-120b}
& low & 0.0$|$19.6 & 0.0$|$3.2 & 0.0$|$13.2 & 0.0$|$56.2 & 0.0$|$41.0 & 0.0$|$4.3 \\
& medium & 2.5$|$61.6 & 20.0$|$20.3 & 0.0$|$30.0 & 0.0$|$73.2 & 0.0$|$62.3 & 0.0$|$13.2 \\
& high & 2.5$|$75.6 & 27.5$|$27.9 & 0.0$|$44.9 & 2.5$|$75.6 & 0.0$|$66.0 & 0.0$|$24.0 \\

\midrule
\multicolumn{8}{l}{\textit{Qwen}} \\
\texttt{Qwen/Qwen3.5-0.8B}
& off & 0.0$|$0.0 & 0.0$|$0.0 & 0.0$|$3.7 & 0.0$|$6.8 & 0.0$|$4.1 & 0.0$|$0.2 \\
& on & 0.0$|$0.0 & 0.0$|$0.0 & 0.0$|$0.0 & 0.0$|$0.0 & 0.0$|$0.0 & 0.0$|$0.0 \\

\texttt{Qwen/Qwen3.5-2B}
& off & 0.0$|$0.0 & 0.0$|$0.7 & 0.0$|$3.0 & 0.0$|$15.1 & 0.0$|$20.4 & 0.0$|$0.3 \\
& on & 0.0$|$2.1 & 0.0$|$0.0 & 0.0$|$1.1 & 0.0$|$0.0 & 0.0$|$13.3 & 0.0$|$0.0 \\

\texttt{Qwen/Qwen3.5-4B}
& off & 0.0$|$12.6 & 12.5$|$19.5 & 0.0$|$22.7 & 2.5$|$76.8 & 0.0$|$54.2 & 0.0$|$13.4 \\
& on & 2.5$|$7.2 & 27.5$|$32.2 & 2.5$|$32.3 & 5.0$|$60.7 & 0.0$|$60.5 & 0.0$|$13.0 \\

\texttt{Qwen/Qwen3.5-9B}
& off & 0.0$|$23.4 & 15.0$|$19.3 & 0.0$|$33.5 & 7.5$|$84.9 & 0.0$|$59.5 & 0.0$|$28.6 \\
& on & 7.5$|$25.7 & 20.0$|$26.1 & 2.5$|$47.4 & 2.5$|$76.9 & 0.0$|$69.6 & 0.0$|$27.8 \\

\texttt{Qwen/Qwen3.5-35B-A3B}
& off & 2.5$|$31.6 & 17.5$|$23.3 & 0.0$|$38.9 & 2.5$|$84.3 & 7.5$|$78.8 & 0.0$|$39.5 \\
& on & 5.0$|$26.3 & 10.0$|$13.5 & 5.0$|$58.4 & 12.5$|$87.6 & 2.5$|$79.6 & 0.0$|$39.1 \\

\texttt{Qwen/Qwen3.5-27B}
& off & 10.0$|$51.2 & 17.5$|$17.5 & 0.0$|$58.4 & 20.0$|$88.8 & 2.5$|$73.3 & 0.0$|$56.0 \\
& on & 22.5$|$67.7 & 27.5$|$27.5 & 15.0$|$78.6 & 10.0$|$84.5 & 5.0$|$79.8 & 0.0$|$60.4 \\

\texttt{Qwen/Qwen3.8-27B}
& low & 30.0$|$76.9 & 15.0$|$15.0 & 5.0$|$57.6 & 2.5$|$66.1 & 0.0$|$61.2 & 0.0$|$33.7 \\
& medium & 25.0$|$67.0 & 27.5$|$27.8 & 17.5$|$67.7 & 2.5$|$72.4 & 2.5$|$65.2 & 0.0$|$51.4 \\
& xhigh & 32.5$|$87.5 & 27.5$|$29.6 & 42.5$|$91.2 & 25.0$|$85.3 & 5.0$|$71.9 & 7.5$|$66.1 \\

\midrule
\multicolumn{8}{l}{\textit{DeepSeek}} \\
\texttt{deepseek-v4.1-flash}
& high & 52.5$|$93.1 & 37.5$|$37.8 & 37.5$|$83.1 & 22.5$|$88.3 & 2.5$|$66.9 & 40.0$|$81.8 \\

\midrule
\multicolumn{8}{l}{\textit{Gemini}} \\
\texttt{gemini-2.5-pro}
& low & 27.5$|$51.6 & 5.0$|$13.8 & 0.0$|$15.6 & 0.0$|$66.9 & 2.5$|$55.0 & 0.0$|$17.0 \\
& medium & 47.5$|$75.6 & 20.0$|$28.4 & 17.5$|$83.6 & 10.0$|$91.0 & 12.5$|$75.4 & 2.5$|$81.5 \\
& high & 32.5$|$67.9 & 20.0$|$29.3 & 20.0$|$83.8 & 12.5$|$90.4 & 12.5$|$76.0 & 2.5$|$87.2 \\

\texttt{gemini-3.5-flash-lite}
& low & 5.0$|$34.4 & 10.0$|$12.1 & 0.0$|$10.8 & 0.0$|$48.8 & 2.5$|$59.8 & 0.0$|$11.7 \\
& medium & 47.5$|$88.7 & 10.0$|$10.6 & 7.5$|$66.8 & 5.0$|$75.8 & 2.5$|$66.7 & 0.0$|$20.0 \\
& high & 80.0$|$89.2 & 22.5$|$22.8 & 12.5$|$79.2 & 7.5$|$88.0 & 2.5$|$72.3 & 0.0$|$60.3 \\

\texttt{gemini-3.1-pro-preview}
& low & 95.0$|$91.6 & 25.0$|$45.2 & 45.0$|$91.3 & 2.5$|$27.7 & 2.5$|$81.5 & 10.0$|$68.8 \\
& medium & 100.0$|$93.5 & 15.0$|$68.8 & 55.0$|$93.0 & 0.0$|$4.7 & 40.0$|$92.5 & 30.0$|$92.1 \\
& high & 95.0$|$93.5 & 27.5$|$84.4 & 55.0$|$93.3 & 12.5$|$47.5 & 57.5$|$94.6 & 45.0$|$95.6 \\

\texttt{gemini-3.8-flash}
& low & 97.5$|$93.3 & 15.0$|$22.6 & 50.0$|$91.3 & 30.0$|$92.9 & 47.5$|$93.8 & 30.0$|$61.3 \\
& medium & 97.5$|$96.6 & 80.0$|$88.8 & 50.0$|$92.5 & 50.0$|$97.1 & 92.5$|$98.3 & 80.0$|$96.2 \\
& high & 97.5$|$98.3 & 65.0$|$86.3 & 50.0$|$92.5 & 50.0$|$97.1 & 97.5$|$99.6 & 82.5$|$98.7 \\

\bottomrule
\end{tabular}}
\end{table*}

\begin{table*}[t]
\centering
\tiny
\setlength{\tabcolsep}{2.4pt}
\caption{\textbf{Task-specific constructive performance for curriculum configurations L11--L15.} Each cell reports complete-task success (\%) $\,|\, $ task-native score $\times 100$.}
\label{tab:curriculum-late-construction}
\resizebox{\textwidth}{!}{
\begin{tabular}{@{}llrrrrrr@{}}
\toprule
Model & Reasoning & Pref. & Counter & Counter+S & Attack & Defence & Attack--Def. \\
\midrule

\multicolumn{8}{l}{\textit{Gemma}} \\
\texttt{google/gemma-4-E2B-it}
& off & 0.0$|$0.0 & 0.0$|$0.0 & 0.0$|$0.0 & 0.0$|$0.1 & 0.0$|$0.0 & 0.0$|$0.1 \\
& on & 0.0$|$0.0 & 0.0$|$0.0 & 0.0$|$0.0 & 0.0$|$0.1 & 0.0$|$0.0 & 0.0$|$0.4 \\

\texttt{google/gemma-4-E4B-it}
& off & 2.5$|$5.5 & 0.0$|$0.2 & 0.0$|$0.0 & 0.0$|$2.2 & 0.0$|$4.0 & 0.0$|$1.1 \\
& on & 2.5$|$6.0 & 0.0$|$0.8 & 0.0$|$2.4 & 0.0$|$2.3 & 0.0$|$6.3 & 0.0$|$3.5 \\

\texttt{google/gemma-4-26B-A4B-it}
& off & 22.5$|$26.2 & 0.0$|$2.2 & 5.0$|$6.7 & 0.0$|$0.8 & 10.0$|$13.2 & 10.0$|$13.8 \\
& on & 32.5$|$34.3 & 5.0$|$11.0 & 2.5$|$7.3 & 5.0$|$7.6 & 10.0$|$16.9 & 15.0$|$20.5 \\

\texttt{google/gemma-4-31B-it}
& off & 32.5$|$35.6 & 0.0$|$4.8 & 12.5$|$15.1 & 0.0$|$2.6 & 5.0$|$12.0 & 5.0$|$12.0 \\
& on & 45.0$|$45.8 & 15.0$|$20.0 & 12.5$|$15.0 & 20.0$|$21.7 & 12.5$|$19.4 & 27.5$|$32.9 \\

\midrule
\multicolumn{8}{l}{\textit{GPT-OSS}} \\
\texttt{openai/gpt-oss-20b}
& low & 0.0$|$0.1 & 0.0$|$0.2 & 2.5$|$2.5 & 0.0$|$0.2 & 0.0$|$1.0 & 10.0$|$14.9 \\
& medium & 10.0$|$11.9 & 12.5$|$12.7 & 10.0$|$10.0 & 2.5$|$4.9 & 17.5$|$18.6 & 17.5$|$21.2 \\
& high & 7.5$|$11.8 & 17.5$|$17.1 & 20.0$|$20.3 & 17.5$|$17.8 & 22.5$|$24.5 & 47.5$|$50.2 \\

\texttt{openai/gpt-oss-120b}
& low & 0.0$|$2.8 & 2.5$|$3.8 & 0.0$|$0.8 & 0.0$|$1.3 & 12.5$|$13.2 & 17.5$|$19.2 \\
& medium & 32.5$|$35.0 & 5.0$|$6.6 & 7.5$|$8.7 & 2.5$|$5.3 & 27.5$|$29.0 & 37.5$|$36.9 \\
& high & 45.0$|$46.0 & 15.0$|$15.1 & 7.5$|$6.9 & 5.0$|$6.4 & 30.0$|$30.1 & 57.5$|$56.9 \\

\midrule
\multicolumn{8}{l}{\textit{Qwen}} \\
\texttt{Qwen/Qwen3.5-0.8B}
& off & 0.0$|$0.0 & 0.0$|$0.0 & 0.0$|$0.0 & 0.0$|$0.0 & 0.0$|$0.0 & 0.0$|$0.1 \\
& on & 0.0$|$0.0 & 0.0$|$0.0 & 0.0$|$0.0 & 0.0$|$0.0 & 0.0$|$0.0 & 0.0$|$0.0 \\

\texttt{Qwen/Qwen3.5-2B}
& off & 0.0$|$0.0 & 0.0$|$0.0 & 0.0$|$0.0 & 0.0$|$0.0 & 0.0$|$0.0 & 0.0$|$0.1 \\
& on & 0.0$|$0.0 & 0.0$|$0.0 & 0.0$|$0.0 & 0.0$|$0.0 & 0.0$|$0.0 & 0.0$|$0.0 \\

\texttt{Qwen/Qwen3.5-4B}
& off & 20.0$|$25.2 & 0.0$|$2.5 & 2.5$|$3.8 & 0.0$|$2.4 & 0.0$|$7.0 & 2.5$|$8.3 \\
& on & 37.5$|$40.0 & 5.0$|$7.3 & 5.0$|$7.3 & 12.5$|$14.1 & 12.5$|$16.4 & 12.5$|$16.4 \\

\texttt{Qwen/Qwen3.5-9B}
& off & 35.0$|$37.4 & 0.0$|$4.5 & 2.5$|$6.2 & 0.0$|$4.3 & 17.5$|$22.2 & 15.0$|$20.4 \\
& on & 37.5$|$38.6 & 5.0$|$10.1 & 27.5$|$29.7 & 20.0$|$21.2 & 15.0$|$19.9 & 27.5$|$32.1 \\

\texttt{Qwen/Qwen3.5-35B-A3B}
& off & 30.0$|$33.7 & 2.5$|$3.6 & 2.5$|$5.0 & 0.0$|$2.5 & 10.0$|$15.8 & 15.0$|$19.6 \\
& on & 45.0$|$46.3 & 10.0$|$13.8 & 25.0$|$28.7 & 22.5$|$22.7 & 30.0$|$35.2 & 37.5$|$40.7 \\

\texttt{Qwen/Qwen3.5-27B}
& off & 47.5$|$47.5 & 10.0$|$12.4 & 7.5$|$8.5 & 15.0$|$15.0 & 25.0$|$29.4 & 45.0$|$47.3 \\
& on & 47.5$|$48.0 & 35.0$|$38.2 & 32.5$|$31.2 & 32.5$|$36.2 & 32.5$|$37.2 & 55.0$|$56.9 \\

\texttt{Qwen/Qwen3.8-27B}
& low & 37.5$|$39.5 & 25.0$|$29.5 & 42.5$|$40.8 & 15.0$|$19.6 & 22.5$|$28.2 & 30.0$|$34.1 \\
& medium & 47.5$|$48.0 & 37.5$|$42.6 & 37.5$|$37.4 & 7.5$|$11.8 & 37.5$|$41.9 & 42.5$|$44.4 \\
& xhigh & 55.0$|$55.0 & 42.5$|$44.1 & 7.5$|$7.5 & 22.5$|$21.3 & 22.5$|$28.0 & 55.0$|$54.1 \\

\midrule
\multicolumn{8}{l}{\textit{DeepSeek}} \\
\texttt{deepseek-v4.1-flash}
& high & 60.0$|$60.2 & 32.5$|$35.0 & 2.5$|$3.4 & 27.5$|$27.7 & 30.0$|$30.2 & 75.0$|$72.3 \\

\midrule
\multicolumn{8}{l}{\textit{Gemini}} \\
\texttt{gemini-2.5-pro}
& low & 37.5$|$39.3 & 0.0$|$1.6 & 0.0$|$0.5 & 0.0$|$1.3 & 5.0$|$7.3 & 12.5$|$17.5 \\
& medium & 50.0$|$50.4 & 30.0$|$33.4 & 17.5$|$18.9 & 10.0$|$11.2 & 20.0$|$24.7 & 40.0$|$41.8 \\
& high & 55.0$|$55.0 & 30.0$|$33.1 & 2.5$|$4.7 & 20.0$|$21.1 & 20.0$|$24.2 & 37.5$|$40.5 \\

\texttt{gemini-3.5-flash-lite}
& low & 0.0$|$4.4 & 2.5$|$4.0 & 0.0$|$0.2 & 5.0$|$5.3 & 20.0$|$21.1 & 10.0$|$16.5 \\
& medium & 42.5$|$43.6 & 17.5$|$20.5 & 0.0$|$1.2 & 7.5$|$8.9 & 47.5$|$48.5 & 35.0$|$40.1 \\
& high & 35.0$|$37.9 & 7.5$|$8.9 & 7.5$|$7.8 & 10.0$|$11.1 & 30.0$|$32.5 & 32.5$|$40.3 \\

\texttt{gemini-3.1-pro-preview}
& low & 47.5$|$47.6 & 70.0$|$70.8 & 32.5$|$29.6 & 32.5$|$32.0 & 47.5$|$49.1 & 62.5$|$64.9 \\
& medium & 52.5$|$52.5 & 67.5$|$69.9 & 17.5$|$15.8 & 17.5$|$17.5 & 42.5$|$44.0 & 80.0$|$80.8 \\
& high & 55.0$|$55.0 & 70.0$|$72.8 & 10.0$|$9.5 & 17.5$|$18.0 & 57.5$|$59.8 & 92.5$|$93.0 \\

\texttt{gemini-3.8-flash}
& low & 55.0$|$55.6 & 52.5$|$52.4 & 10.0$|$8.8 & 50.0$|$50.2 & 77.5$|$77.8 & 52.5$|$55.7 \\
& medium & 65.0$|$65.0 & 90.0$|$90.5 & 5.0$|$4.4 & 95.0$|$95.0 & 87.5$|$87.5 & 90.0$|$90.9 \\
& high & 75.0$|$75.0 & 90.0$|$90.3 & 25.0$|$22.5 & 97.5$|$97.5 & 90.0$|$90.0 & 97.5$|$97.8 \\

\bottomrule
\end{tabular}}
\end{table*}

The task-level decomposition shows that structural richness does not affect all argumentative operations uniformly. For many analysis tasks, complete-task success falls much more sharply than the corresponding native score. For example, Qwen3.5-27B with thinking enabled reaches 72.5\% complete success with a 95.1 native score on \texttt{status\_query} in L1--L5, but only 10.0\% complete success with an 84.5 native score in L11--L15. This pattern indicates that substantial local correctness can persist even when models increasingly fail to assemble the complete argumentative state required by the task.

Constructive tasks show a different profile. Some intervention types remain comparatively robust for stronger models, while others remain difficult or vary non-monotonically across bands. In particular, the strict-rule counter-argument condition remains challenging for many systems even when other constructive operations improve substantially. The curriculum bands should therefore not be interpreted as calibrated scalar difficulty levels: individual tasks respond differently to the structural properties varied by their generators, and occasional non-monotonic changes occur. These results characterize task-specific robustness under increasingly rich generated configurations rather than a universal difficulty progression.

\FloatBarrier
\subsection{Argument-Ordering Analysis}
\label{app:ordering-analysis}

ArgGYM varies argument comparison along two formal dimensions: the argument
comparison principle (last-link versus weakest-link) and the set ordering used
to lift preferences (democratic versus elitist). These dimensions are not
equally active in every generated theory. Democratic and elitist comparison
can differ only when the relevant comparison involves multi-element sets. Under
last-link, arguments in the current generators almost always contribute a
single last defeasible rule; consequently, the democratic and elitist variants
induce the same defeat relation on eleven of the twelve tasks, with
\texttt{status\_query} providing the designed exception. Under weakest-link,
multi-element comparison is more readily available, with the set-ordering
distinction becoming directly task-relevant in particular for
\texttt{preference\_construction}. Table~\ref{tab:ordering-results} reports the
benchmark-wide condition slices for completeness, but the democratic and
elitist columns should therefore not be interpreted as a uniform experimental
contrast across all tasks.

As a separate implementation diagnostic, we also recomputed each frozen theory
after switching only the argument comparison principle between last-link and
weakest-link while preserving the set ordering. The resulting grounded status
map changed for 107 of 120 \texttt{status\_query} theories and 101 of 120
\texttt{defeat\_diagnosis} theories, as well as for all 120 theories in
\texttt{preference\_construction}, \texttt{claim\_chain}, and
\texttt{perturbation}. Thus, the link principle is formally active in the
generated theories underlying several of the largest observed ordering
differences. This same-theory diagnostic is distinct from the benchmark
performance comparison below, whose last-link and weakest-link instances are
independently generated and therefore remain descriptive rather than causal. For the model comparison, each of the 120 frozen \texttt{status\_query} theories was re-posed with the same queried claims under the other link principle, keeping its set ordering, with gold statuses recomputed by the engine. The untrained Qwen3.5-4B answered the re-posed set once under the protocol of Section~\ref{sec:model-setup}; each answer is paired with the model's answer to the original question in its frozen-benchmark re-evaluation (seed 0), and intervals are paired bootstrap intervals over theories.

\begin{table*}[t]
\centering
\scriptsize
\setlength{\tabcolsep}{5pt}
\caption{\textbf{Performance by argument and set ordering.} Values are complete-task success (\%). Last-link and weakest-link refer to the argument comparison principle; democratic and elitist refer to the set ordering used for preference comparison.}
\label{tab:ordering-results}
\begin{tabular}{@{}llrrrr@{}}
\toprule
Model & Reasoning & Last-link & Weakest-link & Democratic & Elitist \\
\midrule

\multicolumn{6}{l}{\textit{Gemma}} \\
\texttt{google/gemma-4-E2B-it}
& off & 1.0 & 1.2 & 1.4 & 0.8 \\
& on & 1.5 & 1.1 & 1.2 & 1.4 \\

\texttt{google/gemma-4-E4B-it}
& off & 5.1 & 1.9 & 3.9 & 3.2 \\
& on & 4.9 & 2.6 & 3.9 & 3.6 \\

\texttt{google/gemma-4-26B-A4B-it}
& off & 14.0 & 4.0 & 9.3 & 8.8 \\
& on & 20.8 & 7.4 & 14.7 & 13.5 \\

\texttt{google/gemma-4-31B-it}
& off & 22.6 & 6.1 & 16.0 & 12.8 \\
& on & 37.4 & 17.2 & 27.6 & 26.9 \\

\midrule
\multicolumn{6}{l}{\textit{GPT-OSS}} \\
\texttt{openai/gpt-oss-20b}
& low & 9.9 & 2.6 & 6.5 & 6.0 \\
& medium & 23.3 & 8.8 & 16.4 & 15.7 \\
& high & 29.9 & 12.6 & 23.2 & 19.3 \\

\texttt{openai/gpt-oss-120b}
& low & 19.6 & 5.1 & 12.1 & 12.6 \\
& medium & 32.4 & 10.7 & 21.5 & 21.5 \\
& high & 35.4 & 22.5 & 28.5 & 29.4 \\

\midrule
\multicolumn{6}{l}{\textit{Qwen}} \\
\texttt{Qwen/Qwen3.5-0.8B}
& off & 0.0 & 0.0 & 0.0 & 0.0 \\
& on & 0.0 & 0.0 & 0.0 & 0.0 \\

\texttt{Qwen/Qwen3.5-2B}
& off & 0.6 & 0.1 & 0.3 & 0.4 \\
& on & 1.2 & 0.8 & 1.4 & 0.7 \\

\texttt{Qwen/Qwen3.5-4B}
& off & 18.3 & 6.5 & 12.8 & 12.1 \\
& on & 31.1 & 13.6 & 21.5 & 23.2 \\

\texttt{Qwen/Qwen3.5-9B}
& off & 26.1 & 9.6 & 17.9 & 17.8 \\
& on & 35.8 & 16.4 & 26.7 & 25.6 \\

\texttt{Qwen/Qwen3.5-35B-A3B}
& off & 31.2 & 12.1 & 21.8 & 21.5 \\
& on & 42.8 & 20.6 & 30.8 & 32.5 \\

\texttt{Qwen/Qwen3.5-27B}
& off & 42.2 & 20.7 & 30.4 & 32.5 \\
& on & 50.1 & 28.5 & 38.6 & 40.0 \\

\texttt{Qwen/Qwen3.8-27B}
& low & 35.0 & 21.8 & 28.6 & 28.2 \\
& medium & 36.8 & 22.1 & 29.9 & 29.0 \\
& xhigh & 51.1 & 27.2 & 37.1 & 41.2 \\

\midrule
\multicolumn{6}{l}{\textit{DeepSeek}} \\
\texttt{deepseek-v4.1-flash}
& high & 50.7 & 39.3 & 42.8 & 47.2 \\

\midrule
\multicolumn{6}{l}{\textit{Gemini}} \\
\texttt{gemini-2.5-pro}
& low & 23.3 & 9.9 & 17.4 & 15.8 \\
& medium & 44.0 & 25.0 & 35.7 & 33.3 \\
& high & 43.6 & 22.8 & 33.5 & 32.9 \\

\texttt{gemini-3.5-flash-lite}
& low & 22.5 & 13.9 & 16.8 & 19.6 \\
& medium & 37.5 & 21.1 & 28.6 & 30.0 \\
& high & 44.3 & 26.0 & 37.1 & 33.2 \\

\texttt{gemini-3.1-pro-preview}
& low & 57.6 & 39.0 & 47.2 & 49.4 \\
& medium & 64.2 & 45.7 & 54.2 & 55.7 \\
& high & 70.8 & 55.0 & 63.9 & 61.9 \\

\texttt{gemini-3.8-flash}
& low & 75.1 & 47.8 & 59.2 & 63.7 \\
& medium & 89.6 & 66.0 & 77.4 & 78.2 \\
& high & 92.9 & 71.8 & 83.2 & 81.5 \\

\bottomrule
\end{tabular}
\end{table*}

Many configurations score very differently on the two link-principle slices. Gemini 3.8 Flash at high effort, for example, scores 92.9\% under last-link and 71.8\% under weakest-link; corresponding differences are 50.1\% versus 28.5\% for Qwen3.5-27B with thinking enabled and 50.7\% versus 39.3\% for DeepSeek-v4.1-Flash. The benchmark-wide democratic and elitist aggregates are often much closer, but those differences require a different interpretation. Because the set-ordering distinction is formally inactive in many cells, particularly under last-link, differences between the aggregate democratic and elitist columns can reflect both genuinely set-ordering-sensitive instances and variation among independently generated theories. They should therefore not be read as a benchmark-wide estimate of model sensitivity to democratic versus elitist comparison.

The aggregate direction should not be interpreted as weakest-link being uniformly harder. Task-level reversals occur. For example, DeepSeek-v4.1-Flash performs better under weakest-link on \texttt{claim\_chain} (66.7\% versus 53.3\%) and \texttt{defeat\_diagnosis} (73.3\% versus 28.3\%). The ordering analysis therefore describes model sensitivity to different preference regimes rather than defining a universal ordering of task difficulty.

\subsubsection{Task-Specific Last-Link and Weakest-Link Results}

The benchmark-wide ordering aggregates can mask substantial task-level
variation. Tables~\ref{tab:ordering-link-analysis} and
\ref{tab:ordering-link-construction} therefore report complete-task success
separately for last-link and weakest-link within each task. Each task-specific
condition contains 60 frozen benchmark instances for a complete
model--reasoning configuration. Cells report
\textit{last-link (\%) $|$ weakest-link (\%)}.

\begin{table*}[t]
\centering
\tiny
\setlength{\tabcolsep}{2.4pt}
\caption{\textbf{Task-specific last-link versus weakest-link performance on
analysis/evaluation tasks.} Each cell reports complete-task success as
last-link (\%) $|$ weakest-link (\%).}
\label{tab:ordering-link-analysis}
\resizebox{\textwidth}{!}{
\begin{tabular}{@{}llrrrrrr@{}}
\toprule
Model & Reasoning & Formal. & Chain & Diagnosis & Status & Semantics & Perturb. \\
\midrule

\multicolumn{8}{l}{\textit{Gemma}} \\
\texttt{google/gemma-4-E2B-it}
& off & 0.0$|$1.7 & 0.0$|$5.0 & 5.0$|$5.0 & 0.0$|$3.3 & 0.0$|$0.0 & 0.0$|$0.0 \\
& on & 0.0$|$1.7 & 3.3$|$6.7 & 0.0$|$0.0 & 3.3$|$5.0 & 0.0$|$0.0 & 0.0$|$0.0 \\

\texttt{google/gemma-4-E4B-it}
& off & 5.0$|$3.3 & 3.3$|$1.7 & 6.7$|$5.0 & 13.3$|$10.0 & 3.3$|$1.7 & 0.0$|$1.7 \\
& on & 5.0$|$3.3 & 5.0$|$8.3 & 10.0$|$8.3 & 11.7$|$10.0 & 0.0$|$0.0 & 1.7$|$0.0 \\

\texttt{google/gemma-4-26B-A4B-it}
& off & 10.0$|$5.0 & 3.3$|$1.7 & 10.0$|$8.3 & 15.0$|$15.0 & 3.3$|$5.0 & 3.3$|$0.0 \\
& on & 16.7$|$5.0 & 16.7$|$15.0 & 10.0$|$11.7 & 11.7$|$10.0 & 11.7$|$8.3 & 1.7$|$3.3 \\

\texttt{google/gemma-4-31B-it}
& off & 15.0$|$13.3 & 21.7$|$26.7 & 11.7$|$3.3 & 8.3$|$5.0 & 11.7$|$8.3 & 5.0$|$1.7 \\
& on & 53.3$|$48.3 & 30.0$|$31.7 & 30.0$|$21.7 & 25.0$|$15.0 & 21.7$|$23.3 & 5.0$|$8.3 \\

\midrule
\multicolumn{8}{l}{\textit{GPT-OSS}} \\
\texttt{openai/gpt-oss-20b}
& low & 5.0$|$6.7 & 11.7$|$6.7 & 10.0$|$6.7 & 6.7$|$1.7 & 1.7$|$0.0 & 0.0$|$0.0 \\
& medium & 8.3$|$3.3 & 30.0$|$33.3 & 16.7$|$11.7 & 3.3$|$3.3 & 5.0$|$5.0 & 0.0$|$0.0 \\
& high & 13.3$|$8.3 & 36.7$|$43.3 & 20.0$|$13.3 & 5.0$|$5.0 & 3.3$|$0.0 & 0.0$|$0.0 \\

\texttt{openai/gpt-oss-120b}
& low & 10.0$|$5.0 & 28.3$|$26.7 & 10.0$|$5.0 & 6.7$|$5.0 & 1.7$|$5.0 & 0.0$|$0.0 \\
& medium & 13.3$|$5.0 & 46.7$|$51.7 & 13.3$|$8.3 & 20.0$|$10.0 & 10.0$|$11.7 & 0.0$|$0.0 \\
& high & 13.3$|$15.0 & 51.7$|$38.3 & 23.3$|$16.7 & 18.3$|$13.3 & 15.0$|$11.7 & 5.0$|$5.0 \\

\midrule
\multicolumn{8}{l}{\textit{Qwen}} \\
\texttt{Qwen/Qwen3.5-0.8B}
& off & 0.0$|$0.0 & 0.0$|$0.0 & 0.0$|$0.0 & 0.0$|$0.0 & 0.0$|$0.0 & 0.0$|$0.0 \\
& on & 0.0$|$0.0 & 0.0$|$0.0 & 0.0$|$0.0 & 0.0$|$0.0 & 0.0$|$0.0 & 0.0$|$0.0 \\

\texttt{Qwen/Qwen3.5-2B}
& off & 1.7$|$1.7 & 1.7$|$0.0 & 1.7$|$0.0 & 0.0$|$0.0 & 0.0$|$0.0 & 0.0$|$0.0 \\
& on & 0.0$|$1.7 & 0.0$|$5.0 & 8.3$|$3.3 & 1.7$|$0.0 & 0.0$|$0.0 & 0.0$|$0.0 \\

\texttt{Qwen/Qwen3.5-4B}
& off & 6.7$|$5.0 & 35.0$|$28.3 & 16.7$|$8.3 & 18.3$|$6.7 & 13.3$|$10.0 & 0.0$|$3.3 \\
& on & 11.7$|$3.3 & 50.0$|$50.0 & 25.0$|$15.0 & 28.3$|$13.3 & 8.3$|$8.3 & 0.0$|$5.0 \\

\texttt{Qwen/Qwen3.5-9B}
& off & 8.3$|$8.3 & 35.0$|$41.7 & 21.7$|$11.7 & 35.0$|$11.7 & 8.3$|$6.7 & 6.7$|$5.0 \\
& on & 11.7$|$8.3 & 35.0$|$50.0 & 26.7$|$18.3 & 23.3$|$16.7 & 18.3$|$8.3 & 1.7$|$5.0 \\

\texttt{Qwen/Qwen3.8-27B}
& low & 38.3$|$31.7 & 36.7$|$40.0 & 25.0$|$30.0 & 5.0$|$6.7 & 5.0$|$6.7 & 5.0$|$10.0 \\
& medium & 40.0$|$21.7 & 46.7$|$46.7 & 21.7$|$48.3 & 10.0$|$3.3 & 11.7$|$6.7 & 1.7$|$1.7 \\
& xhigh & 35.0$|$33.3 & 58.3$|$65.0 & 63.3$|$36.7 & 71.7$|$16.7 & 21.7$|$15.0 & 25.0$|$23.3 \\

\texttt{Qwen/Qwen3.5-35B-A3B}
& off & 20.0$|$11.7 & 38.3$|$35.0 & 25.0$|$15.0 & 40.0$|$18.3 & 28.3$|$31.7 & 6.7$|$8.3 \\
& on & 10.0$|$15.0 & 45.0$|$43.3 & 36.7$|$26.7 & 56.7$|$21.7 & 21.7$|$26.7 & 16.7$|$15.0 \\

\texttt{Qwen/Qwen3.5-27B}
& off & 20.0$|$16.7 & 48.3$|$43.3 & 23.3$|$16.7 & 38.3$|$13.3 & 23.3$|$20.0 & 8.3$|$6.7 \\
& on & 18.3$|$26.7 & 46.7$|$53.3 & 53.3$|$25.0 & 51.7$|$20.0 & 23.3$|$21.7 & 20.0$|$10.0 \\

\midrule
\multicolumn{8}{l}{\textit{DeepSeek}} \\
\texttt{deepseek-v4.1-flash}
& high & 63.3$|$71.7 & 53.3$|$66.7 & 28.3$|$73.3 & 46.7$|$21.7 & 16.7$|$15.0 & 53.3$|$40.0 \\

\midrule
\multicolumn{8}{l}{\textit{Gemini}} \\
\texttt{gemini-2.5-pro}
& low & 43.3$|$40.0 & 28.3$|$20.0 & 16.7$|$11.7 & 6.7$|$6.7 & 1.7$|$1.7 & 0.0$|$1.7 \\
& medium & 43.3$|$45.0 & 48.3$|$50.0 & 45.0$|$26.7 & 48.3$|$21.7 & 28.3$|$25.0 & 23.3$|$26.7 \\
& high & 43.3$|$36.7 & 43.3$|$55.0 & 50.0$|$26.7 & 53.3$|$21.7 & 25.0$|$18.3 & 21.7$|$23.3 \\

\texttt{gemini-3.5-flash-lite}
& low & 21.7$|$21.7 & 28.3$|$36.7 & 18.3$|$15.0 & 20.0$|$13.3 & 10.0$|$11.7 & 0.0$|$1.7 \\
& medium & 60.0$|$56.7 & 45.0$|$31.7 & 31.7$|$13.3 & 26.7$|$20.0 & 18.3$|$10.0 & 0.0$|$6.7 \\
& high & 80.0$|$88.3 & 56.7$|$43.3 & 53.3$|$16.7 & 35.0$|$16.7 & 20.0$|$18.3 & 10.0$|$10.0 \\

\texttt{gemini-3.1-pro-preview}
& low & 95.0$|$96.7 & 53.3$|$51.7 & 85.0$|$18.3 & 31.7$|$15.0 & 30.0$|$28.3 & 33.3$|$31.7 \\
& medium & 98.3$|$100.0 & 58.3$|$56.7 & 95.0$|$18.3 & 26.7$|$15.0 & 60.0$|$53.3 & 56.7$|$43.3 \\
& high & 95.0$|$100.0 & 68.3$|$48.3 & 100.0$|$21.7 & 46.7$|$18.3 & 75.0$|$78.3 & 66.7$|$61.7 \\

\texttt{gemini-3.8-flash}
& low & 95.0$|$100.0 & 56.7$|$60.0 & 96.7$|$18.3 & 85.0$|$25.0 & 70.0$|$65.0 & 55.0$|$51.7 \\
& medium & 96.7$|$100.0 & 88.3$|$88.3 & 100.0$|$15.0 & 100.0$|$26.7 & 90.0$|$95.0 & 96.7$|$71.7 \\
& high & 98.3$|$100.0 & 80.0$|$91.7 & 100.0$|$15.0 & 96.7$|$26.7 & 98.3$|$98.3 & 95.0$|$78.3 \\

\bottomrule
\end{tabular}}
\end{table*}

\begin{table*}[t]
\centering
\tiny
\setlength{\tabcolsep}{2.4pt}
\caption{\textbf{Task-specific last-link versus weakest-link performance on
constructive tasks.} Each cell reports complete-task success as
last-link (\%) $|$ weakest-link (\%).}
\label{tab:ordering-link-construction}
\resizebox{\textwidth}{!}{
\begin{tabular}{@{}llrrrrrr@{}}
\toprule
Model & Reasoning & Pref. & Counter & Counter+S & Attack & Defence & Attack--Def. \\
\midrule

\multicolumn{8}{l}{\textit{Gemma}} \\
\texttt{google/gemma-4-E2B-it}
& off & 5.0$|$0.0 & 0.0$|$0.0 & 0.0$|$0.0 & 0.0$|$0.0 & 1.7$|$0.0 & 0.0$|$0.0 \\
& on & 6.7$|$0.0 & 0.0$|$0.0 & 0.0$|$0.0 & 0.0$|$0.0 & 5.0$|$0.0 & 0.0$|$0.0 \\

\texttt{google/gemma-4-E4B-it}
& off & 25.0$|$0.0 & 0.0$|$0.0 & 0.0$|$0.0 & 0.0$|$0.0 & 5.0$|$0.0 & 0.0$|$0.0 \\
& on & 16.7$|$0.0 & 0.0$|$0.0 & 0.0$|$0.0 & 0.0$|$0.0 & 8.3$|$0.0 & 0.0$|$1.7 \\

\texttt{google/gemma-4-26B-A4B-it}
& off & 61.7$|$0.0 & 3.3$|$1.7 & 3.3$|$5.0 & 0.0$|$0.0 & 35.0$|$6.7 & 20.0$|$0.0 \\
& on & 70.0$|$0.0 & 10.0$|$6.7 & 23.3$|$8.3 & 8.3$|$10.0 & 40.0$|$5.0 & 30.0$|$5.0 \\

\texttt{google/gemma-4-31B-it}
& off & 81.7$|$0.0 & 10.0$|$0.0 & 21.7$|$8.3 & 15.0$|$1.7 & 53.3$|$3.3 & 16.7$|$1.7 \\
& on & 93.3$|$0.0 & 18.3$|$10.0 & 26.7$|$6.7 & 25.0$|$15.0 & 56.7$|$21.7 & 63.3$|$5.0 \\

\midrule
\multicolumn{8}{l}{\textit{GPT-OSS}} \\
\texttt{openai/gpt-oss-20b}
& low & 16.7$|$0.0 & 5.0$|$1.7 & 5.0$|$0.0 & 5.0$|$3.3 & 28.3$|$3.3 & 23.3$|$1.7 \\
& medium & 36.7$|$0.0 & 31.7$|$6.7 & 15.0$|$8.3 & 18.3$|$13.3 & 66.7$|$10.0 & 48.3$|$10.0 \\
& high & 46.7$|$0.0 & 40.0$|$16.7 & 36.7$|$11.7 & 28.3$|$8.3 & 60.0$|$20.0 & 68.3$|$25.0 \\

\texttt{openai/gpt-oss-120b}
& low & 26.7$|$0.0 & 21.7$|$0.0 & 8.3$|$5.0 & 18.3$|$6.7 & 56.7$|$3.3 & 46.7$|$0.0 \\
& medium & 68.3$|$0.0 & 25.0$|$6.7 & 23.3$|$3.3 & 21.7$|$5.0 & 73.3$|$10.0 & 73.3$|$16.7 \\
& high & 88.3$|$25.0 & 23.3$|$38.3 & 21.7$|$16.7 & 16.7$|$10.0 & 65.0$|$55.0 & 83.3$|$25.0 \\

\midrule
\multicolumn{8}{l}{\textit{Qwen}} \\
\texttt{Qwen/Qwen3.5-0.8B}
& off & 0.0$|$0.0 & 0.0$|$0.0 & 0.0$|$0.0 & 0.0$|$0.0 & 0.0$|$0.0 & 0.0$|$0.0 \\
& on & 0.0$|$0.0 & 0.0$|$0.0 & 0.0$|$0.0 & 0.0$|$0.0 & 0.0$|$0.0 & 0.0$|$0.0 \\

\texttt{Qwen/Qwen3.5-2B}
& off & 1.7$|$0.0 & 0.0$|$0.0 & 0.0$|$0.0 & 0.0$|$0.0 & 0.0$|$0.0 & 0.0$|$0.0 \\
& on & 3.3$|$0.0 & 0.0$|$0.0 & 0.0$|$0.0 & 0.0$|$0.0 & 1.7$|$0.0 & 0.0$|$0.0 \\

\texttt{Qwen/Qwen3.5-4B}
& off & 63.3$|$0.0 & 6.7$|$0.0 & 8.3$|$3.3 & 8.3$|$10.0 & 33.3$|$1.7 & 10.0$|$1.7 \\
& on & 86.7$|$0.0 & 33.3$|$10.0 & 20.0$|$23.3 & 20.0$|$15.0 & 53.3$|$6.7 & 36.7$|$13.3 \\

\texttt{Qwen/Qwen3.5-9B}
& off & 81.7$|$0.0 & 16.7$|$5.0 & 13.3$|$5.0 & 6.7$|$5.0 & 45.0$|$5.0 & 35.0$|$10.0 \\
& on & 86.7$|$0.0 & 31.7$|$10.0 & 50.0$|$38.3 & 28.3$|$20.0 & 53.3$|$11.7 & 63.3$|$10.0 \\

\texttt{Qwen/Qwen3.8-27B}
& low & 65.0$|$1.7 & 26.7$|$25.0 & 56.7$|$45.0 & 26.7$|$16.7 & 53.3$|$38.3 & 76.7$|$10.0 \\
& medium & 71.7$|$6.7 & 31.7$|$23.3 & 40.0$|$41.7 & 23.3$|$10.0 & 73.3$|$41.7 & 70.0$|$13.3 \\
& xhigh & 91.7$|$8.3 & 63.3$|$21.7 & 21.7$|$1.7 & 36.7$|$18.3 & 53.3$|$38.3 & 71.7$|$48.3 \\

\texttt{Qwen/Qwen3.5-35B-A3B}
& off & 78.3$|$1.7 & 16.7$|$5.0 & 25.0$|$6.7 & 11.7$|$0.0 & 48.3$|$5.0 & 36.7$|$6.7 \\
& on & 93.3$|$0.0 & 23.3$|$8.3 & 50.0$|$35.0 & 41.7$|$20.0 & 60.0$|$15.0 & 58.3$|$20.0 \\

\texttt{Qwen/Qwen3.5-27B}
& off & 93.3$|$1.7 & 26.7$|$20.0 & 31.7$|$31.7 & 43.3$|$31.7 & 68.3$|$25.0 & 81.7$|$21.7 \\
& on & 98.3$|$0.0 & 38.3$|$45.0 & 48.3$|$36.7 & 41.7$|$46.7 & 71.7$|$31.7 & 90.0$|$25.0 \\

\midrule
\multicolumn{8}{l}{\textit{DeepSeek}} \\
\texttt{deepseek-v4.1-flash}
& high & 100.0$|$20.0 & 33.3$|$35.0 & 13.3$|$13.3 & 35.0$|$30.0 & 66.7$|$41.7 & 98.3$|$43.3 \\

\midrule
\multicolumn{8}{l}{\textit{Gemini}} \\
\texttt{gemini-2.5-pro}
& low & 83.3$|$6.7 & 5.0$|$1.7 & 5.0$|$5.0 & 13.3$|$16.7 & 53.3$|$5.0 & 23.3$|$1.7 \\
& medium & 88.3$|$18.3 & 23.3$|$23.3 & 23.3$|$8.3 & 16.7$|$13.3 & 58.3$|$21.7 & 81.7$|$20.0 \\
& high & 95.0$|$10.0 & 26.7$|$25.0 & 6.7$|$1.7 & 20.0$|$20.0 & 60.0$|$23.3 & 78.3$|$11.7 \\

\texttt{gemini-3.5-flash-lite}
& low & 40.0$|$0.0 & 16.7$|$1.7 & 5.0$|$3.3 & 16.7$|$6.7 & 60.0$|$30.0 & 33.3$|$25.0 \\
& medium & 83.3$|$0.0 & 25.0$|$20.0 & 13.3$|$8.3 & 21.7$|$15.0 & 75.0$|$46.7 & 50.0$|$25.0 \\
& high & 86.7$|$1.7 & 33.3$|$26.7 & 18.3$|$11.7 & 10.0$|$16.7 & 68.3$|$38.3 & 60.0$|$23.3 \\

\texttt{gemini-3.1-pro-preview}
& low & 98.3$|$0.0 & 45.0$|$48.3 & 13.3$|$23.3 & 40.0$|$43.3 & 81.7$|$58.3 & 85.0$|$53.3 \\
& medium & 100.0$|$5.0 & 48.3$|$65.0 & 30.0$|$11.7 & 33.3$|$38.3 & 68.3$|$63.3 & 95.0$|$78.3 \\
& high & 100.0$|$23.3 & 48.3$|$66.7 & 25.0$|$21.7 & 58.3$|$50.0 & 66.7$|$86.7 & 100.0$|$83.3 \\

\texttt{gemini-3.8-flash}
& low & 100.0$|$3.3 & 53.3$|$50.0 & 33.3$|$26.7 & 71.7$|$71.7 & 90.0$|$76.7 & 95.0$|$25.0 \\
& medium & 100.0$|$15.0 & 80.0$|$73.3 & 40.0$|$23.3 & 98.3$|$98.3 & 85.0$|$96.7 & 100.0$|$88.3 \\
& high & 100.0$|$26.7 & 95.0$|$78.3 & 61.7$|$58.3 & 100.0$|$98.3 & 90.0$|$95.0 & 100.0$|$95.0 \\

\bottomrule
\end{tabular}}
\end{table*}

\FloatBarrier
\subsection{Generation and Answer-Protocol Diagnostics}
\label{app:protocol-diagnostics}

Canonical complete-task success remains the benchmark score throughout the paper. We additionally track generation truncation and whether a response contains the required \texttt{<answer>} region. To quantify the extent to which missing answer regions affect a run, Table~\ref{tab:protocol-diagnostics} also reports success conditioned on the presence of an answer region. This conditioned value is diagnostic only: it divides canonical successes by responses containing an answer region and does not recover or rescore malformed outputs.

\begin{table*}[t]
\centering
\scriptsize
\setlength{\tabcolsep}{4pt}
\caption{\textbf{Generation and answer-protocol diagnostics.} Trunc.\ is the percentage of truncated generations; No answer is the percentage without the required \texttt{<answer>} region; Compliance is the corresponding answer-region compliance rate. Cond.\ denotes success conditioned on the presence of an answer region, and $\Delta$ is conditioned minus canonical success in percentage points.}
\label{tab:protocol-diagnostics}
\begin{tabular}{@{}llrrrrrr@{}}
\toprule
Model & Reasoning & Trunc. & No answer & Compliance & Canonical & Cond. & $\Delta$ \\
\midrule

\multicolumn{8}{l}{\textit{Gemma}} \\
\texttt{google/gemma-4-E2B-it}
& off & 0.4 & 0.8 & 99.2 & 1.1 & 1.1 & +0.0 \\
& on & 0.1 & 20.9 & 79.1 & 1.3 & 1.7 & +0.3 \\

\texttt{google/gemma-4-E4B-it}
& off & 0.0 & 0.2 & 99.8 & 3.5 & 3.5 & +0.0 \\
& on & 0.1 & 4.2 & 95.8 & 3.8 & 3.9 & +0.2 \\

\texttt{google/gemma-4-26B-A4B-it}
& off & 10.3 & 9.9 & 90.1 & 9.0 & 10.0 & +1.0 \\
& on & 1.9 & 2.0 & 98.0 & 14.1 & 14.4 & +0.3 \\

\texttt{google/gemma-4-31B-it}
& off & 1.7 & 2.0 & 98.0 & 14.4 & 14.7 & +0.3 \\
& on & 0.0 & 0.1 & 99.9 & 27.3 & 27.3 & +0.0 \\

\midrule
\multicolumn{8}{l}{\textit{GPT-OSS}} \\
\texttt{openai/gpt-oss-20b}
& low & 0.0 & 3.6 & 96.4 & 6.2 & 6.5 & +0.2 \\
& medium & 0.0 & 4.0 & 96.0 & 16.0 & 16.7 & +0.7 \\
& high & 0.4 & 1.1 & 98.9 & 21.2 & 21.5 & +0.2 \\

\texttt{openai/gpt-oss-120b}
& low & 0.0 & 0.4 & 99.6 & 12.4 & 12.4 & +0.1 \\
& medium & 0.0 & 0.1 & 99.9 & 21.5 & 21.6 & +0.0 \\
& high & 0.0 & 0.4 & 99.7 & 29.0 & 29.1 & +0.1 \\

\midrule
\multicolumn{8}{l}{\textit{Qwen}} \\
\texttt{Qwen/Qwen3.5-0.8B}
& off & 4.7 & 15.9 & 84.1 & 0.0 & 0.0 & +0.0 \\
& on & 99.2 & 67.6 & 32.4 & 0.0 & 0.0 & +0.0 \\

\texttt{Qwen/Qwen3.5-2B}
& off & 1.7 & 16.2 & 83.8 & 0.3 & 0.4 & +0.1 \\
& on & 82.6 & 71.3 & 28.7 & 1.0 & 3.6 & +2.6 \\

\texttt{Qwen/Qwen3.5-4B}
& off & 0.1 & 0.4 & 99.7 & 12.4 & 12.5 & +0.0 \\
& on & 8.3 & 7.8 & 92.2 & 22.4 & 24.3 & +1.9 \\

\texttt{Qwen/Qwen3.5-9B}
& off & 0.4 & 0.2 & 99.8 & 17.8 & 17.9 & +0.0 \\
& on & 1.2 & 4.3 & 95.7 & 26.1 & 27.3 & +1.2 \\

\texttt{Qwen/Qwen3.5-35B-A3B}
& off & 4.4 & 0.7 & 99.3 & 21.7 & 21.8 & +0.2 \\
& on & 3.3 & 2.8 & 97.2 & 31.7 & 32.6 & +0.9 \\

\texttt{Qwen/Qwen3.5-27B}
& off & 0.1 & 0.1 & 99.9 & 31.5 & 31.5 & +0.0 \\
& on & 0.6 & 1.4 & 98.6 & 39.3 & 39.9 & +0.6 \\

\texttt{Qwen/Qwen3.8-27B}
& low & 0.0 & 6.5 & 93.5 & 28.4 & 30.4 & +2.0 \\
& medium & 0.0 & 1.5 & 98.5 & 29.4 & 29.9 & +0.5 \\
& xhigh & 0.2 & 0.2 & 99.8 & 39.2 & 39.2 & +0.1 \\

\midrule
\multicolumn{8}{l}{\textit{DeepSeek}} \\
\texttt{deepseek-v4.1-flash}
& high & 0.0 & 0.0 & 100.0 & 45.0 & 45.0 & +0.0 \\

\midrule
\multicolumn{8}{l}{\textit{Gemini}} \\
\texttt{gemini-2.5-pro}
& low & 0.0 & 16.0 & 84.0 & 16.6 & 19.8 & +3.2 \\
& medium & 0.0 & 10.2 & 89.8 & 34.5 & 38.4 & +3.9 \\
& high & 0.0 & 9.9 & 90.1 & 33.2 & 36.9 & +3.7 \\

\texttt{gemini-3.5-flash-lite}
& low & 0.0 & 0.3 & 99.7 & 18.2 & 18.2 & +0.1 \\
& medium & 0.0 & 0.0 & 100.0 & 29.3 & 29.3 & +0.0 \\
& high & 0.0 & 0.4 & 99.6 & 35.1 & 35.3 & +0.1 \\

\texttt{gemini-3.1-pro-preview}
& low & 0.0 & 0.3 & 99.7 & 48.3 & 48.5 & +0.1 \\
& medium & 0.0 & 0.4 & 99.6 & 54.9 & 55.2 & +0.2 \\
& high & 0.0 & 0.5 & 99.5 & 62.9 & 63.2 & +0.3 \\

\texttt{gemini-3.8-flash}
& low & 0.0 & 0.8 & 99.2 & 61.5 & 61.9 & +0.5 \\
& medium & 0.0 & 0.1 & 99.9 & 77.8 & 77.9 & +0.1 \\
& high & 0.0 & 0.0 & 100.0 & 82.4 & 82.4 & +0.0 \\

\bottomrule
\end{tabular}
\end{table*}

Most configurations have high answer-region compliance. The two smallest Qwen3.5 thinking-on configurations are notable exceptions: Qwen3.5-0.8B and Qwen3.5-2B truncate 99.2\% and 82.6\% of generations and achieve answer-region compliance of 32.4\% and 28.7\%, respectively. Their near-zero canonical success therefore reflects both task performance and severe generation/protocol effects. Across the remaining configurations, conditioning on compliant responses changes success only modestly. The largest differences occur for Gemini 2.5 Pro, where conditioned success exceeds canonical success by 3.2--3.9 percentage points; Qwen3.8-27B at low effort shows a 2.0-point difference, and Qwen3.5-4B with thinking enabled a 1.9-point difference. The principal benchmark conclusions are therefore not explained by answer-protocol failures alone.

There are no scorer refusals or API errors in the final scored runs. Canonical success is retained throughout because producing a valid submitted answer is part of the benchmark contract.

\FloatBarrier

\FloatBarrier
\section{Training Details}
\label{app:training}

This part documents the training experiment of Section~\ref{sec:training}: the setup, checkpoint selection, results at every evaluated checkpoint, and the analyses summarized in the main text.

\subsection{Setup}
\label{app:training-setup}

We post-train Qwen3.5-4B with thinking enabled using GRPO \citep{deepseekmath} with a LoRA adapter \citep{hu2022lora}, implemented on the verl framework \citep{sheng2025hybridflow}. Table~\ref{tab:training-setup} lists the full configuration.

\begin{table}[h]
\centering
\caption{\textbf{Training configuration.}}
\label{tab:training-setup}
\small
\begin{tabular}{@{}p{0.22\linewidth}p{0.74\linewidth}@{}}
\toprule
Aspect & Setting \\
\midrule
Model and adapter & Qwen3.5-4B, thinking enabled; LoRA rank 64, alpha 32, on every linear layer of the language model (vision tower frozen). \\
Algorithm & GRPO with group-standardized advantages; 18 prompts $\times$ 16 completions per step; two optimizer updates per step (mini-batch of 9 prompts); PPO clipping 0.2; token-mean loss; no KL penalty. \\
Rollouts & Temperature 1.0, top-$p$ 0.95, top-$k$ 20, presence penalty 1.5 (the model's recommended settings); log-probabilities are computed after the same penalty and top-$k$/top-$p$ mask. Each step generates its own batch with the current weights; token-level importance weights between the trainer's and the inference engine's log-probabilities are truncated at 2. \\
Reward & 1 for complete-task success and 0 otherwise; 0 for answers truncated at 40,960 tokens. On \texttt{status\_query}, \texttt{defeat\_diagnosis}, \texttt{formalization}, and \texttt{perturbation}, a failed answer earns $0.3 \cdot \max\!\big(0, (s-b)/(1-b)\big)$, where $s$ is the task-native score and $b$ the best score of a fixed uninformed answer on that task and level. \\
Training cells & 82 task--level cells in levels 1--12, kept where the estimated expected reward lies between 0.15 and 0.85 (between 0.12 and 0.85 for the four partial-credit tasks), estimated from the base model's frozen-benchmark evaluation and, for levels 1--8, also from the training groups and step-20 development answers of earlier pilot runs. No test-set result of a trained checkpoint was used. \\
Curriculum & Adaptive sampling over the training cells from running estimates $q$ of each cell's expected reward (exact success, or the paid reward on the four partial-credit tasks), whose counts decay with a half-life of 10 groups of that cell. Cell weight $w(q) = \big(\max(0, q(1-q) - 0.09)/0.16\big)^2 + 0.03$; each step allocates prompts to tasks in proportion to each task's best cell weight, then to that task's cells in proportion to their weights, so cells solved about half the time are favoured, tasks are balanced, and the floor of 0.03 keeps every cell in rotation. \\
Training pool & 8,200 freshly generated problems (100 per cell), with seeds starting at 1,000,000 and disjoint from every evaluation set; 200 steps use 3,600 prompts, less than half a pass. \\
Schedule & Steps 1--100: learning rate $3\times10^{-5}$, 3-step warmup, cosine decay to $3\times10^{-6}$. Steps 101--200: resumed from the step-100 adapter and optimizer state at $1\times10^{-5}$ without warmup, under a cosine schedule toward $1\times10^{-6}$ over 200 steps ($5.4\times10^{-6}$ at step 200). \\
Compute & 3 A100-80GB GPUs: steps 1--100 took 151 hours (453 GPU-hours), steps 101--140 took 56.5 hours (170 GPU-hours), and steps 141--200 took 78.5 hours (235 GPU-hours), about 860 GPU-hours in total for the reported run, including its development evaluations; stopped and diagnostic runs are not included. All evaluations ran on H200 GPUs. \\
\bottomrule
\end{tabular}
\end{table}

\subsection{Checkpoint Selection}
\label{app:training-selection}

Checkpoints were evaluated every 20 steps on a development split of 96 freshly generated problems (one per task and level 1--8), with 8 samples per problem and a 65,536-token output limit. Table~\ref{tab:dev-curve} reports the development curve. We select step 200, the last checkpoint of the run and the best of steps 20--200 on this split; no test-set result was used for selection. The development split shares no instance with the frozen benchmark, the held-out set, or the training pool.

\begin{table}[h]
\centering
\caption{\textbf{Development-split success by training step} (96 problems, 8 samples each). Step 0 is the base model; steps 101--200 belong to the continued run.}
\label{tab:dev-curve}
\small
\setlength{\tabcolsep}{3.5pt}
\begin{tabular}{@{}lrrrrrrrrrrr@{}}
\toprule
Step & 0 & 20 & 40 & 60 & 80 & 100 & 120 & 140 & 160 & 180 & 200 \\
\midrule
Success (\%) & 30.9 & 41.8 & 44.7 & 52.7 & 56.9 & 53.9 & 57.2 & 61.3 & 61.1 & 64.1 & \textbf{64.5} \\
\bottomrule
\end{tabular}
\end{table}

\subsection{Results at Every Checkpoint}
\label{app:training-results}

Table~\ref{tab:training-checkpoints} reports complete-task success at every evaluated checkpoint. Steps 80, 140 and 200 are pooled over three inference runs of the same checkpoint (seeds 0, 1000, and 2000); the other checkpoints are single runs. $\Delta$ is the change from the base model at step 200, with a 95\% paired bootstrap interval over items.

\begin{table}[h]
\centering
\caption{\textbf{Complete-task success (\%) by checkpoint.} Base is the base model re-evaluated alongside the checkpoints under the identical protocol.}
\label{tab:training-checkpoints}
\small
\setlength{\tabcolsep}{3.5pt}
\resizebox{\textwidth}{!}{
\begin{tabular}{@{}lrrrrrrc@{}}
\toprule
& Base & Step 60 & Step 80 & Step 100 & Step 140 & Step 200 & $\Delta$ step 200 [95\% CI] \\
& & (1 run) & (3 runs) & (1 run) & (3 runs) & (3 runs) & \\
\midrule
\multicolumn{8}{@{}l}{\emph{Held-out set (1,440 instances)}} \\
All tasks & 20.4 & 39.6 & 41.7 & 44.4 & 47.9 & 53.3 & $+32.9$ [30.5, 35.3] \\
\quad Construction & 24.4 & 58.1 & 63.1 & 66.8 & 70.6 & 80.0 & $+55.6$ [52.1, 58.9] \\
\quad Analysis & 16.4 & 21.1 & 20.3 & 22.1 & 25.2 & 26.6 & $+10.2$ [7.7, 12.7] \\
\quad Trained cells (L1--12) & 30.3 & 51.5 & 53.6 & 56.4 & 60.7 & 66.0 & $+35.7$ [31.9, 39.5] \\
\quad Untrained cells (L1--12) & 15.5 & 32.9 & 34.4 & 36.7 & 40.7 & 45.2 & $+29.6$ [25.5, 33.7] \\
\quad Levels 13--15 & 6.2 & 24.0 & 27.2 & 30.6 & 31.0 & 38.4 & $+32.2$ [27.1, 37.3] \\
\quad Last-link & 29.3 & 49.4 & 49.0 & 52.8 & 55.1 & 60.0 & $+30.6$ [27.3, 34.1] \\
\quad Weakest-link & 11.5 & 29.7 & 34.4 & 36.1 & 40.6 & 46.7 & $+35.1$ [31.6, 38.7] \\
\midrule
\multicolumn{8}{@{}l}{\emph{Frozen benchmark (1,440 instances)}} \\
All tasks & 20.8 & 41.2 & 43.1 & 44.7 & 48.3 & 52.7 & $+31.8$ [29.9, 33.9] \\
\quad Construction & 25.3 & 61.2 & 64.8 & 67.0 & 73.1 & 79.3 & $+54.0$ [51.3, 56.9] \\
\quad Analysis & 16.4 & 21.2 & 21.3 & 22.4 & 23.6 & 26.1 & $+9.7$ [7.9, 11.5] \\
\bottomrule
\end{tabular}}
\end{table}

\paragraph{Run-to-run variation.} The three inference runs of step 200 give 53.7, 54.0 and 52.2\% on the held-out set and 53.4, 53.0 and 51.6\% on the frozen benchmark. Because all three runs sample the same trained checkpoint, this variation reflects sampling at temperature 1.0, not variation between training runs.

\paragraph{Per-task results.} Table~\ref{tab:training-tasks} reports success per task at every checkpoint. On the held-out set, step 200 improves on eleven of the twelve tasks with intervals excluding zero; claim chain does not change. On the frozen benchmark, all twelve intervals exclude zero.

\begin{table}[h]
\centering
\caption{\textbf{Complete-task success (\%) per task by checkpoint} (120 instances per task). Steps 80, 140 and 200 are pooled over three inference runs. $\Delta$: change from the base model at step 200 with a 95\% paired bootstrap interval.}
\label{tab:training-tasks}
\small
\setlength{\tabcolsep}{3.5pt}
\resizebox{\textwidth}{!}{
\begin{tabular}{@{}llrrrrrrc@{}}
\toprule
& Task & Base & Step 60 & Step 80 & Step 100 & Step 140 & Step 200 & $\Delta$ step 200 [95\% CI] \\
\midrule
\multicolumn{9}{@{}l}{\emph{Held-out set}} \\
Construction & \texttt{attack} & 15.0 & 55.0 & 69.2 & 73.3 & 74.7 & 89.7 & $+74.7$ [67.8, 81.7] \\
 & \texttt{defence} & 21.7 & 61.7 & 65.8 & 75.0 & 79.2 & 89.4 & $+67.8$ [60.3, 75.0] \\
 & \texttt{attack\_defense} & 25.8 & 71.7 & 77.2 & 80.8 & 89.4 & 92.5 & $+66.7$ [58.1, 74.7] \\
 & \texttt{counter\_argument} & 20.0 & 60.0 & 62.5 & 62.5 & 70.6 & 83.6 & $+63.6$ [56.1, 70.6] \\
 & \texttt{counter\_argument\_strict} & 24.2 & 62.5 & 59.4 & 67.5 & 65.0 & 79.2 & $+55.0$ [46.4, 63.6] \\
 & \texttt{preference\_construction} & 40.0 & 37.5 & 44.2 & 41.7 & 44.4 & 45.8 & $+5.8$ [0.6, 11.7] \\
\addlinespace[2pt]
Analysis & \texttt{status\_query} & 13.3 & 25.0 & 23.6 & 25.8 & 29.7 & 33.9 & $+20.6$ [13.9, 27.2] \\
 & \texttt{formalization} & 7.5 & 17.5 & 16.1 & 13.3 & 25.6 & 23.1 & $+15.6$ [10.3, 20.8] \\
 & \texttt{perturbation} & 1.7 & 3.3 & 3.3 & 5.8 & 7.2 & 13.1 & $+11.4$ [6.9, 16.4] \\
 & \texttt{defeat\_diagnosis} & 15.8 & 20.8 & 19.4 & 20.8 & 21.1 & 24.4 & $+8.6$ [4.4, 13.3] \\
 & \texttt{semantics\_query} & 12.5 & 15.0 & 17.2 & 16.7 & 17.8 & 17.8 & $+5.3$ [0.3, 10.3] \\
 & \texttt{claim\_chain} & 47.5 & 45.0 & 42.2 & 50.0 & 49.7 & 47.2 & $-0.3$ [$-9.2$, 8.9] \\
\midrule
\multicolumn{9}{@{}l}{\emph{Frozen benchmark}} \\
Construction & \texttt{attack} & 21.3 & 65.8 & 70.8 & 70.8 & 78.6 & 87.1 & $+65.8$ [60.1, 71.5] \\
 & \texttt{defence} & 27.5 & 63.8 & 67.2 & 75.4 & 83.2 & 88.8 & $+61.2$ [55.0, 67.2] \\
 & \texttt{attack\_defense} & 27.1 & 74.2 & 77.4 & 82.5 & 87.1 & 92.5 & $+65.4$ [59.0, 71.7] \\
 & \texttt{counter\_argument} & 17.1 & 59.2 & 67.2 & 67.1 & 76.9 & 82.8 & $+65.7$ [60.6, 70.6] \\
 & \texttt{counter\_argument\_strict} & 22.5 & 58.3 & 62.1 & 60.4 & 67.5 & 77.5 & $+55.0$ [48.2, 61.8] \\
 & \texttt{preference\_construction} & 36.3 & 45.8 & 44.0 & 45.8 & 45.1 & 46.9 & $+10.7$ [6.9, 14.7] \\
\addlinespace[2pt]
Analysis & \texttt{status\_query} & 21.3 & 30.4 & 28.1 & 30.0 & 34.4 & 36.7 & $+15.4$ [10.4, 21.3] \\
 & \texttt{formalization} & 6.3 & 15.4 & 14.6 & 12.5 & 18.1 & 18.5 & $+12.2$ [7.9, 16.7] \\
 & \texttt{perturbation} & 1.3 & 2.5 & 2.6 & 2.1 & 5.1 & 11.7 & $+10.4$ [6.5, 14.6] \\
 & \texttt{defeat\_diagnosis} & 20.4 & 21.7 & 19.0 & 21.3 & 20.8 & 23.5 & $+3.1$ [1.0, 5.4] \\
 & \texttt{semantics\_query} & 10.0 & 15.4 & 17.8 & 17.1 & 16.1 & 16.9 & $+6.9$ [4.2, 10.1] \\
 & \texttt{claim\_chain} & 39.2 & 42.1 & 46.0 & 51.3 & 46.9 & 49.2 & $+10.0$ [3.5, 16.5] \\
\bottomrule
\end{tabular}}
\end{table}

\paragraph{Truncation.} The base model truncates 9.9\% of its held-out answers at the 65,536-token limit (12.8\% on constructive tasks), against 0.8\% for step 200 (0.4\% on constructive tasks; three runs). Truncation does not explain the constructive gain: even if every truncated base answer were counted as solved, 78\% of the gain (43.2 of 55.6 points) would remain.

\paragraph{Cells without learning signal.} Weakest-link \texttt{preference\_construction} in its generated instances requires a preference between premises as well as between rules. During training it was rewarded in 1 of 2,480 sampled answers, against 2,325 of 2,512 for last-link \texttt{preference\_construction}; no evaluated weakest-link answer succeeds, for the base model or for step 200 (0 of 480 answers on the frozen benchmark). GRPO receives no gradient from a cell whose group rewards are all zero, so this cell receives almost no gradient under the present reward.

\paragraph{Matched link principle.} Step 200 also answered the 120 frozen \texttt{status\_query} theories re-posed under the other link principle (Appendix~\ref{app:ordering-analysis}; run~1, paired with its frozen-benchmark answers of the same run). Last-link theories re-posed under weakest-link are solved less often (55.0\% to 30.0\%, $-25.0$ [$-38.3$, $-11.7$]), and weakest-link theories re-posed under last-link more often (23.3\% to 41.7\%, $+18.3$ [6.7, 31.7]). For the trained model, the effect of the link principle is significant in both directions.

\paragraph{Base-model re-evaluation.} The paper's main results report 22.4\% overall success for Qwen3.5-4B with thinking enabled (one run). Re-evaluated alongside the checkpoints under the identical protocol, the same model reaches 21.2\% and 20.5\% in two samples on the frozen benchmark (20.8\% pooled), a difference consistent with sampling variation at temperature 1.0. All training gains are measured against this re-evaluated base.

\FloatBarrier
\section{External Benchmarks}
\label{app:external}

This part documents the evaluation of the trained model on external benchmarks (Section~\ref{sec:training}): the protocol and scoring rules, results at every checkpoint, and results by subset at step 200.

\subsection{Protocol and Scoring}
\label{app:external-protocol}

The base model and steps 60, 80, 100, 140 and 200 are evaluated on the full test set of each benchmark, with the decoding settings of the ArgGYM evaluation (temperature 1.0, top-$p$ 0.95, top-$k$ 20, presence penalty 1.5, thinking enabled) and a 65,536-token output limit. Each run draws one sample per item, except GPQA Diamond (four samples) and MATH-500 (two samples); item scores are averaged over samples and, for steps 80, 140 and 200, over the three inference runs. A truncated answer counts as incorrect. Intervals are 95\% paired bootstrap intervals over items, stratified by task for BBEH and by step for Belief-R.

Answers are scored as follows. Classification benchmarks are scored by exact match of the extracted label. GSM8K and MATH-500 answers count as correct if either a string normalizer or the \texttt{math-verify} equivalence checker\footnote{\url{https://github.com/huggingface/Math-Verify}} accepts them as matching the gold answer. LiveCodeBench (release v6) programs count as correct if they pass every test case when executed in a sandbox. Belief-R is scored by BREU, the mean of update and maintain accuracy on the revision step; BBEH by its micro-averaged accuracy and by the adjusted harmonic mean over its 23 tasks. On Belief-R, the base model and step 200 both almost never revise a conclusion on its update items (update accuracy 2.4\% and 1.6\%), consistent with the original benchmark's findings \citep{wilie2024belief}, so update accuracy is at floor for both models.

\paragraph{Answers neither model truncated.} Finishing within the 65,536-token limit is part of solving an item, and training reduces truncation on the benchmarks with long answers. To show how much of each gain comes from completing more answers and how much from answering completed items differently, we repeat the comparison on the item-samples that neither the base model nor step 200 (run~1) truncated. This comparison drops exactly the items on which completing the answer is the improvement.

\subsection{Results at Every Checkpoint}
\label{app:external-results}

Table~\ref{tab:external-checkpoints} reports accuracy at every evaluated checkpoint. On $\delta$-NLI, the share of answers without an answer region falls from 2.6\% for the base model to 1.1\% for step 200 (run~1), and accuracy among answers both models gave changes from 84.5\% to 85.6\%.

\begin{table}[h]
\centering
\caption{\textbf{External benchmarks by checkpoint} (accuracy, \%). Steps 80, 140 and 200 are pooled over three inference runs; other checkpoints are single runs. $\Delta$: change from the base model at step 200 with a 95\% paired bootstrap interval.}
\label{tab:external-checkpoints}
\scriptsize
\setlength{\tabcolsep}{3pt}
\resizebox{\textwidth}{!}{
\begin{tabular}{@{}llrrrrrrrc@{}}
\toprule
Domain & Benchmark & Items & Base & Step 60 & Step 80 & Step 100 & Step 140 & Step 200 & $\Delta$ step 200 [95\% CI] \\
\midrule
Defeasible \& non-monotonic & BoardgameQA & 15,000 & 92.8 & 94.6 & 94.8 & 94.9 & 94.7 & 93.9 & $+1.1$ [0.8, 1.5] \\
 & LogicNMR & 10,000 & 94.1 & 95.8 & 95.8 & 95.9 & 96.2 & 95.5 & $+1.4$ [1.0, 1.8] \\
 & $\delta$-NLI & 15,414 & 82.4 & 83.5 & 84.1 & 84.0 & 84.1 & 84.4 & $+2.0$ [1.5, 2.5] \\
 & DEFREASING & 2,392 & 70.0 & 69.8 & 69.5 & 69.8 & 69.7 & 70.2 & $+0.2$ [$-$0.4, 0.8] \\
 & Belief-R (BREU) & 3,656 & 49.9 & 49.6 & 49.6 & 49.7 & 49.6 & 49.6 & $-0.3$ [$-$0.9, 0.3] \\
 & Belief-R (update acc.) & 3,656 & 2.4 & 2.0 & 2.1 & 2.4 & 1.8 & 1.6 & $-0.8$ [$-$1.6, 0.0] \\
\addlinespace[2pt]
Logical reasoning & LogicBench & 2,020 & 84.4 & 83.9 & 84.0 & 84.4 & 84.4 & 84.2 & $-0.2$ [$-$1.0, 0.7] \\
 & Multi-LogiEval & 1,556 & 75.6 & 72.6 & 73.5 & 72.9 & 73.8 & 74.8 & $-0.8$ [$-$1.8, 0.3] \\
\addlinespace[2pt]
Mathematics & GSM8K & 1,319 & 95.7 & 95.1 & 96.0 & 95.7 & 95.7 & 96.2 & $+0.5$ [$-$0.2, 1.2] \\
 & MATH-500 & 500 & 98.0 & 98.2 & 98.7 & 98.2 & 98.4 & 98.6 & $+0.6$ [$-$0.1, 1.3] \\
\addlinespace[2pt]
Knowledge \& science & MMLU-Pro & 12,032 & 79.7 & 80.1 & 80.1 & 80.3 & 79.9 & 80.2 & $+0.5$ [0.1, 0.9] \\
 & GPQA Diamond & 198 & 74.4 & 77.1 & 79.2 & 77.5 & 79.5 & 79.8 & $+5.4$ [3.2, 7.8] \\
\addlinespace[2pt]
Code & LiveCodeBench v6 & 1,055 & 73.2 & 75.9 & 77.4 & 76.4 & 76.9 & 78.5 & $+5.3$ [3.5, 7.0] \\
\addlinespace[2pt]
Broad hard reasoning & BBEH (micro) & 4,520 & 48.3 & 48.5 & 50.1 & 49.0 & 50.6 & 50.9 & $+2.6$ [1.4, 3.7] \\
 & BBEH (harmonic mean) & 4,520 & 32.1 & 34.3 & 35.7 & 33.3 & 36.5 & 35.4 & $+3.2$ [0.3, 7.0] \\
\bottomrule
\end{tabular}}
\end{table}

\subsection{Results by Subset}
\label{app:external-subsets}

Tables~\ref{tab:subsets-small}--\ref{tab:subsets-mmlupro} report every benchmark that has subsets, comparing the base model with step 200 (three runs). GSM8K has no subsets. Subset intervals use 10,000 bootstrap resamples.

On BoardgameQA, the gain grows with reasoning depth in the main subsets ($-1.2$ at depth~1, $+4.4$ at depth~3), with the level of conflict between rules, and with the number of distractors; the binary depth-1 subset declines. On LogicBench, only the binary non-monotonic subset declines ($-2.0$); within it, the largest declines are on the third type of reasoning about exceptions ($-6.7$) and on reasoning about priority ($-6.3$). On BBEH, the largest gains are on Dyck languages ($+13.0$) and multistep arithmetic ($+9.3$), and the largest declines on hyperbaton ($-10.7$) and temporal sequence ($-7.7$).

\begin{table}[h]
\centering
\caption{\textbf{Subsets of LogicNMR, $\delta$-NLI, DEFREASING, Belief-R, GPQA Diamond, LiveCodeBench and MATH-500} (accuracy, \%; step 200 pooled over three runs). $\Delta$: change from the base model with a 95\% paired bootstrap interval. Belief-R rows are the benchmark's own metrics, with the intervals of Table~\ref{tab:external-checkpoints}.}
\label{tab:subsets-small}
\scriptsize
\begin{tabular}{@{}llrrrc@{}}
\toprule
Benchmark & Subset & Items & Base & Step 200 & $\Delta$ [95\% CI] \\
\midrule
LogicNMR & gold true & 3,306 & 92.4 & 94.1 & $+1.6$ [0.7, 2.6] \\
 & gold false & 3,326 & 92.0 & 93.7 & $+1.7$ [0.8, 2.7] \\
 & gold unknown & 3,368 & 97.8 & 98.7 & $+0.9$ [0.5, 1.3] \\
\addlinespace[2pt]
$\delta$-NLI & ATOMIC & 4,138 & 82.8 & 85.0 & $+2.3$ [1.4, 3.2] \\
 & SNLI & 1,837 & 85.2 & 87.3 & $+2.1$ [0.9, 3.3] \\
 & Social & 9,439 & 81.6 & 83.5 & $+1.9$ [1.2, 2.5] \\
\addlinespace[2pt]
DEFREASING & N-alt & 368 & 99.2 & 99.6 & $+0.5$ [$-$0.5, 1.5] \\
 & N-prop & 184 & 92.9 & 91.1 & $-1.8$ [$-$4.5, 0.9] \\
 & S-alt & 368 & 98.9 & 99.4 & $+0.5$ [$-$0.2, 1.3] \\
 & S-case & 736 & 5.4 & 6.1 & $+0.6$ [$-$1.1, 2.4] \\
 & W-case & 736 & 99.7 & 99.9 & $+0.1$ [$-$0.2, 0.6] \\
\addlinespace[2pt]
Belief-R & step $t$ & 1,912 & 98.7 & 98.8 & $+0.1$ [$-$0.3, 0.5] \\
 & step $t{+}1$, maintain & 670 & 97.3 & 97.5 & $+0.1$ [$-$0.7, 1.1] \\
 & step $t{+}1$, update & 1,074 & 2.4 & 1.6 & $-0.8$ [$-$1.6, 0.0] \\
\addlinespace[2pt]
GPQA Diamond & Biology & 19 & 57.9 & 64.9 & $+7.0$ [0.0, 15.4] \\
 & Chemistry & 93 & 61.0 & 69.7 & $+8.7$ [4.6, 13.0] \\
 & Physics & 86 & 92.4 & 94.0 & $+1.6$ [$-$0.1, 3.6] \\
\addlinespace[2pt]
LiveCodeBench v6 & easy & 322 & 97.8 & 98.6 & $+0.7$ [$-$0.3, 2.0] \\
 & medium & 383 & 81.2 & 87.6 & $+6.4$ [3.2, 9.7] \\
 & hard & 350 & 41.7 & 49.9 & $+8.2$ [4.4, 12.0] \\
\addlinespace[2pt]
MATH-500 & level 1 & 43 & 100.0 & 100.0 & $0.0$ [0.0, 0.0] \\
 & level 2 & 90 & 100.0 & 99.8 & $-0.2$ [$-$0.6, 0.0] \\
 & level 3 & 105 & 99.0 & 100.0 & $+1.0$ [0.0, 2.4] \\
 & level 4 & 128 & 98.0 & 98.6 & $+0.5$ [$-$1.2, 2.2] \\
 & level 5 & 134 & 95.1 & 96.1 & $+1.0$ [$-$0.5, 2.9] \\
\bottomrule
\end{tabular}
\end{table}

\begin{table}[h]
\centering
\caption{\textbf{BoardgameQA by subset} (accuracy, \%; 1,000 items each; step 200 pooled over three runs).}
\label{tab:boardgameqa-subsets}
\scriptsize
\begin{tabular}{@{}llrrc@{}}
\toprule
Dimension & Subset & Base & Step 200 & $\Delta$ [95\% CI] \\
\midrule
Depth (main) & depth 1 & 96.0 & 94.8 & $-1.2$ [$-$2.2, $-$0.1] \\
 & depth 2 & 93.5 & 94.5 & $+1.0$ [$-$0.3, 2.3] \\
 & depth 3 & 88.9 & 93.3 & $+4.4$ [2.9, 6.1] \\
\addlinespace[2pt]
Depth (binary) & depth 1 & 94.1 & 89.2 & $-4.9$ [$-$6.3, $-$3.4] \\
 & depth 2 & 91.5 & 92.3 & $+0.8$ [$-$0.5, 2.1] \\
 & depth 3 & 90.2 & 90.7 & $+0.5$ [$-$0.9, 1.9] \\
\addlinespace[2pt]
Conflict & zero & 96.9 & 97.3 & $+0.4$ [$-$0.2, 1.1] \\
 & low & 95.7 & 95.9 & $+0.2$ [$-$0.8, 1.3] \\
 & easy & 94.1 & 95.5 & $+1.4$ [0.3, 2.5] \\
 & high & 91.5 & 93.7 & $+2.2$ [0.9, 3.6] \\
 & difficult & 91.2 & 93.7 & $+2.5$ [1.1, 4.0] \\
\addlinespace[2pt]
Distractors & some & 93.3 & 95.4 & $+2.1$ [0.7, 3.5] \\
 & many & 90.2 & 94.0 & $+3.8$ [2.3, 5.3] \\
\addlinespace[2pt]
Knowledge & light & 92.5 & 94.8 & $+2.3$ [1.1, 3.6] \\
 & heavy & 92.3 & 93.5 & $+1.2$ [$-$0.1, 2.6] \\
\bottomrule
\end{tabular}
\end{table}

\begin{table}[h]
\centering
\caption{\textbf{LogicBench by subset, and its binary non-monotonic items by reasoning type} (accuracy, \%; full test set; step 200 pooled over three runs).}
\label{tab:subsets-logicbench}
\scriptsize
\begin{tabular}{@{}llrrrc@{}}
\toprule
& Subset & Items & Base & Step 200 & $\Delta$ [95\% CI] \\
\midrule
Binary QA & first-order & 520 & 85.0 & 84.3 & $-0.7$ [$-$2.6, 1.2] \\
 & non-monotonic & 440 & 77.0 & 75.1 & $-2.0$ [$-$3.9, $-$0.2] \\
 & propositional & 560 & 86.2 & 87.5 & $+1.2$ [$-$0.2, 2.7] \\
Multiple choice & first-order & 180 & 93.3 & 93.5 & $+0.2$ [$-$1.9, 2.4] \\
 & non-monotonic & 160 & 73.8 & 74.6 & $+0.8$ [$-$2.5, 4.4] \\
 & propositional & 160 & 96.9 & 96.7 & $-0.2$ [$-$2.3, 1.9] \\
\midrule
\multicolumn{6}{@{}l}{\emph{Binary non-monotonic items by reasoning type}} \\
 & default reasoning (default) & 40 & 87.5 & 90.8 & $+3.3$ [0.0, 9.2] \\
 & default reasoning (irrelevant information) & 40 & 80.0 & 79.2 & $-0.8$ [$-$7.5, 5.0] \\
 & default reasoning (open) & 40 & 67.5 & 63.3 & $-4.2$ [$-$13.3, 4.2] \\
 & default reasoning (several defaults) & 80 & 77.5 & 77.1 & $-0.4$ [$-$3.3, 2.5] \\
 & reasoning about exceptions 1 & 80 & 75.0 & 75.4 & $+0.4$ [$-$4.2, 5.0] \\
 & reasoning about exceptions 2 & 40 & 72.5 & 71.7 & $-0.8$ [$-$2.5, 0.0] \\
 & reasoning about exceptions 3 & 40 & 85.0 & 78.3 & $-6.7$ [$-$13.3, $-$1.7] \\
 & reasoning about priority & 80 & 75.0 & 68.8 & $-6.3$ [$-$12.5, $-$0.4] \\
\bottomrule
\end{tabular}
\end{table}

\begin{table}[h]
\centering
\caption{\textbf{Multi-LogiEval by reasoning depth and logic} (accuracy, \%; step 200 pooled over three runs).}
\label{tab:subsets-multilogieval}
\scriptsize
\begin{tabular}{@{}llrrrc@{}}
\toprule
Depth & Logic & Items & Base & Step 200 & $\Delta$ [95\% CI] \\
\midrule
1 & first-order & 130 & 70.0 & 69.0 & $-1.0$ [$-$3.1, 0.8] \\
 & non-monotonic & 160 & 28.1 & 24.6 & $-3.5$ [$-$6.5, $-$1.0] \\
 & propositional & 120 & 92.5 & 92.2 & $-0.3$ [$-$3.9, 3.6] \\
2 & first-order & 105 & 90.5 & 88.6 & $-1.9$ [$-$5.7, 1.9] \\
 & non-monotonic & 236 & 73.7 & 73.2 & $-0.6$ [$-$3.0, 1.6] \\
 & propositional & 105 & 82.9 & 83.8 & $+1.0$ [$-$2.9, 4.8] \\
3 & first-order & 135 & 83.7 & 81.7 & $-2.0$ [$-$5.7, 1.7] \\
 & non-monotonic & 40 & 57.5 & 55.0 & $-2.5$ [$-$10.0, 4.2] \\
 & propositional & 135 & 91.9 & 89.1 & $-2.7$ [$-$6.4, 0.7] \\
4 & first-order & 120 & 83.3 & 89.2 & $+5.8$ [1.7, 10.6] \\
 & non-monotonic & 40 & 77.5 & 71.7 & $-5.8$ [$-$14.2, 0.8] \\
 & propositional & 120 & 80.8 & 83.6 & $+2.8$ [$-$2.8, 8.3] \\
5 & first-order & 45 & 88.9 & 85.2 & $-3.7$ [$-$11.9, 5.2] \\
 & non-monotonic & 20 & 50.0 & 40.0 & $-10.0$ [$-$23.3, 0.0] \\
 & propositional & 45 & 77.8 & 79.3 & $+1.5$ [$-$3.7, 7.4] \\
\bottomrule
\end{tabular}
\end{table}

\begin{table}[h]
\centering
\caption{\textbf{BBEH by task} (accuracy, \%; 200 items per task, 120 for Disambiguation QA; step 200 pooled over three runs).}
\label{tab:subsets-bbeh}
\scriptsize
\begin{tabular}{@{}lrrc@{}}
\toprule
Task & Base & Step 200 & $\Delta$ [95\% CI] \\
\midrule
BoardgameQA & 85.5 & 88.0 & $+2.5$ [$-$2.2, 7.3] \\
Boolean expressions & 52.5 & 54.3 & $+1.8$ [$-$5.2, 8.7] \\
Buggy tables & 43.5 & 45.5 & $+2.0$ [$-$4.3, 8.3] \\
Causal understanding & 54.0 & 57.7 & $+3.7$ [$-$0.3, 7.8] \\
Disambiguation QA & 56.7 & 58.6 & $+1.9$ [$-$5.0, 8.9] \\
Dyck languages & 57.5 & 70.5 & $+13.0$ [5.7, 20.0] \\
Geometric shapes & 28.5 & 31.8 & $+3.3$ [$-$3.2, 9.7] \\
Hyperbaton & 39.0 & 28.3 & $-10.7$ [$-$17.8, $-$3.5] \\
Linguini & 5.5 & 9.5 & $+4.0$ [1.3, 6.8] \\
Movie recommendation & 71.5 & 74.8 & $+3.3$ [$-$2.0, 9.0] \\
Multistep arithmetic & 38.0 & 47.3 & $+9.3$ [4.3, 14.5] \\
NYCC & 14.5 & 12.3 & $-2.2$ [$-$6.8, 2.2] \\
Object counting & 75.5 & 82.2 & $+6.7$ [1.7, 12.0] \\
Object properties & 48.5 & 55.3 & $+6.8$ [$-$0.2, 13.7] \\
SARC triples & 21.0 & 20.8 & $-0.2$ [$-$5.3, 5.0] \\
Shuffled objects & 11.5 & 11.8 & $+0.3$ [$-$5.0, 5.5] \\
Spatial reasoning & 48.0 & 52.0 & $+4.0$ [0.5, 7.5] \\
SportQA & 23.0 & 24.8 & $+1.8$ [$-$1.8, 5.5] \\
Temporal sequence & 67.5 & 59.8 & $-7.7$ [$-$14.0, $-$1.3] \\
Time arithmetic & 76.5 & 83.3 & $+6.8$ [1.2, 12.7] \\
Web of lies & 63.5 & 63.8 & $+0.3$ [$-$6.3, 6.8] \\
Word sorting & 74.5 & 77.2 & $+2.7$ [$-$1.2, 6.7] \\
Zebra puzzles & 57.5 & 62.8 & $+5.3$ [$-$1.5, 12.3] \\
\bottomrule
\end{tabular}
\end{table}

\begin{table}[h]
\centering
\caption{\textbf{MMLU-Pro by subject} (accuracy, \%; step 200 pooled over three runs).}
\label{tab:subsets-mmlupro}
\scriptsize
\begin{tabular}{@{}lrrrc@{}}
\toprule
Subject & Items & Base & Step 200 & $\Delta$ [95\% CI] \\
\midrule
Biology & 717 & 89.4 & 90.3 & $+0.9$ [$-$0.5, 2.4] \\
Business & 789 & 86.8 & 87.2 & $+0.4$ [$-$1.1, 1.8] \\
Chemistry & 1,132 & 90.6 & 90.4 & $-0.2$ [$-$1.1, 0.6] \\
Computer science & 410 & 82.4 & 82.5 & $+0.1$ [$-$1.9, 2.1] \\
Economics & 844 & 83.8 & 85.7 & $+2.0$ [0.5, 3.5] \\
Engineering & 969 & 77.5 & 77.7 & $+0.2$ [$-$1.5, 2.0] \\
Health & 818 & 74.9 & 74.1 & $-0.8$ [$-$2.5, 0.9] \\
History & 381 & 61.4 & 61.7 & $+0.3$ [$-$3.0, 3.5] \\
Law & 1,101 & 53.3 & 54.2 & $+0.9$ [$-$1.3, 3.1] \\
Math & 1,351 & 94.8 & 94.7 & $-0.1$ [$-$0.7, 0.5] \\
Other & 924 & 68.2 & 68.7 & $+0.5$ [$-$1.5, 2.6] \\
Philosophy & 499 & 65.5 & 66.5 & $+1.0$ [$-$1.5, 3.5] \\
Physics & 1,299 & 88.9 & 90.1 & $+1.2$ [0.3, 2.2] \\
Psychology & 798 & 76.9 & 77.9 & $+0.9$ [$-$0.7, 2.5] \\
\bottomrule
\end{tabular}
\end{table}

\FloatBarrier
\section{Relation to Other Defeasible-Reasoning Benchmarks}
\label{app:panel}

This part compares ArgGYM with seven defeasible- and logical-reasoning benchmarks on a common panel of untrained models, to show how ArgGYM's view of these models relates to theirs.

\subsection{Protocol}
\label{app:panel-protocol}

The panel comprises 12 model configurations from Table~\ref{tab:overall-results}, spanning three families and a wide range of ArgGYM success; each configuration's ArgGYM score is its frozen-benchmark complete-task success from that table. Each configuration is evaluated once on fixed subsets of seven benchmarks, under its serving profile from Appendix~\ref{app:model-configs}: BoardgameQA (1,500 items, 100 from each of its 15 subsets), LogicNMR (1,000), $\delta$-NLI (900, 300 per source), DEFREASING (1,196, half of each stratum), Belief-R (1,044 answers, both steps of each sampled problem), and LogicBench (2,020) and Multi-LogiEval (1,556) in full. The subsets are identical for all configurations. Unreadable answers are rare (at most 4\% of any configuration's answers on any benchmark, and none for most), so the scores reflect answers rather than format failures.

\subsection{Results}
\label{app:panel-results}

Table~\ref{tab:panel} reports every configuration's accuracy. The last row gives the Spearman rank correlation between each benchmark and ArgGYM across the 12 configurations: how similarly the two order these models.

\begin{table}[h]
\centering
\caption{\textbf{Panel accuracy (\%)}, ordered by ArgGYM complete-task success. Belief-R is scored by BREU. Last row: Spearman rank correlation with ArgGYM across the 12 configurations (not computed for Belief-R, whose scores barely vary).}
\label{tab:panel}
\scriptsize
\setlength{\tabcolsep}{3pt}
\begin{tabular}{@{}lrrrrrrrr@{}}
\toprule
Configuration & ArgGYM & BoardgameQA & LogicNMR & $\delta$-NLI & DEFREASING & LogicBench & Multi-LogiEval & Belief-R \\
\midrule
Gemma 4 E2B & 1.3 & 78.7 & 34.7 & 76.2 & 83.0 & 81.7 & 68.6 & 50.6 \\
Gemma 4 E4B & 3.8 & 73.2 & 39.9 & 81.6 & 78.8 & 83.3 & 75.3 & 49.5 \\
Gemma 4 26B-A4B & 14.1 & 73.9 & 91.2 & 87.3 & 80.5 & 85.3 & 77.6 & 49.6 \\
GPT-OSS-20B (high) & 21.2 & 95.0 & 98.6 & 81.8 & 72.8 & 84.1 & 76.0 & 49.5 \\
Qwen3.5-4B & 22.4 & 92.3 & 94.7 & 81.7 & 69.9 & 84.4 & 75.6 & 48.7 \\
Qwen3.5-9B & 26.1 & 91.5 & 97.8 & 87.8 & 71.2 & 83.4 & 72.1 & 49.1 \\
Gemma 4 31B & 27.3 & 84.8 & 98.7 & 89.8 & 76.2 & 83.7 & 76.2 & 49.4 \\
GPT-OSS-120B (high) & 29.0 & 96.5 & 98.7 & 87.6 & 70.1 & 85.5 & 76.8 & 49.9 \\
Qwen3.5-27B (thinking off) & 31.5 & 91.2 & 97.9 & 86.8 & 79.8 & 83.9 & 76.6 & 49.7 \\
Qwen3.5-35B-A3B & 31.7 & 90.2 & 97.8 & 88.6 & 70.7 & 84.6 & 74.9 & 48.2 \\
Qwen3.8-27B (xhigh) & 39.2 & 93.5 & 98.9 & 89.7 & 74.1 & 85.0 & 72.7 & 49.5 \\
Qwen3.5-27B & 39.3 & 91.5 & 98.4 & 90.3 & 73.2 & 84.8 & 76.5 & 49.9 \\
\midrule
Rank corr.\ with ArgGYM & -- & +0.46 & +0.70 & +0.82 & $-$0.38 & +0.52 & +0.17 & -- \\
\bottomrule
\end{tabular}
\end{table}

Four observations follow from the table.

\paragraph{ArgGYM spreads the models apart where several benchmarks are near ceiling.} The twelve configurations span 1.3\% to 39.3\% on ArgGYM. On LogicBench they all fall between 81.7\% and 85.5\%; on LogicNMR, ten of the twelve score between 91.2\% and 98.9\%; and eight score at least 90\% on BoardgameQA. Configurations that these benchmarks place close together can differ substantially on ArgGYM: Qwen3.5-4B and Qwen3.5-27B score 84.4\% and 84.8\% on LogicBench but 22.4\% and 39.3\% on ArgGYM.

\paragraph{The ordering agrees most with $\delta$-NLI and LogicNMR.} Models that score higher on ArgGYM tend to score higher on these two benchmarks (rank correlations 0.82 and 0.70); the two smallest Gemma configurations, for instance, are the weakest on both ArgGYM and LogicNMR. The correspondence is weaker for BoardgameQA and LogicBench and weak for Multi-LogiEval.

\paragraph{DEFREASING orders the models differently.} On DEFREASING, three of the four highest scores belong to the three configurations ranked lowest on ArgGYM (83.0\% for Gemma 4 E2B, against 69.9--76.2\% for most configurations that are stronger on ArgGYM), so its ordering tends to run opposite to ArgGYM's (rank correlation $-0.38$).

\paragraph{The differences run in both directions.} Configurations with similar ArgGYM success can differ on other benchmarks: Gemma 4 31B and GPT-OSS-120B (27.3\% and 29.0\% on ArgGYM) score 84.8\% and 96.5\% on BoardgameQA, and Qwen3.5-27B with thinking disabled and Qwen3.5-35B-A3B (31.5\% and 31.7\%) score 79.8\% and 70.7\% on DEFREASING. ArgGYM therefore complements these benchmarks rather than substituting for them.

\paragraph{Belief-R.} All twelve configurations score between 48.2 and 50.6 BREU: every configuration almost always maintains its earlier conclusion and rarely revises it (update accuracy below 6\% for all but Gemma 4 E2B, which revises more often at the cost of maintaining less), the trade-off reported by the benchmark's authors \citep{wilie2024belief}. Belief-R therefore does not distinguish these models.

\FloatBarrier
\section{Extended Discussion}
\label{app:discussion}

The results point to a common pattern in current models: \emph{local
argumentative judgments are more reliable than complete answers}. This distinction is especially important in defeasible
reasoning. Correctly identifying individual supports, attacks, statuses, or
defeaters is not sufficient when the required answer depends on their joint
consequences. Attacks can be defeated, preferences can determine whether an
attack succeeds, and changes to one part of a theory can propagate through
multiple support and defeat relations. An answer can therefore contain many
correct local judgments while still representing the overall argumentative
state incorrectly. The failure analysis of Appendix~\ref{app:failure-analysis} shows that most of these
failures are errors of judgment, and that they are not spread evenly over an
answer: they are associated with claims whose status depends on longer
derivations or remains undecided, rather than being spread evenly over the
claims of a large instance.

The curriculum and ordering results further indicate that this gap depends on
the structure of the argumentation problem, not simply on a single underlying
notion of task difficulty; once the structural properties of the generated
theories are controlled, the curriculum level itself adds little. As generated theories introduce
longer dependencies, interacting attackers and defenders, preference-sensitive
conflicts, and multi-goal structure, complete solutions become less reliable
even when local correctness remains comparatively high. Preference regimes add
another form of structural demand. Under last-link, comparison can depend
primarily on the final defeasible steps of an argument, whereas weakest-link
can make preference information distributed across premises and defeasible
rules relevant to the comparison. The resulting pattern is consistent with a
greater burden on maintaining and combining preference-relevant state under
some weakest-link instances. At the same time, the observed task-level
reversals rule out a simple interpretation in which weakest-link is merely a
harder version of last-link. Model behavior depends on the interaction between
the reasoning operation and the formal regime in which it is performed.

The constructive tasks expose a different limitation. Defeasible reasoning is
not only about determining the state induced by a theory; it can also require
reasoning backwards from a desired argumentative outcome to an intervention
that produces it. The matched counter-argument conditions show that changing the intervention contract does not uniformly make this problem easier for a model. Expanding the available intervention language
changes the search problem, and models differ in whether they recognize and
use the newly available route. This separates two capabilities that ordinary
answer-only evaluation would conflate: understanding which argumentative state
should result and finding an effective sequence of operations that reaches
that state. A model may be strong at evaluating a formal state while still
being unreliable at constructive search over that formal system.

Taken together, the findings suggest at least two separable bottlenecks in
current structured defeasible reasoning: tracking the argumentative state
across long derivations and unresolved conflicts, and searching effectively over interventions that
change the argumentative state. Neither bottleneck is well represented by a
single benchmark-wide accuracy value. Models with similar aggregate scores can
fail for different reasons, and even the strongest evaluated systems retain
operation-specific weaknesses. This matters for both evaluation and training:
improvement on one reasoning operation should not be assumed to imply
improvement on another, and high partial correctness should not be taken as evidence of
complete solutions.

ArgGYM makes these distinctions testable within one executable formal
environment. Because the tasks share the same ASPIC+ substrate, failures can be
localized to support recovery, defeat diagnosis, semantic evaluation,
preference reasoning, revision, or constructive intervention without changing
the underlying argumentation formalism. The procedural generators can then
produce fresh verified instances targeting a particular operation or
structural regime while the frozen benchmark remains unchanged. The training experiment of Section~\ref{sec:training} is a first test of
this kind: verifier rewards improved constructive intervention far more than
state analysis. The same setting allows future work to test whether such gains
transfer to other operations, and whether improvements in local reasoning
translate into more reliable complete argumentative states.

\FloatBarrier
\section{Limitations in Detail}
\label{app:limitations}

We acknowledge that our evaluation does not cover the full range of currently available models. Gemini models are accessed through the provider API, DeepSeek-v4.1-Flash through a hosted API, and the remaining systems in our panel are open-weight models. Other major proprietary API-served systems, including Anthropic Claude and OpenAI's proprietary GPT-series models, are not included. Such systems may exhibit different capability profiles, and evaluating them with ArgGYM would be an important direction for future work.

Secondly, almost all ArgGYM tasks use symbolic argumentation theories to isolate the reasoning problem and enable exact verification. This leaves open how well the same capabilities transfer to more natural language. The \texttt{formalization} task provides a controlled natural-language setting, but its descriptions are template-generated. Extending ArgGYM with more varied language in future work could help separate argumentative competence from effects of lexical choice, semantic plausibility, and world knowledge. The training experiment, however, offers partial evidence: gains from training on symbolic theories transfer to some natural-language defeasible benchmarks (BoardgameQA, LogicNMR) but not to all of them (Section~\ref{sec:training}).

ArgGYM also focuses on a single structured-argumentation framework. Most tasks use grounded semantics. This provides a consistent executable foundation, but it does not establish whether the same findings would hold more broadly across alternative argumentation semantics. Extending ArgGYM in this direction would allow future work to study whether the observed model weaknesses persist under different notions of argumentative acceptance.

Our training experiment covers one model (Qwen3.5-4B), one training run, and one reward design, and the three step-200 evaluations repeat inference on a single checkpoint. We did not ablate the adaptive curriculum and therefore make no claim about its contribution. Training cells were partly selected where the base model had a learning signal, which limits how cleanly the untrained-cell results separate generalization from difficulty.

Finally, it is important to note that the curriculum and ordering analyses should be interpreted as diagnostic. Curriculum configurations can vary several structural properties at once, while last-link and weakest-link instances are generated independently rather than as matched versions of the same underlying theory. Requiring every theory to remain both structurally informative and semantically suitable under both ordering principles would impose a stronger constraint on generation. This would reduce generation efficiency and, more importantly, could bias the accepted instances toward a narrow set of specially constructed theories that happen to behave meaningfully under both orderings. We therefore generate the ordering conditions separately and interpret benchmark-wide differences as systematic associations. A matched re-evaluation of the same status-query theories under both principles (Section~\ref{sec:results}) points the same way in both directions, significantly in one for the untrained model and in both for the trained model (Appendix~\ref{app:training}), and covers a single task.

\end{document}